\documentclass[acmsmall]{acmart}

\acmJournal{PACMPL}
\acmVolume{1}
\acmNumber{CONF} 
\acmArticle{1}
\acmYear{2018}
\acmMonth{1}
\acmDOI{} 
\startPage{1}

\setcopyright{none}

\usepackage{booktabs}   
\usepackage{subcaption} 

\pdfoutput=1
\usepackage{graphicx}
\usepackage{enumerate}
\usepackage{enumitem}

\usepackage{stmaryrd} 
\usepackage{todonotes}
\usepackage{tikzit}
\usepackage{scalefnt}

\tikzstyle{node}=[fill=black, draw=black, shape=circle, scale=0.5]
\tikzstyle{medium_box}=[fill=white, draw=black, shape=rectangle, minimum height=0.8cm, minimum width=0.5cm]

\tikzstyle{pointy}=[->]
\tikzstyle{bluearrow}=[->, fill=none, draw={rgb,255: red,29; green,206; blue,255}, thick]
\tikzstyle{lightnone}=[-, draw={rgb,255: red,191; green,191; blue,191}]

\newtheorem{remark}{Remark}[section]

\DeclareMathOperator{\Ca}{\mathcal{C}}
\DeclareMathOperator{\Da}{\mathcal{D}}

\DeclareMathOperator{\Rb}{\mathbb{R}}

\DeclareMathOperator{\Zb}{\mathbb{Z}}

\newcommand{\R}{\ensuremath{\mathbb R}}
\newcommand{\Z}{\ensuremath{\mathbb Z}}

\newcommand{\Smooth}{\textsf{Smooth}}
\newcommand{\PolyZ}{\mathrm{POLY}_{\Z_2}}

\newcommand{\para}{\mathbf{Para}}
\newcommand{\Para}[1]{\mathbf{Para}(#1)}

\newcommand{\lens}[1]{\mathbf{Lens}(#1)}

\newcommand{\Softmax}{\mathrm{Softmax}}

\newcommand{\<}{\langle}
\renewcommand{\>}{\rangle}

\newcommand{\loss}{\mathsf{loss}}

\newcommand{\gett}{\ensuremath{\mathsf{get}}}
\newcommand{\putt}{\ensuremath{\mathsf{put}}}

\newcommand{\Learn}{\mathbf{Learn}}

\newcommand{\cp}[0]{\ensuremath{\fatsemi}} 

\newcommand{\codelink}[0]{https://github.com/statusfailed/numeric-optics-python/}

\usepackage{listings}
\makeatletter
\def\lst@makecaption{%
  \def\@captype{table}%
  \@makecaption
}
\makeatother

\begin{document}

\title[Categorical Foundations of Gradient-Based Learning]{Categorical Foundations of Gradient-Based Learning}         


\author{G.S.H. Cruttwell}
\affiliation{
  \position{Position1}
  \department{Department of Mathematics and Computer Science} 
  \institution{Mount Allison University}            
  \country{Canada}                    
}

\author{Bruno Gavranovi\'{c}}

\author{Neil Ghani}
\affiliation{
  \institution{University of Strathclyde}            
  \country{UK}
}
\email{bruno@brunogavranovic.com}          

\author{Paul Wilson}
\email{paul@statusfailed.com}         

\author{Fabio Zanasi}
\affiliation{
  \institution{University College London}           
  \country{UK}
}

\begin{abstract}
Text of abstract \ldots.
\end{abstract}



\begin{abstract}
  \label{section:abstract}
  We propose a categorical semantics of gradient-based machine learning algorithms in
terms of lenses, parametrised maps, and reverse derivative categories.  This foundation provides a powerful explanatory and unifying framework: it encompasses a variety of gradient
descent algorithms such as ADAM, AdaGrad, and Nesterov momentum,
as well as a variety of loss functions such as as MSE and Softmax cross-entropy, shedding new light on their similarities and differences.  Our approach to gradient-based learning has examples generalising beyond the familiar continuous domains (modelled in categories of smooth maps) and can be realized in the discrete setting of boolean circuits.
Finally, we demonstrate the practical significance of our framework with an implementation in Python.

\end{abstract}

\maketitle

\section{Introduction}
\label{section:introduction}

The last decade has witnessed a surge of interest in machine learning, fuelled by the numerous successes and applications that these methodologies have found in many fields of science and technology. 
As machine learning techniques become increasingly pervasive, algorithms and models become more sophisticated, posing a significant challenge both to the software developers and the users that need to interface, execute and maintain these systems.
In spite of this rapidly evolving picture, the formal analysis of many
learning algorithms mostly takes place at a heuristic level~\cite{SeshiaS16}, or
using definitions that fail to provide a general and scalable framework for
describing machine learning.
Indeed, it is commonly acknowledged through academia, industry, policy makers
and funding agencies that there is a pressing need for a unifying perspective,
which can make this growing body of work more systematic, rigorous, transparent
and accessible both for users and developers \cite{DeepLearningAdHoc, ExplainableAI}.

\begin{figure*}[ht] 
	\includegraphics[width=7.6cm]{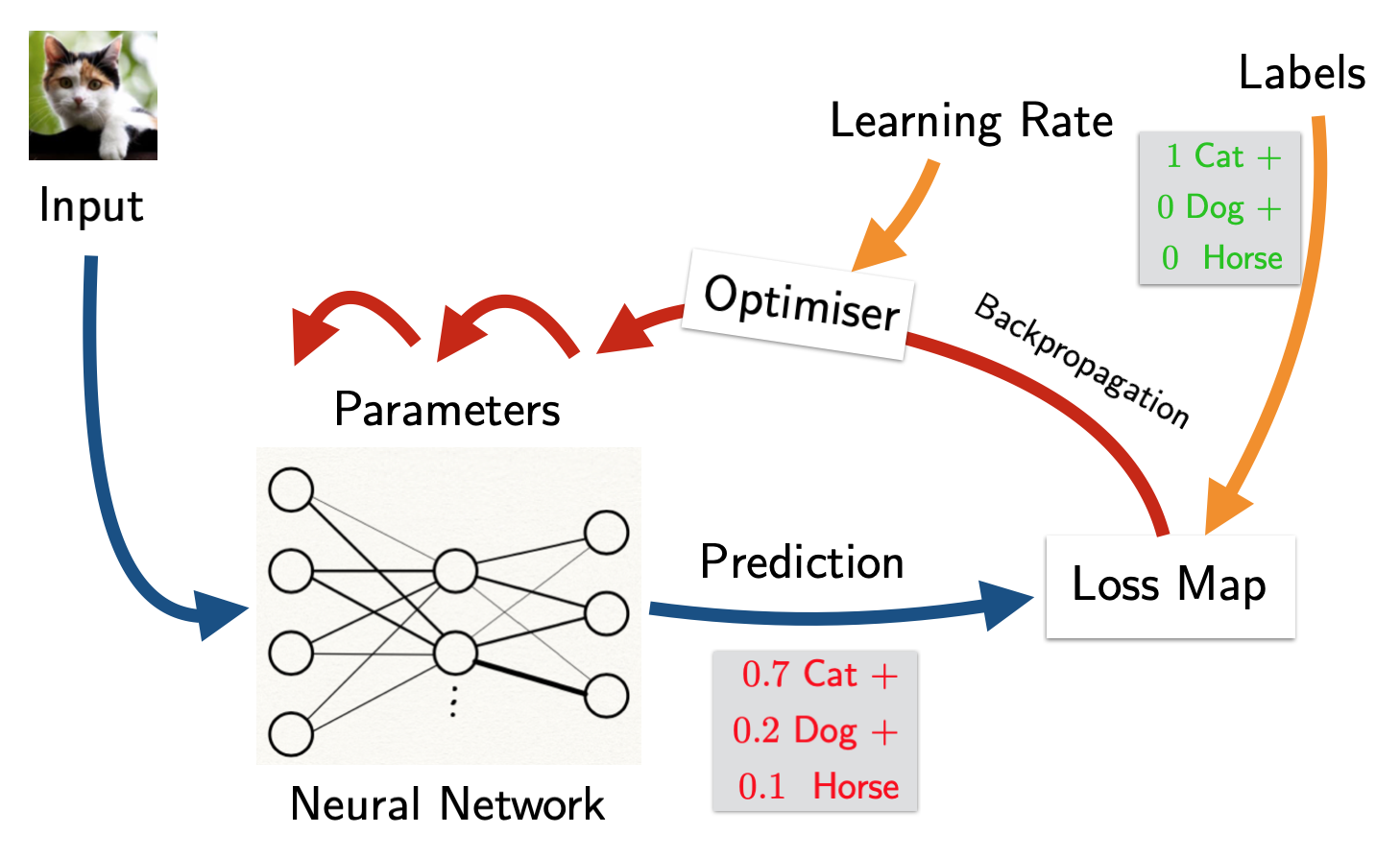} \caption{An
    informal illustration of gradient-based learning. This neural network is
    trained to distinguish different kinds of animals in the input image. Given
    an input $X$, the network predicts an output $Y$, which is compared by a
    `loss map' with what would be the correct answer (`label'). The loss map
    returns a real value expressing the error of the prediction; this
    information, together with the \emph{learning rate} (a weight controlling
    how much the model should be changed in response to error) is used by an
    \emph{optimiser}, which computes by gradient-descent the update of the
    parameters of the network, with the aim of improving its accuracy. The
    neural network, the loss map, the optimiser and the learning rate are all
    components of a supervised learning system, and can vary independently of
    one another.}\label{fig:informalGD}
\end{figure*}

Consider, for example, one of the most common machine
learning scenarios: supervised learning with a neural network. This technique trains the model towards a certain task, e.g. the recognition of patterns in a data set
(\emph{cf.} Figure~\ref{fig:informalGD}). There are several different ways of
implementing this scenario. Typically, at their core, there is a
\emph{gradient update} algorithm (often called the ``optimiser''), depending on a given \emph{loss function}, which updates in steps the parameters of the network, based on some \textit{learning rate} controlling the ``scaling'' of the update.
All of these components can vary independently in a supervised learning
algorithm and a number of choices is available for loss maps (quadratic error,
Softmax cross entropy, dot product, etc.) and optimisers (Adagrad \cite{Adagrad}, Momentum
\cite{Momentum}, and Adam \cite{Adam}, etc.).

This scenario highlights several questions: is there a uniform mathematical language capturing the different components of the learning process?
 Can we develop a unifying
picture of the various optimisation techniques, allowing for their comparative analysis? Moreover, it should be noted that supervised
learning is not limited to neural networks. For example, supervised learning is surprisingly applicable to the
discrete setting of boolean circuits \cite{rda} where continuous functions are replaced by boolean-valued functions. Can we identify an abstract perspective encompassing both the real-valued and the boolean case? In a nutshell, this paper seeks to answer the question:
 \begin{quote}\hspace{-.2cm}{\em what are the fundamental mathematical structures underpinning gradient-based learning?}
 \end{quote}

Our approach to this question stems from the identification of three fundamental aspects of the gradient-descent learning process: \vspace{-.15cm}

\begin{enumerate}[label=(\Roman*)]
	\item computation is \textbf{parameterised}, e.g. in the most simple case we
    are given a function $f : P \times X \to Y$ and learning consists of finding
    a parameter $p : P$ such that $f(p, -)$ is the best function according to
    some criteria. More generally, the weights on the internal nodes of a neural
    network are a parameter which the learning is seeking to optimize.
    Parameters also arise elsewhere, e.g. in the loss function (see later).
	\item information flows \textbf{bidirectionally}: in the forward direction the
    computation turns inputs via a sequence of \textit{layers} into predicted
    outputs, and then into a loss value; in the reverse direction,
    backpropagation is used propagate the changes \textit{backwards} through the layers,
    and then turn them into parameter updates.
	\item the basis of parameter update via gradient descent is \textbf{differentiation} e.g. in the simple case we differentiate the function mapping a parameter to its associated loss to reduce that loss.
  \end{enumerate}

We model bidirectionality via lenses \cite{ProfunctorOptics, Lens,
  BimorphicLenses} and based upon the above three insights, we propose the
notion of \textbf{parametric lens} as the fundamental semantic structure of
learning. In a nutshell, a parametric lens is a process with three kinds of
interfaces: inputs, outputs, and parameters. On each interface, information
flows both ways, i.e. computations are bidirectional. These data are best
explained with our graphical representation of parametric lenses, with inputs
$A$, $A'$, outputs $B$, $B'$, parameters $P$, $P'$, and arrows indicating
information flow (below left). The graphical notation also makes evident that
parametric lenses are \emph{open systems}, which may be composed along their
interfaces (below center and right).
\begin{equation*}
  \quad \qquad
  \scalebox{0.7}{ \begin{tikzpicture}
	\begin{pgfonlayer}{nodelayer}
		\node [style=none] (20) at (-1, 1) {};
		\node [style=none] (21) at (-1, -1) {};
		\node [style=none] (22) at (1, 1) {};
		\node [style=none] (23) at (1, -1) {};
		\node [style=none] (26) at (-2.5, 0.5) {$A$};
		\node [style=none] (27) at (-2.5, -0.5) {$A'$};
		\node [style=none] (28) at (2.5, 0.5) {$B$};
		\node [style=none] (29) at (2.5, -0.5) {$B'$};
		\node [style=none] (39) at (-2, 0.5) {};
		\node [style=none] (40) at (-1, 0.5) {};
		\node [style=none] (43) at (-1, -0.5) {};
		\node [style=none] (44) at (-2, -0.5) {};
		\node [style=none] (45) at (1, 0.5) {};
		\node [style=none] (46) at (2, 0.5) {};
		\node [style=none] (57) at (2, -0.5) {};
		\node [style=none] (58) at (1, -0.5) {};
		\node [style=none] (60) at (-0.5, 1) {};
		\node [style=none] (61) at (0.5, 1) {};
		\node [style=none] (62) at (-0.5, 2) {};
		\node [style=none] (63) at (0.5, 2) {};
		\node [style=none] (64) at (-0.5, 2.5) {$P$};
		\node [style=none] (65) at (0.5, 2.5) {$P'$};
	\end{pgfonlayer}
	\begin{pgfonlayer}{edgelayer}
		\draw (20.center) to (22.center);
		\draw (22.center) to (23.center);
		\draw (23.center) to (21.center);
		\draw (21.center) to (20.center);
		\draw [style=pointy] (39.center) to (40.center);
		\draw [style=pointy] (43.center) to (44.center);
		\draw [style=pointy] (45.center) to (46.center);
		\draw [style=pointy] (57.center) to (58.center);
		\draw [style=pointy] (62.center) to (60.center);
		\draw [style=pointy] (61.center) to (63.center);
	\end{pgfonlayer}
\end{tikzpicture}}
  \qquad\qquad \qquad
  \scalebox{0.6}{\begin{tikzpicture}
	\begin{pgfonlayer}{nodelayer}
		\node [style=none] (20) at (-3, 1) {};
		\node [style=none] (21) at (-3, -1) {};
		\node [style=none] (22) at (-1, 1) {};
		\node [style=none] (23) at (-1, -1) {};
		\node [style=none] (26) at (-4.5, 0.5) {$A$};
		\node [style=none] (27) at (-4.5, -0.5) {$A'$};
		\node [style=none] (28) at (0, 1) {$B$};
		\node [style=none] (29) at (0, -1) {$B'$};
		\node [style=none] (39) at (-4, 0.5) {};
		\node [style=none] (40) at (-3, 0.5) {};
		\node [style=none] (43) at (-3, -0.5) {};
		\node [style=none] (44) at (-4, -0.5) {};
		\node [style=none] (45) at (-1, 0.5) {};
		\node [style=none] (46) at (0, 0.5) {};
		\node [style=none] (57) at (0, -0.5) {};
		\node [style=none] (58) at (-1, -0.5) {};
		\node [style=none] (60) at (-2.5, 1) {};
		\node [style=none] (61) at (-1.5, 1) {};
		\node [style=none] (62) at (-2.5, 2) {};
		\node [style=none] (63) at (-1.5, 2) {};
		\node [style=none] (64) at (-2.5, 2.5) {$P$};
		\node [style=none] (65) at (-1.5, 2.5) {$P'$};
		\node [style=none] (66) at (1, 1) {};
		\node [style=none] (67) at (1, -1) {};
		\node [style=none] (68) at (3, 1) {};
		\node [style=none] (69) at (3, -1) {};
		\node [style=none] (72) at (4.5, 0.5) {$C$};
		\node [style=none] (73) at (4.5, -0.5) {$C'$};
		\node [style=none] (74) at (0, 0.5) {};
		\node [style=none] (75) at (1, 0.5) {};
		\node [style=none] (76) at (1, -0.5) {};
		\node [style=none] (77) at (0, -0.5) {};
		\node [style=none] (78) at (3, 0.5) {};
		\node [style=none] (79) at (4, 0.5) {};
		\node [style=none] (80) at (4, -0.5) {};
		\node [style=none] (81) at (3, -0.5) {};
		\node [style=none] (82) at (1.5, 1) {};
		\node [style=none] (83) at (2.5, 1) {};
		\node [style=none] (84) at (1.5, 2) {};
		\node [style=none] (85) at (2.5, 2) {};
		\node [style=none] (86) at (1.5, 2.5) {$Q$};
		\node [style=none] (87) at (2.5, 2.5) {$Q'$};
	\end{pgfonlayer}
	\begin{pgfonlayer}{edgelayer}
		\draw (20.center) to (22.center);
		\draw (22.center) to (23.center);
		\draw (23.center) to (21.center);
		\draw (21.center) to (20.center);
		\draw [style=pointy] (39.center) to (40.center);
		\draw [style=pointy] (43.center) to (44.center);
		\draw (45.center) to (46.center);
		\draw [style=pointy] (57.center) to (58.center);
		\draw [style=pointy] (62.center) to (60.center);
		\draw [style=pointy] (61.center) to (63.center);
		\draw (66.center) to (68.center);
		\draw (68.center) to (69.center);
		\draw (69.center) to (67.center);
		\draw (67.center) to (66.center);
		\draw [style=pointy] (74.center) to (75.center);
		\draw (76.center) to (77.center);
		\draw [style=pointy] (78.center) to (79.center);
		\draw [style=pointy] (80.center) to (81.center);
		\draw [style=pointy] (84.center) to (82.center);
		\draw [style=pointy] (83.center) to (85.center);
	\end{pgfonlayer}
\end{tikzpicture}}
  \qquad \qquad
  \vcenter{ \scalebox{0.6}{\begin{tikzpicture}
	\begin{pgfonlayer}{nodelayer}
		\node [style=none] (20) at (-1, 1) {};
		\node [style=none] (21) at (-1, -1) {};
		\node [style=none] (22) at (1, 1) {};
		\node [style=none] (23) at (1, -1) {};
		\node [style=none] (26) at (-2.5, 0.5) {$A$};
		\node [style=none] (27) at (-2.5, -0.5) {$A'$};
		\node [style=none] (28) at (2.5, 0.5) {$B$};
		\node [style=none] (29) at (2.5, -0.5) {$B'$};
		\node [style=none] (39) at (-2, 0.5) {};
		\node [style=none] (40) at (-1, 0.5) {};
		\node [style=none] (43) at (-1, -0.5) {};
		\node [style=none] (44) at (-2, -0.5) {};
		\node [style=none] (45) at (1, 0.5) {};
		\node [style=none] (46) at (2, 0.5) {};
		\node [style=none] (57) at (2, -0.5) {};
		\node [style=none] (58) at (1, -0.5) {};
		\node [style=none] (60) at (-0.5, 1) {};
		\node [style=none] (61) at (0.5, 1) {};
		\node [style=none] (62) at (-0.5, 3) {};
		\node [style=none] (63) at (0.5, 3) {};
		\node [style=none] (64) at (-1, 2) {$P$};
		\node [style=none] (65) at (1, 2) {$P'$};
		\node [style=none] (66) at (-1, 5) {};
		\node [style=none] (67) at (-1, 3) {};
		\node [style=none] (68) at (1, 5) {};
		\node [style=none] (69) at (1, 3) {};
		\node [style=none] (74) at (-0.5, 5) {};
		\node [style=none] (75) at (0.5, 5) {};
		\node [style=none] (76) at (-0.5, 6) {};
		\node [style=none] (77) at (0.5, 6) {};
		\node [style=none] (78) at (-0.5, 6.5) {$Q$};
		\node [style=none] (79) at (0.5, 6.5) {$Q'$};
	\end{pgfonlayer}
	\begin{pgfonlayer}{edgelayer}
		\draw (20.center) to (22.center);
		\draw (22.center) to (23.center);
		\draw (23.center) to (21.center);
		\draw (21.center) to (20.center);
		\draw [style=pointy] (39.center) to (40.center);
		\draw [style=pointy] (43.center) to (44.center);
		\draw [style=pointy] (45.center) to (46.center);
		\draw [style=pointy] (57.center) to (58.center);
		\draw [style=pointy] (62.center) to (60.center);
		\draw [style=pointy] (61.center) to (63.center);
		\draw (66.center) to (68.center);
		\draw (68.center) to (69.center);
		\draw (69.center) to (67.center);
		\draw (67.center) to (66.center);
		\draw [style=pointy] (76.center) to (74.center);
		\draw [style=pointy] (75.center) to (77.center);
	\end{pgfonlayer}
\end{tikzpicture}}}
\end{equation*}
This pictorial formalism is not just an intuitive sketch: as we will show, it
can be understood as a completely formal (graphical) syntax using the formalism
of \emph{string diagrams}~\cite{Selinger}, in a way similar to how other
computational phenomena have been recently analysed e.g. in quantum theory
~\cite{coecke_kissinger_2017}, control theory ~\cite{CategoriesInControl,
  BonchiSZ17}, and digital circuit theory ~\cite{GhicaCircuits}. 

It is intuitively clear how parametric lens express aspects (I) and (II) above, whereas
(III) will be achieved by studying them in a space of `differentiable objects'
(in a sense that will be made precise). The main technical contribution of our
paper is showing how the various ingredients involved in learning (the model,
the optimiser, the error map and the learning rate) can be uniformly understood as being built from parametric lenses.


\begin{figure}[h]
  \centering
  {\scalefont{0.8}
    \begin{tikzpicture}
	\begin{pgfonlayer}{nodelayer}
		\node [style=none] (20) at (-3.5, 1) {};
		\node [style=none] (21) at (-3.5, -1) {};
		\node [style=none] (22) at (-1, 1) {};
		\node [style=none] (23) at (-1, -1) {};
		\node [style=none] (26) at (-4.5, 1) {$A$};
		\node [style=none] (27) at (-4.5, -1) {$A'$};
		\node [style=none] (28) at (0, 1) {$B$};
		\node [style=none] (29) at (0, -1) {$B'$};
		\node [style=none] (39) at (-4.5, 0.5) {};
		\node [style=none] (40) at (-3.5, 0.5) {};
		\node [style=none] (43) at (-3.5, -0.5) {};
		\node [style=none] (44) at (-4.5, -0.5) {};
		\node [style=none] (45) at (-1, 0.5) {};
		\node [style=none] (46) at (0, 0.5) {};
		\node [style=none] (57) at (0, -0.5) {};
		\node [style=none] (58) at (-1, -0.5) {};
		\node [style=none] (60) at (-2.75, 1) {};
		\node [style=none] (61) at (-1.75, 1) {};
		\node [style=none] (62) at (-2.75, 2) {};
		\node [style=none] (63) at (-1.75, 2) {};
		\node [style=none] (64) at (-3.25, 2) {$P$};
		\node [style=none] (65) at (-1.25, 2) {$P'$};
		\node [style=none] (66) at (1, 1) {};
		\node [style=none] (67) at (1, -1) {};
		\node [style=none] (68) at (3.5, 1) {};
		\node [style=none] (69) at (3.5, -1) {};
		\node [style=none] (72) at (4.5, 1) {$L$};
		\node [style=none] (73) at (4.5, -1) {$L'$};
		\node [style=none] (74) at (0, 0.5) {};
		\node [style=none] (75) at (1, 0.5) {};
		\node [style=none] (76) at (1, -0.5) {};
		\node [style=none] (77) at (0, -0.5) {};
		\node [style=none] (78) at (3.5, 0.5) {};
		\node [style=none] (79) at (4.5, 0.5) {};
		\node [style=none] (80) at (4.5, -0.5) {};
		\node [style=none] (81) at (3.5, -0.5) {};
		\node [style=none] (82) at (1.75, 1) {};
		\node [style=none] (83) at (2.75, 1) {};
		\node [style=none] (84) at (1.75, 6) {};
		\node [style=node] (85) at (2.75, 2.5) {};
		\node [style=none] (86) at (1.75, 6.5) {$B$};
		\node [style=none] (87) at (3.25, 1.5) {$B'$};
		\node [style=none] (89) at (2.25, 0) {\textsf{Loss}};
		\node [style=none] (92) at (5.5, 1) {};
		\node [style=none] (93) at (5.5, -1) {};
		\node [style=none] (94) at (7.5, -1) {};
		\node [style=none] (95) at (8.5, 0) {};
		\node [style=none] (96) at (7.5, 1) {};
		\node [style=none] (97) at (5.5, -0.5) {};
		\node [style=none] (98) at (5.5, 0.5) {};
		\node [style=none] (99) at (6.725, 0.5) {\textsf{Learning}};
		\node [style=none] (100) at (-3.5, 3) {};
		\node [style=none] (101) at (-1, 3) {};
		\node [style=none] (102) at (-1, 5) {};
		\node [style=none] (103) at (-3.5, 5) {};
		\node [style=none] (105) at (-2.75, 3) {};
		\node [style=none] (106) at (-1.75, 3) {};
		\node [style=none] (108) at (-2.75, 5) {};
		\node [style=none] (110) at (-1.75, 6) {};
		\node [style=none] (111) at (-2.75, 6) {};
		\node [style=none] (117) at (-1.75, 5) {};
		\node [style=none] (120) at (-8, 1) {};
		\node [style=none] (121) at (-8, -1) {};
		\node [style=none] (122) at (-5.5, 1) {};
		\node [style=none] (123) at (-5.5, -1) {};
		\node [style=none] (124) at (-8, 0.5) {};
		\node [style=none] (125) at (-8, -0.5) {};
		\node [style=none] (126) at (-5.5, 0.5) {};
		\node [style=none] (127) at (-5.5, -0.5) {};
		\node [style=none] (128) at (-7.25, 1) {};
		\node [style=none] (129) at (-6.25, 1) {};
		\node [style=none] (130) at (-2.25, 0) {\textsf{Model}};
		\node [style=none] (131) at (-6.75, 0.5) {};
		\node [style=none] (132) at (-6.25, 0) {};
		\node [style=none] (133) at (-5.75, -0.5) {};
		\node [style=none] (134) at (-7.25, 6) {};
		\node [style=node] (135) at (-6.25, 2.5) {};
		\node [style=none] (136) at (-3.25, 6.75) {$P$};
		\node [style=none] (137) at (-1.25, 6.75) {$P$};
		\node [style=none] (138) at (-7.25, 6.75) {$A$};
		\node [style=none] (139) at (-2.25, 4) {\textsf{Optimiser}};
		\node [style=none] (140) at (6.725, -0.475) {\textsf{rate}};
	\end{pgfonlayer}
	\begin{pgfonlayer}{edgelayer}
		\draw (20.center) to (22.center);
		\draw (22.center) to (23.center);
		\draw (23.center) to (21.center);
		\draw (21.center) to (20.center);
		\draw [style=pointy] (39.center) to (40.center);
		\draw (43.center) to (44.center);
		\draw (45.center) to (46.center);
		\draw [style=pointy] (57.center) to (58.center);
		\draw [style=pointy] (62.center) to (60.center);
		\draw (61.center) to (63.center);
		\draw (66.center) to (68.center);
		\draw (68.center) to (69.center);
		\draw (69.center) to (67.center);
		\draw (67.center) to (66.center);
		\draw [style=pointy] (74.center) to (75.center);
		\draw (76.center) to (77.center);
		\draw (78.center) to (79.center);
		\draw [style=pointy] (80.center) to (81.center);
		\draw [style=pointy] (84.center) to (82.center);
		\draw (83.center) to (85);
		\draw (92.center) to (96.center);
		\draw (96.center) to (95.center);
		\draw (95.center) to (94.center);
		\draw (94.center) to (93.center);
		\draw (93.center) to (92.center);
		\draw [style=pointy] (79.center) to (98.center);
		\draw (97.center) to (80.center);
		\draw (103.center) to (102.center);
		\draw (102.center) to (101.center);
		\draw (101.center) to (100.center);
		\draw (100.center) to (103.center);
		\draw (62.center) to (105.center);
		\draw [style=pointy] (63.center) to (106.center);
		\draw (111.center) to (108.center);
		\draw [style=pointy] (117.center) to (110.center);
		\draw [style=lightnone] (120.center) to (122.center);
		\draw [style=lightnone] (122.center) to (123.center);
		\draw [style=lightnone] (123.center) to (121.center);
		\draw [style=lightnone] (121.center) to (120.center);
		\draw (126.center) to (39.center);
		\draw [style=pointy] (44.center) to (127.center);
		\draw (126.center) to (131.center);
		\draw [in=-90, out=-180, looseness=1.25] (131.center) to (128.center);
		\draw [bend right=315, looseness=1.25] (133.center) to (132.center);
		\draw (127.center) to (133.center);
		\draw (132.center) to (129.center);
		\draw [style=pointy] (134.center) to (128.center);
		\draw (129.center) to (135);
	\end{pgfonlayer}
\end{tikzpicture}
    }
  
  \caption[Supervised learning diagram]{The parametric lens that captures the learning process informally
    sketched in Figure~\ref{fig:informalGD}. Note each component is a lens
    itself, whose composition yields the interactions described in
    Figure~\ref{fig:informalGD}. Defining this picture formally will be the
    subject of
    Sections~\ref{section:components-as-lenses}-\ref{section:learning-with-lenses}.
    Also, an animation of this supervised learning system is available online. \footnotemark
  }
  \label{fig:roadmap}
\end{figure}
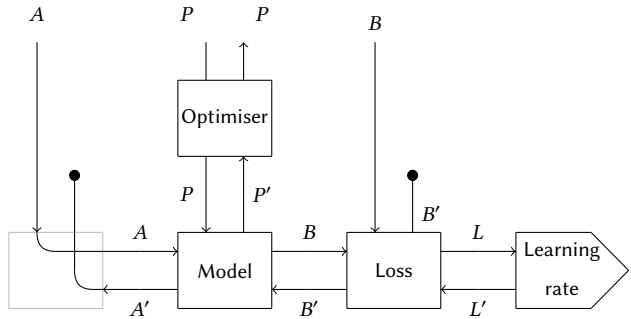

\footnotetext{\label{note1} \url{https://giphy.com/gifs/xXl4CSLvGsd7fmeTOJ} }
We will use \emph{category theory} as the formal language to develop our notion of parametric lenses, and make Figure ~\ref{fig:roadmap} mathematically precise.
The categorical perspective brings several advantages, which are well-known, established principles in programming language semantics~\cite{plotkin_semantics, abramsky_semantics, selinger2001control}. Three of them are particularly important to our contribution, as they constitute distinctive advantages of our semantic foundations:
\begin{description}
	\item[Abstraction] Our approach studies which categorical structures are
    sufficient to perform gradient-based learning. This analysis abstracts away
    from the standard case of neural networks in several different ways: as we
    will see, it encompasses other models (namely Boolean circuits), different
    kinds of optimisers (including Adagrad, Adam, Nesterov momentum), and error
    maps (including quadratic and softmax cross entropy loss). These can be all understood as parametric lenses, and different forms of learning result from their interaction.
	\item[Uniformity] As seen in Figure~\ref{fig:informalGD}, learning involves ingredients that are seemingly quite different: a model, an optimiser, a loss map, etc. We will show how all these notions may be seen as instances of the categorical definition of a parametric lens, thus yielding a remarkably uniform description of the learning process, and supporting our claim of parametric lenses being a fundamental semantic structure of learning.
	\item[Compositionality] The use of categorical structures to describe
    computation naturally enables \emph{compositional reasoning} whereby complex
    systems are analysed in terms of smaller, and hence easier to understand,
    components. Compositionality is a fundamental tenet of programming language
    semantics; in the last few years, it has found application in the  study of
    diverse kinds of computational models, across different fields--- see e.g.
    \cite{FunctorialDataMigration, CompositionalGameTheory,
      coecke_kissinger_2017,BonchiSZ17}. As made evident by
    Figure~\ref{fig:roadmap}, our approach models a neural network as a
    parametric lens, resulting from the \emph{composition} of simpler parametric
    lenses, capturing the different ingredients involved in the learning
    process. Moreover, as all the simpler parametric lenses are themselves
    composable, one may engineer a different learning process by simply plugging
    a new lens on the left or right of existing ones. This means that one can glue together smaller and relatively simple networks to create larger and more sophisticated neural networks.
\end{description}

We now give a synopsis of our contributions:
\begin{itemize}
	\item In Section~\ref{section:background}, we introduce the tools necessary to define our notion of \textbf{parametric lens}. First, in Section~\ref{sec:para}, we introduce a notion of parametrisation for categories, which amounts to a functor $\para(-)$ turning a category $\Ca$ into one $\para(\Ca)$ of `parametrised $\Ca$-maps'. Second, we recall \emph{lenses} (Section~\ref{sec:lenses}). In a nutshell, a lens is a categorical morphism equipped with operations to view and update values in a certain data structure. Lenses play a prominent role in functional programming~\cite{lenses_functional}, as well as in the foundations of database theory~\cite{lenses_database} and more recently game theory~\cite{CompositionalGameTheory}. Considering lenses in $\Ca$ simply amounts to the application of a functorial construction $\lens{-}$, yielding $\lens{\Ca}$. Finally, we recall the notion of a \emph{cartesian reverse differential category} (CRDC): a categorical structure axiomatising the notion of differentiation \cite{CRDC} (Section~\ref{sec:crdc}). We wrap up in Section~\ref{sec:paralenses}, by combining these ingredients into the notion of parametric lens, formally defined as a morphism in $\para(\lens{\Ca})$ for a CRDC $\Ca$. In terms of our desiderata (I)-(III) above, note that $\para(-)$ accounts for (I), $\lens{-}$ accounts for (II), and the CRDC structure accounts for (III).
	\item As seen in Figure~\ref{fig:informalGD}, in the learning process there
    are many components at work: the model, the optimiser, the loss map, the
    learning rate, etc.. In Section~\ref{section:learning-with-lenses}, we show
    how the previously introduced notion of parametric lens provides a uniform characterisation for such components. Moreover, for each of them, we show how different variations appearing in the literature become instances of our abstract characterisation. The plan is as follows: 
	 \begin{itemize}[label=$\circ$]
	 \item In Section~\ref{subsec:models}, we show how the combinatorial \textbf{model} subject of the training can be seen as a parametric lens. The conditions we provide are met by the `standard' case of neural networks, but also enables the study of learning for other classes of models. In particular,  another instance are Boolean circuits: learning of these structures is relevant to binarisation~\cite{BinaryConnect} and it has been explored recently using a categorical approach~\cite{rda}, which turns out to be a particular case of our framework. 
	 \item In Section~\ref{subsec:lossmaps}, we show how the \textbf{loss maps}
     associated with training are also parametric lenses. We also show how our
     approach covers the cases of quadratic error, Boolean error, Softmax cross
     entropy, but also the `dot product loss' associated with the phenomenon of deep
     dreaming~\cite{DeepDreaming1, DeepDreaming2, DeepDreaming3, DeepDreaming4}.
		 \item In Section~\ref{subsec:learningrate}, we model the \textbf{learning rate} as a parametric lens. This analysis also allows us to contrast how learning rate is handled in the `real-valued' case of neural networks with respect to the `Boolean-valued' case of Boolean circuits.
\item In Section~\ref{subsec:optimisers}, we show how \textbf{optimisers} can be
  modelled as `reparametrisations' of models as parametric lenses. As case
  studies, in addition to basic gradient update, we consider the stateful
  variants: Momentum~\cite{Momentum}, Nesterov Momentum~\cite{NesterovMomentum}, Adagrad~\cite{Adagrad}, and Adam (Adaptive Moment Estimation)~\cite{Adam}. Also, on Boolean circuits, we show how the reverse derivative ascent of~\cite{rda} can be also regarded in such way.
\end{itemize}
	\item In Section~\ref{section:learning-with-lenses}, we study how the composition of the lenses defined in Section~\ref{section:components-as-lenses} yields a description of different kinds of learning processes.
	\begin{itemize}[label=$\circ$]
	\item Section~\ref{subsec:learning-parameters} is dedicated to modelling
    supervised \textbf{learning of parameters}, in the way described in
    Figure~\ref{fig:informalGD}. This amounts essentially to study the composite
    of lenses expressed in Figure~\ref{fig:roadmap}, for different choices of
    the various components. In particular we show (i) quadratic loss with basic
    gradient descent, (ii) softmax cross entropy loss with basic gradient
    descent, (iii) quadratic loss with Nesterov momentum, and (iv) learning in
    Boolean circuits with XOR loss and basic gradient ascent.
	\item In order to showcase the flexibility of our approach, in Section~\ref{subsec:deep_dreaming} we depart from our `core' case study of parameter learning, and turn attention to supervised \textbf{learning of inputs}. The idea behind this technique, sometimes called \textbf{deep dreaming}, is that, instead of the network parameters, one updates the inputs, in order to elicit a particular interpretation~\cite{DeepDreaming1, DeepDreaming2,
  DeepDreaming3, DeepDreaming4}.

  Deep dreaming can be easily expressed within our approach, with a different rearrangement of the various parametric lenses involved in the learning process, see Figure~\ref{fig:deep_dreaming} below. The abstract viewpoint of categorical semantics provides a mathematically precise and visually captivating description of the differences between the usual parameter learning process and deep dreaming.
	\end{itemize} 
\item In Section~\ref{section:implementation} we describe a proof-of-concept Python \textbf{implementation}\footnote{
    Available at \href{\codelink}{\codelink}.
} based on the theory developed in this paper. This code is intended to show more concretely the payoff of our approach. Model architectures, as well as the various components participating in the learning process, are now expressed in a uniform, principled mathematical language, in terms of lenses. As a result, computing network gradients is greatly simplified, as it amounts to lens composition. Moreover, the modularity of this approach allows one to more easily tune the various parameters of training. 
	
	We give a demonstration of our library via a number of experiments, and prove
  correctness by achieving accuracy on par with an equivalent model in Keras, a mainstream deep learning framework \cite{Keras}. In particular, we create a working
  non-trivial neural network model for the MNIST image-classification problem \cite{mnist}. \item Finally, in Sections~\ref{section:related-work} and \ref{section:conclusions}, we discuss related and future work.
\end{itemize}

\section{Categorical Toolkit}
\label{section:background}
In this section, we describe the three categorical components of our framework, each corresponding to an aspect of gradient-based learning.
In Section \ref{sec:para} we review the $\para$ construction which builds a category of parametrised maps from a monoidal category, and describe its graphical language.  
In Section \ref{sec:lenses}, we review the $\textbf{Lens}$ construction, which builds a category of ``bidirectional'' maps out of a Cartesian category, and describe its graphical language.   In Section \ref{sec:paralenses}, we look at what happens when we combine these two constructions, and the resulting graphical language of ``parametric lenses''.  In Section \ref{sec:crdc}, we review Cartesian reverse differential categories, a setting for categories equipped with an abstract gradient operator, and how their structure relates to categories of lenses and parameteric lenses.

Then, in the following sections, we will see how these components fit together,
allowing us to describe parametrised models and the algorithms used to train
them.

\subsection{Parametrized Maps}\label{sec:para}

In supervised learning one is typically interested in approximating a function 
	\[ g: \Rb^n \to \Rb^m \]
  for some $n$ and $m$.
  To do this, one begins by building a neural network, which is a smooth map 
	\[ f: \Rb^p \times \Rb^n \to \Rb^m \]
where $\Rb^p$ is the set of possible weights of that neural network.  Then one
looks for a value of $p \in \Rb^p$ such that the function $f(p,-): \Rb^n \to
\Rb^m$ closely approximates $g$.  The first thing we need to is formalize these
types of maps categorically, and this is done via the $\para$ construction
\cite{BackpropAsFunctor, CompDL, TowardsCatCyber}.  

\begin{definition}[Parametrised category] \label{def:paracat}
If $\Ca$ is a strict\footnote{One can also define $\Para{\Ca}$ in the case when $\Ca$ is non-strict; however, the result would be not a category but a bicategory.} symmetric monoidal category (with monoidal product $\otimes$ and monoidal unit $I$) then we define a category $\Para{\Ca}$ with
\begin{itemize}
	\item objects those of $\Ca$
	\item a map from $A$ to $B$ in $\Para{\Ca}$ is a pair $(P,f)$ where $P$ is an
    object of $\Ca$ and $f: P \otimes A \to B$
	\item the identity on $A$ is the pair $(I,1_A)$ (since $\otimes$ is strict monoidal, $I \otimes A = A$)
	\item the composite of $(P,f): A \to B$ with $(P',f'): B \to C$ is the pair 
		\[ (P' \otimes P, (1 \otimes f);g). \]
\end{itemize}
\end{definition}

\begin{example}\label{ex:parasmooth}
Our primary example for the above construction is the category $\Smooth$ whose objects are natural numbers with a map $f: n \to m$ a smooth map from $\Rb^n$ to $\Rb^m$.  As described above, the category $\Para{\Smooth}$ can be thought of as a category of neural networks: a map in this category from $n$ to $m$ consists of a choice of $p$ and a map $f: \Rb^p \times \Rb^n \to \Rb^m$ with $\Rb^p$ representing the set of possible weights of the neural network.  
\end{example}

As anticipated in the introduction, we represent the morphisms of $\Para{\Ca}$ graphically using the formalism of \emph{string diagrams} ---see~\cite{Selinger} for a general overview. As we will see in the next sections, the interplay of the various components at work in the learning process becomes much clearer once represented with this pictorial notation.

 In fact, we will mildly massage the traditional notation for string diagrams, which would represent a morphism in
$\Para{\Ca}$ from $A$ to $B$ as below left.
\[
  \scalebox{0.7}{\begin{tikzpicture}
	\begin{pgfonlayer}{nodelayer}
		\node [style={medium_box}] (0) at (0, 0) {$f$};
		\node [style=none] (1) at (-2, 1) {};
		\node [style=none] (2) at (-2, -1) {};
		\node [style=none] (3) at (-0.5, 0.475) {};
		\node [style=none] (4) at (-0.5, -0.475) {};
		\node [style=none] (5) at (0.25, 0) {};
		\node [style=none] (6) at (2, 0) {};
		\node [style=none] (7) at (-3, 1) {$P$};
		\node [style=none] (8) at (-3, -1) {$A$};
		\node [style=none] (9) at (3, 0) {$B$};
		\node [style=none] (10) at (0, 1.75) {};
	\end{pgfonlayer}
	\begin{pgfonlayer}{edgelayer}
		\draw [style=pointy, in=180, out=0, looseness=0.75] (1.center) to (3.center);
		\draw [style=pointy, in=-180, out=0, looseness=0.75] (2.center) to (4.center);
		\draw [style=pointy] (5.center) to (6.center);
	\end{pgfonlayer}
\end{tikzpicture}}
\qquad \qquad\qquad \scalebox{0.7}{\begin{tikzpicture}
	\begin{pgfonlayer}{nodelayer}
		\node [style={medium_box}] (0) at (0, 0) {$f$};
		\node [style=none] (1) at (0, 1.5) {};
		\node [style=none] (2) at (-2, 0) {};
		\node [style=none] (3) at (0, 0.75) {};
		\node [style=none] (4) at (-0.5, 0) {};
		\node [style=none] (5) at (0.25, 0) {};
		\node [style=none] (6) at (2, 0) {};
		\node [style=none] (7) at (0, 2) {$P$};
		\node [style=none] (8) at (-3, 0) {$A$};
		\node [style=none] (9) at (3, 0) {$B$};
	\end{pgfonlayer}
	\begin{pgfonlayer}{edgelayer}
		\draw [style=pointy, in=-180, out=0, looseness=0.75] (2.center) to (4.center);
		\draw [style=pointy] (5.center) to (6.center);
		\draw [style=pointy] (1.center) to (3.center);
	\end{pgfonlayer}
\end{tikzpicture}}
\]
Note the standard notation does not emphasise the special role played by $P$, which is part of the data of the morphism itself. Parameters and data in machine learning have different semantics: by separating them on two different axes, we obtain a graphical language which is more closely tied to these semantics. Thus, we
will use a slightly different convention for $\Para{\Ca}$, writing a
morphism $(P,f): A \to B$ as on the right above. Incidentally, this clarifies why composition in $\Para{\Ca}$ is defined the way it is: the composite of $(P,f): A \to B$ with $(P',f'): B \to C$ is simply given by hooking up the $B$ wires:

\begin{equation}\label{eq:para-composition-doublecat}
   \scalebox{0.8}{\begin{tikzpicture}
	\begin{pgfonlayer}{nodelayer}
		\node [style={medium_box}] (0) at (0, 0) {$f$};
		\node [style=none] (1) at (0, 2) {};
		\node [style=none] (2) at (-2, 0) {};
		\node [style=none] (3) at (0, 0.25) {};
		\node [style=none] (4) at (-0.25, 0) {};
		\node [style=none] (5) at (0.25, 0) {};
		\node [style=none] (6) at (2, 0) {};
		\node [style=none] (7) at (0, 3) {$P$};
		\node [style=none] (8) at (-3, 0) {$A$};
		\node [style=none] (9) at (2, 0.75) {$B$};
		\node [style={medium_box}] (10) at (4, 0) {$f'$};
		\node [style=none] (11) at (4, 2) {};
		\node [style=none] (12) at (2, 0) {};
		\node [style=none] (13) at (4, 0.25) {};
		\node [style=none] (14) at (3.75, 0) {};
		\node [style=none] (15) at (4.25, 0) {};
		\node [style=none] (16) at (6, 0) {};
		\node [style=none] (17) at (4, 3) {$P'$};
		\node [style=none] (19) at (7, 0) {$C$};
	\end{pgfonlayer}
	\begin{pgfonlayer}{edgelayer}
		\draw [in=-180, out=0, looseness=0.75] (2.center) to (4.center);
		\draw (5.center) to (6.center);
		\draw (1.center) to (3.center);
		\draw [in=-180, out=0, looseness=0.75] (12.center) to (14.center);
		\draw (15.center) to (16.center);
		\draw (11.center) to (13.center);
	\end{pgfonlayer}
\end{tikzpicture}}
\end{equation}

This notation also yields a neat visualisation of ``reparameterisation'', as defined below.  

\begin{definition}\label{def:reparameterisation}
A \textbf{reparametrisation} of $(P,f): A \to B$ in $\Para{\Ca}$ by a map $\alpha: Q \to P$ (below left) is the $\Para{\Ca}$ map $(Q, (\alpha \otimes 1_A);f): A \to B$ (represented below right).
\begin{equation}\label{eq:para-reparametrisation-doublecat}
   \scalebox{0.8}{\begin{tikzpicture}
	\begin{pgfonlayer}{nodelayer}
		\node [style={medium_box}] (0) at (0, 0) {$f$};
		\node [style=none] (1) at (0, 1.5) {};
		\node [style=none] (2) at (-2, 0) {};
		\node [style=none] (3) at (0, 0.75) {};
		\node [style=none] (4) at (-0.5, 0) {};
		\node [style=none] (5) at (0.25, 0) {};
		\node [style=none] (6) at (2, 0) {};
		\node [style=none] (7) at (0, 2) {$P$};
		\node [style=none] (8) at (-3, 0) {$A$};
		\node [style=none] (9) at (3, 0) {$B$};
	\end{pgfonlayer}
	\begin{pgfonlayer}{edgelayer}
		\draw [style=pointy, in=-180, out=0, looseness=0.75] (2.center) to (4.center);
		\draw [style=pointy] (5.center) to (6.center);
		\draw [style=pointy] (1.center) to (3.center);
	\end{pgfonlayer}
\end{tikzpicture}} \qquad \qquad \qquad \scalebox{0.8}{\begin{tikzpicture}
	\begin{pgfonlayer}{nodelayer}
		\node [style={medium_box}] (0) at (0, 0) {$f$};
		\node [style=none] (1) at (0, 2.75) {};
		\node [style=none] (2) at (-2, 0) {};
		\node [style=none] (3) at (0, 0.25) {};
		\node [style=none] (4) at (-0.25, 0) {};
		\node [style=none] (5) at (0.25, 0) {};
		\node [style=none] (6) at (2, 0) {};
		\node [style=none] (7) at (-0.5, 1.5) {$P$};
		\node [style=none] (8) at (-3, 0) {$A$};
		\node [style=none] (9) at (3, 0) {$B$};
		\node [style={medium_box}] (10) at (0, 3) {$\alpha$};
		\node [style=none] (11) at (0, 3.25) {};
		\node [style=none] (12) at (0, 4.5) {};
		\node [style=none] (13) at (0, 5) {$Q$};
	\end{pgfonlayer}
	\begin{pgfonlayer}{edgelayer}
		\draw [in=-180, out=0, looseness=0.75] (2.center) to (4.center);
		\draw (5.center) to (6.center);
		\draw (1.center) to (3.center);
		\draw (12.center) to (11.center);
	\end{pgfonlayer}
\end{tikzpicture}}
\end{equation}
\end{definition}

Intuitively, reparameterisation changes the parameter space of $(P,f): A \to B$ to some other object $Q$, via some map
$\alpha: Q \to P$.  We shall see later that gradient descent and its many
variants can naturally be viewed as reparametrisations. 

Note coherence rules in combining operations \eqref{eq:para-composition-doublecat} and \eqref{eq:para-reparametrisation-doublecat} just work as expected, as these diagrams can be ultimately `compiled' down to string diagrams for monoidal categories.   For example, given maps $(P,f): A \to B$, $(Q,g): B \to C$ with reparametrisations $\alpha: P' \to P$, $\beta: Q' \to Q$, one could either first reparametrise $f$ and $g$ separately and then compose the results (below left), or compose first then reparametrise jointly (below right):
\begin{equation}
   \scalebox{0.8}{\begin{tikzpicture}
	\begin{pgfonlayer}{nodelayer}
		\node [style={medium_box}] (0) at (-2, 0) {$f$};
		\node [style=none] (1) at (-2, 2.75) {};
		\node [style=none] (2) at (-4, 0) {};
		\node [style=none] (3) at (-2, 0.25) {};
		\node [style=none] (4) at (-2.25, 0) {};
		\node [style=none] (5) at (-1.75, 0) {};
		\node [style=none] (6) at (0, 0) {};
		\node [style=none] (7) at (-2.5, 1.5) {$P$};
		\node [style=none] (8) at (-5, 0) {$A$};
		\node [style={medium_box}] (10) at (-2, 3) {$\alpha$};
		\node [style=none] (11) at (-2, 3.25) {};
		\node [style=none] (12) at (-2, 4.5) {};
		\node [style=none] (13) at (-2, 5) {$P'$};
		\node [style={medium_box}] (14) at (6, 0) {$g$};
		\node [style=none] (15) at (6, 2.75) {};
		\node [style=none] (16) at (4, 0) {};
		\node [style=none] (17) at (6, 0.25) {};
		\node [style=none] (18) at (5.75, 0) {};
		\node [style=none] (19) at (6.25, 0) {};
		\node [style=none] (20) at (8, 0) {};
		\node [style=none] (21) at (5.5, 1.5) {$Q$};
		\node [style=none] (23) at (9, 0) {$C$};
		\node [style={medium_box}] (24) at (6, 3) {$\beta$};
		\node [style=none] (25) at (6, 3.25) {};
		\node [style=none] (26) at (6, 4.5) {};
		\node [style=none] (27) at (6, 5) {$Q'$};
		\node [style=none] (28) at (3, 0) {$B$};
		\node [style=none] (29) at (1, 0) {$B$};
		\node [style=none] (30) at (2, 2) {};
		\node [style=none] (31) at (2, 2) {$\cp$};
	\end{pgfonlayer}
	\begin{pgfonlayer}{edgelayer}
		\draw [in=-180, out=0, looseness=0.75] (2.center) to (4.center);
		\draw (5.center) to (6.center);
		\draw (1.center) to (3.center);
		\draw (12.center) to (11.center);
		\draw [in=-180, out=0, looseness=0.75] (16.center) to (18.center);
		\draw (19.center) to (20.center);
		\draw (15.center) to (17.center);
		\draw (26.center) to (25.center);
	\end{pgfonlayer}
\end{tikzpicture}} \qquad \qquad \qquad   \scalebox{0.8}{ \begin{tikzpicture}
	\begin{pgfonlayer}{nodelayer}
		\node [style={medium_box}] (0) at (-2, -2) {$f$};
		\node [style=none] (1) at (-2, 2) {};
		\node [style=none] (2) at (-4, -2) {};
		\node [style=none] (3) at (-2, -1.75) {};
		\node [style=none] (4) at (-2.25, -2) {};
		\node [style=none] (5) at (-1.75, -2) {};
		\node [style=none] (6) at (0, -2) {};
		\node [style=none] (7) at (-2, 0) {$P$};
		\node [style=none] (8) at (-5, -2) {$A$};
		\node [style={medium_box}] (10) at (-2, 2.25) {$\alpha$};
		\node [style=none] (11) at (-2, 2.5) {};
		\node [style=none] (12) at (-2, 3.75) {};
		\node [style=none] (13) at (-2, 4.25) {$P'$};
		\node [style={medium_box}] (14) at (2, -2) {$g$};
		\node [style=none] (15) at (2, 2) {};
		\node [style=none] (16) at (0, -2) {};
		\node [style=none] (17) at (2, -1.75) {};
		\node [style=none] (18) at (1.75, -2) {};
		\node [style=none] (19) at (2.25, -2) {};
		\node [style=none] (20) at (4, -2) {};
		\node [style=none] (21) at (2, 0) {$Q$};
		\node [style=none] (23) at (5, -2) {$C$};
		\node [style={medium_box}] (24) at (2, 2.25) {$\beta$};
		\node [style=none] (25) at (2, 2.5) {};
		\node [style=none] (26) at (2, 3.75) {};
		\node [style=none] (27) at (2, 4.25) {$Q'$};
		\node [style=none] (32) at (-2, -0.75) {};
		\node [style=none] (33) at (2, -0.75) {};
		\node [style=none] (34) at (-2, 0.75) {};
		\node [style=none] (35) at (2, 0.75) {};
	\end{pgfonlayer}
	\begin{pgfonlayer}{edgelayer}
		\draw [in=-180, out=0, looseness=0.75] (2.center) to (4.center);
		\draw (5.center) to (6.center);
		\draw (12.center) to (11.center);
		\draw [in=-180, out=0, looseness=0.75] (16.center) to (18.center);
		\draw (19.center) to (20.center);
		\draw (26.center) to (25.center);
		\draw (3.center) to (32.center);
		\draw (17.center) to (33.center);
		\draw (15.center) to (35.center);
		\draw (1.center) to (34.center);
	\end{pgfonlayer}
\end{tikzpicture}}
\end{equation}
As expected, translating these two operations into string diagrams for monoidal categories yield equivalent representations of the same morphism. 
\begin{equation}\label{eq:para-language-rules-3}
  \scalebox{0.8}{ \begin{tikzpicture}
	\begin{pgfonlayer}{nodelayer}
		\node [style=none] (50) at (-4, -1) {};
		\node [style=none] (52) at (-5, 2) {};
		\node [style=none] (53) at (-5, 0.5) {};
		\node [style=none] (54) at (-5, -1) {};
		\node [style=none] (55) at (-4, 0.5) {};
		\node [style=none] (56) at (-4, 1) {};
		\node [style=none] (57) at (-3, 1) {};
		\node [style=none] (58) at (-3, 0) {};
		\node [style=none] (59) at (-4, 0) {};
		\node [style=none] (60) at (-3.5, 0.5) {$\alpha$};
		\node [style=none] (61) at (-3, 0.5) {};
		\node [style=none] (62) at (-2, 0.5) {};
		\node [style=none] (63) at (-1, 0) {};
		\node [style=none] (64) at (-1, 0.25) {};
		\node [style=none] (65) at (0, 0.25) {};
		\node [style=none] (66) at (0, -0.75) {};
		\node [style=none] (67) at (-1, -0.75) {};
		\node [style=none] (68) at (-0.5, -0.25) {$f$};
		\node [style=none] (69) at (0, -0.25) {};
		\node [style=none] (70) at (-1, -0.5) {};
		\node [style=none] (71) at (-2, -1) {};
		\node [style=none] (72) at (0, -0.25) {};
		\node [style=none] (73) at (0, 2) {};
		\node [style=none] (74) at (0, -0.25) {};
		\node [style=none] (75) at (0, 2) {};
		\node [style=none] (76) at (0, -0.25) {};
		\node [style=none] (77) at (0, 2) {};
		\node [style=none] (78) at (0, 2.5) {};
		\node [style=none] (79) at (1, 2.5) {};
		\node [style=none] (80) at (1, 1.5) {};
		\node [style=none] (81) at (0, 1.5) {};
		\node [style=none] (82) at (0.5, 2) {$\beta$};
		\node [style=none] (83) at (1, 2) {};
		\node [style=none] (84) at (2, 2) {};
		\node [style=none] (85) at (3, 1.25) {};
		\node [style=none] (86) at (3, 1.5) {};
		\node [style=none] (87) at (4, 1.5) {};
		\node [style=none] (88) at (4, 0.5) {};
		\node [style=none] (89) at (3, 0.5) {};
		\node [style=none] (90) at (3.5, 1) {$g$};
		\node [style=none] (91) at (4, 1) {};
		\node [style=none] (92) at (3, 0.75) {};
		\node [style=none] (93) at (2, -0.25) {};
		\node [style=none] (94) at (5, 1) {};
		\node [style=none] (140) at (-2, 1) {$P$};
		\node [style=none] (141) at (-6, 0.5) {$P'$};
		\node [style=none] (142) at (-6, 2) {$Q'$};
		\node [style=none] (143) at (2.5, 2.5) {$Q$};
		\node [style=none] (144) at (-6, -1) {};
		\node [style=none] (145) at (-6, -1) {$A$};
		\node [style=none] (146) at (1, -0.75) {};
		\node [style=none] (147) at (1, -0.75) {$B$};
		\node [style=none] (148) at (6, 1) {$C$};
	\end{pgfonlayer}
	\begin{pgfonlayer}{edgelayer}
		\draw (54.center) to (50.center);
		\draw (56.center) to (57.center);
		\draw (57.center) to (58.center);
		\draw (58.center) to (59.center);
		\draw (59.center) to (56.center);
		\draw (53.center) to (55.center);
		\draw (61.center) to (62.center);
		\draw (64.center) to (65.center);
		\draw (65.center) to (66.center);
		\draw (66.center) to (67.center);
		\draw (67.center) to (64.center);
		\draw (50.center) to (71.center);
		\draw [in=-180, out=0] (62.center) to (63.center);
		\draw [in=180, out=0] (71.center) to (70.center);
		\draw (69.center) to (72.center);
		\draw (76.center) to (74.center);
		\draw (78.center) to (79.center);
		\draw (79.center) to (80.center);
		\draw (80.center) to (81.center);
		\draw (81.center) to (78.center);
		\draw (75.center) to (77.center);
		\draw (83.center) to (84.center);
		\draw (86.center) to (87.center);
		\draw (87.center) to (88.center);
		\draw (88.center) to (89.center);
		\draw (89.center) to (86.center);
		\draw (74.center) to (93.center);
		\draw [in=-180, out=0] (84.center) to (85.center);
		\draw [in=180, out=0] (93.center) to (92.center);
		\draw (91.center) to (94.center);
		\draw (52.center) to (77.center);
	\end{pgfonlayer}
\end{tikzpicture}} \qquad = \qquad  \scalebox{0.8}{\begin{tikzpicture}
	\begin{pgfonlayer}{nodelayer}
		\node [style=none] (50) at (-4, -1) {};
		\node [style=none] (52) at (-5, 2) {};
		\node [style=none] (53) at (-5, 0.5) {};
		\node [style=none] (54) at (-5, -1) {};
		\node [style=none] (55) at (-4, 0.5) {};
		\node [style=none] (56) at (-4, 1) {};
		\node [style=none] (57) at (-3, 1) {};
		\node [style=none] (58) at (-3, 0) {};
		\node [style=none] (59) at (-4, 0) {};
		\node [style=none] (60) at (-3.5, 0.5) {$\alpha$};
		\node [style=none] (61) at (-3, 0.5) {};
		\node [style=none] (62) at (-2, 0.5) {};
		\node [style=none] (63) at (-1, 0) {};
		\node [style=none] (64) at (-1, 0.25) {};
		\node [style=none] (65) at (0, 0.25) {};
		\node [style=none] (66) at (0, -0.75) {};
		\node [style=none] (67) at (-1, -0.75) {};
		\node [style=none] (68) at (-0.5, -0.25) {$f$};
		\node [style=none] (69) at (0, -0.25) {};
		\node [style=none] (70) at (-1, -0.5) {};
		\node [style=none] (71) at (-2, -1) {};
		\node [style=none] (72) at (0, -0.25) {};
		\node [style=none] (73) at (-4, 2) {};
		\node [style=none] (74) at (0, -0.25) {};
		\node [style=none] (75) at (-4, 2) {};
		\node [style=none] (76) at (0, -0.25) {};
		\node [style=none] (77) at (-4, 2) {};
		\node [style=none] (78) at (-4, 2.5) {};
		\node [style=none] (79) at (-3, 2.5) {};
		\node [style=none] (80) at (-3, 1.5) {};
		\node [style=none] (81) at (-4, 1.5) {};
		\node [style=none] (82) at (-3.5, 2) {$\beta$};
		\node [style=none] (83) at (-3, 2) {};
		\node [style=none] (84) at (1, 2) {};
		\node [style=none] (85) at (2, 1.25) {};
		\node [style=none] (86) at (2, 1.5) {};
		\node [style=none] (87) at (3, 1.5) {};
		\node [style=none] (88) at (3, 0.5) {};
		\node [style=none] (89) at (2, 0.5) {};
		\node [style=none] (90) at (2.5, 1) {$g$};
		\node [style=none] (91) at (3, 1) {};
		\node [style=none] (92) at (2, 0.75) {};
		\node [style=none] (93) at (1, -0.25) {};
		\node [style=none] (94) at (4, 1) {};
		\node [style=none] (140) at (-2, 1) {$P$};
		\node [style=none] (141) at (-6, 0.5) {$P'$};
		\node [style=none] (142) at (-6, 2) {$Q'$};
		\node [style=none] (143) at (-2, 2.5) {$Q$};
		\node [style=none] (144) at (-6, -1) {};
		\node [style=none] (145) at (-6, -1) {$A$};
		\node [style=none] (146) at (1, -0.75) {};
		\node [style=none] (147) at (1, -0.75) {$B$};
		\node [style=none] (148) at (5, 1) {$C$};
	\end{pgfonlayer}
	\begin{pgfonlayer}{edgelayer}
		\draw (54.center) to (50.center);
		\draw (56.center) to (57.center);
		\draw (57.center) to (58.center);
		\draw (58.center) to (59.center);
		\draw (59.center) to (56.center);
		\draw (53.center) to (55.center);
		\draw (61.center) to (62.center);
		\draw (64.center) to (65.center);
		\draw (65.center) to (66.center);
		\draw (66.center) to (67.center);
		\draw (67.center) to (64.center);
		\draw (50.center) to (71.center);
		\draw [in=-180, out=0] (62.center) to (63.center);
		\draw [in=180, out=0] (71.center) to (70.center);
		\draw (69.center) to (72.center);
		\draw (76.center) to (74.center);
		\draw (78.center) to (79.center);
		\draw (79.center) to (80.center);
		\draw (80.center) to (81.center);
		\draw (81.center) to (78.center);
		\draw (75.center) to (77.center);
		\draw (83.center) to (84.center);
		\draw (86.center) to (87.center);
		\draw (87.center) to (88.center);
		\draw (88.center) to (89.center);
		\draw (89.center) to (86.center);
		\draw (74.center) to (93.center);
		\draw [in=-180, out=0] (84.center) to (85.center);
		\draw [in=180, out=0] (93.center) to (92.center);
		\draw (91.center) to (94.center);
		\draw (52.center) to (77.center);
	\end{pgfonlayer}
\end{tikzpicture}}
\end{equation}

\begin{remark} There is a 2-categorical perspective on $\Para{\Ca}$, which we glossed over in this paper for the sake of simplicity. In particular, the reparametrisations described above can also be seen as equipping
$\Para{\Ca}$ with 2-cells, giving a 2-categorical structure on $\Para{\Ca}$.
This is also coherent with respect to base change: if $\Ca$ and $\Da$ are strict symmetric monoidal categories, and $F \colon \Ca \to \Da$ a lax symmetric monoidal functor, then there is an induced 2-functor $\Para{F} \colon  \Para{\Ca} \to \Para{\Da}$ which agrees with $F$ on objects. This 2-functor is straightforward: for a 1-cell $(P,f): A \to B$, it applies $F$ to $P$ and $f$ and uses the (lax) comparison to get a map of the correct type.
We will see how this base change becomes important when performing
backpropagation on parameterised maps (Eq. \ref{eq:para_rdc})

Lastly, we mention that $\Para{\Ca}$ inherits the symmetric monoidal structure
from $\Ca$ and that the induced 2-functor $\Para{F}$ respects that structure.
This will allow us to compose neural networks not only in series, but also in
parallel.  For more detail on alternative viewpoints on the $\para$ construction, including how it can be viewed as the Grothendieck construction of a certain indexed category, see \cite{TowardsCatCyber}.  
\end{remark}

\subsection{Lenses}\label{sec:lenses}

We next consider a very different categorical construction.  In machine learning (or even learning in general) it is fundamental that information flows
both forwards and backwards: the `forward' flow corresponds to a model's predictions,
and the `backwards' flow to \emph{corrections} to the model.
The category of lenses is the ideal setting to capture this type of structure,
as it is a category consisting of maps with both a ``forward'' and a
``backward'' part.

\begin{definition}
For any Cartesian category $\Ca$, the category of lenses\footnote{These lenses
  are often called \textit{bimorphic} \cite{BimorphicLenses}, in contrast to
  \textit{simple} lenses whose objects are all of the form $(A, A)$: the forward
  and backward types of simple lenses are the same.} in $\Ca$, $\lens{\Ca}$, is
the category with the following data:
\begin{itemize}
	\item Its objects are pairs $(A,A')$, where both $A$ and $A'$ are objects in $\Ca$
	\item A map from $(A,A')$ to $(B,B')$ consists of a pair $(f,f^*)$ where $f: A \to B$ (called the \textbf{get} or \textbf{forward} part of the lens) and $f^*: A \times B' \to A'$ (called the \textbf{put} or \textbf{backwards} part of the lens)
	\item the composite of $(f,f^*): (A,A') \to (B,B')$ and $(g,g^*): (B,B') \to (C,C')$ is given by get $f;g$ and put
		\[ \<\pi_0,\<\pi_0;f, \pi_1\>;g^*\>;f^* \]
	\item The identity on $(A,A')$ is the pair $(1_A, \pi_1)$.
\end{itemize}
\end{definition}
It is much easier to visualize the morphisms of $\lens{\Ca}$ and their composites with a graphical calculus as described in \cite[Thm. 23]{graphical_optics}.  In this language, a morphism $(f,f^*): (A,A') \to (B,B')$ is written as

\begin{equation*}
  \scalebox{0.7}{\begin{tikzpicture}
	\begin{pgfonlayer}{nodelayer}
		\node [style=none] (20) at (-5, 4) {};
		\node [style=none] (21) at (-5, -4) {};
		\node [style=none] (22) at (5, 4) {};
		\node [style=none] (23) at (5, -4) {};
		\node [style=none] (26) at (-7, 1) {$A$};
		\node [style=none] (27) at (-7, -2) {$A'$};
		\node [style=none] (28) at (7, 2) {$B$};
		\node [style=none] (29) at (7, -2.5) {$B'$};
		\node [style=none] (30) at (-1, 3) {};
		\node [style=none] (31) at (-1, 1) {};
		\node [style=none] (32) at (1, 1) {};
		\node [style=none] (33) at (1, 3) {};
		\node [style=none] (34) at (1, 2) {};
		\node [style=none] (35) at (-1, 2) {};
		\node [style=none] (36) at (-4.25, 1) {};
		\node [style=none] (39) at (-6, 1) {};
		\node [style=none] (40) at (-5, 1) {};
		\node [style=none] (41) at (-3, 2) {};
		\node [style=none] (42) at (-3, 0) {};
		\node [style=none] (43) at (-5, -2) {};
		\node [style=none] (44) at (-6, -2) {};
		\node [style=none] (45) at (5, 2) {};
		\node [style=none] (46) at (6, 2) {};
		\node [style=none] (47) at (3, 0) {};
		\node [style=none] (48) at (3, -1.5) {};
		\node [style=none] (49) at (-1, -1) {};
		\node [style=none] (50) at (-1, -3) {};
		\node [style=none] (51) at (1, -3) {};
		\node [style=none] (52) at (1, -1) {};
		\node [style=none] (53) at (1, -1.5) {};
		\node [style=none] (54) at (-1, -2) {};
		\node [style=none] (55) at (3, -2.5) {};
		\node [style=none] (56) at (1, -2.5) {};
		\node [style=none] (57) at (6, -2.5) {};
		\node [style=none] (58) at (5, -2.5) {};
		\node [style=none] (59) at (0, 2) {$f$};
		\node [style=none] (60) at (0, -2) {$f^*$};
	\end{pgfonlayer}
	\begin{pgfonlayer}{edgelayer}
		\draw (20.center) to (22.center);
		\draw (22.center) to (23.center);
		\draw (23.center) to (21.center);
		\draw (21.center) to (20.center);
		\draw (39.center) to (40.center);
		\draw [style=pointy] (43.center) to (44.center);
		\draw [style=pointy, in=-180, out=0] (36.center) to (41.center);
		\draw [style=pointy, in=180, out=0] (36.center) to (42.center);
		\draw (40.center) to (36.center);
		\draw [style=pointy] (41.center) to (35.center);
		\draw (34.center) to (45.center);
		\draw [style=pointy] (45.center) to (46.center);
		\draw (30.center) to (33.center);
		\draw (33.center) to (34.center);
		\draw (34.center) to (32.center);
		\draw (32.center) to (31.center);
		\draw (31.center) to (35.center);
		\draw (35.center) to (30.center);
		\draw [style=pointy] (42.center) to (47.center);
		\draw [style=pointy, bend left=90, looseness=1.75] (47.center) to (48.center);
		\draw (49.center) to (52.center);
		\draw (52.center) to (53.center);
		\draw (53.center) to (51.center);
		\draw (51.center) to (50.center);
		\draw (50.center) to (54.center);
		\draw (54.center) to (49.center);
		\draw [style=pointy] (48.center) to (53.center);
		\draw [style=pointy] (55.center) to (56.center);
		\draw (54.center) to (43.center);
		\draw (57.center) to (58.center);
		\draw (58.center) to (55.center);
	\end{pgfonlayer}
\end{tikzpicture}}
\end{equation*}

where $\scalebox{0.5}{\begin{tikzpicture}
	\begin{pgfonlayer}{nodelayer}
		\node [style=none] (0) at (0.75, -0.5) {};
		\node [style=none] (1) at (-0.25, 0.25) {};
		\node [style=none] (2) at (0.75, 1) {};
		\node [style=none] (3) at (-0.75, 0.25) {};
	\end{pgfonlayer}
	\begin{pgfonlayer}{edgelayer}
		\draw [in=-180, out=0, looseness=0.75] (1.center) to (0.center);
		\draw [in=-180, out=0, looseness=0.75] (1.center) to (2.center);
		\draw (1.center) to (3.center);
	\end{pgfonlayer}
\end{tikzpicture}}$ is the string diagram (`built-in' in any cartesian category) duplicating the value $A$. 
It is clear in this language how to describe the composite of $(f,f^*): (A,A') \to (B,B')$ and $(g,g^*): (B,B') \to (C,C')$: simply join the $B/B'$ wires together to get the composite lens

\begin{equation*}\label{eq:lens-composition}
  \scalebox{0.7}{\begin{tikzpicture}
	\begin{pgfonlayer}{nodelayer}
		\node [style=none] (20) at (-5, 4) {};
		\node [style=none] (21) at (-5, -4) {};
		\node [style=none] (22) at (5, 4) {};
		\node [style=none] (23) at (5, -4) {};
		\node [style=none] (26) at (-7, 1) {$A$};
		\node [style=none] (27) at (-7, -2) {$A'$};
		\node [style=none] (30) at (-1, 3) {};
		\node [style=none] (31) at (-1, 1) {};
		\node [style=none] (32) at (1, 1) {};
		\node [style=none] (33) at (1, 3) {};
		\node [style=none] (34) at (1, 2) {};
		\node [style=none] (35) at (-1, 2) {};
		\node [style=none] (36) at (-4.25, 1) {};
		\node [style=none] (39) at (-6, 1) {};
		\node [style=none] (40) at (-5, 1) {};
		\node [style=none] (41) at (-3, 2) {};
		\node [style=none] (42) at (-3, 0) {};
		\node [style=none] (43) at (-5, -2) {};
		\node [style=none] (44) at (-6, -2) {};
		\node [style=none] (45) at (5, 2) {};
		\node [style=none] (46) at (6, 2) {};
		\node [style=none] (47) at (3, 0) {};
		\node [style=none] (48) at (3, -1.5) {};
		\node [style=none] (49) at (-1, -1) {};
		\node [style=none] (50) at (-1, -3) {};
		\node [style=none] (51) at (1, -3) {};
		\node [style=none] (52) at (1, -1) {};
		\node [style=none] (53) at (1, -1.5) {};
		\node [style=none] (54) at (-1, -2) {};
		\node [style=none] (55) at (3, -2.5) {};
		\node [style=none] (56) at (1, -2.5) {};
		\node [style=none] (57) at (6, -2.5) {};
		\node [style=none] (58) at (5, -2.5) {};
		\node [style=none] (59) at (0, 2) {$f$};
		\node [style=none] (60) at (0, -2) {$f^*$};
		\node [style=none] (61) at (9, 4) {};
		\node [style=none] (62) at (9, -4) {};
		\node [style=none] (63) at (19, 4) {};
		\node [style=none] (64) at (19, -4) {};
		\node [style=none] (67) at (21, 2) {$C$};
		\node [style=none] (68) at (21, -2.5) {$C'$};
		\node [style=none] (69) at (13, 3) {};
		\node [style=none] (70) at (13, 1) {};
		\node [style=none] (71) at (15, 1) {};
		\node [style=none] (72) at (15, 3) {};
		\node [style=none] (73) at (15, 2) {};
		\node [style=none] (74) at (13, 2) {};
		\node [style=none] (75) at (9.75, 1) {};
		\node [style=none] (76) at (8, 1) {};
		\node [style=none] (77) at (9, 1) {};
		\node [style=none] (78) at (11, 2) {};
		\node [style=none] (79) at (11, 0) {};
		\node [style=none] (80) at (9, -2) {};
		\node [style=none] (81) at (8, -2) {};
		\node [style=none] (82) at (19, 2) {};
		\node [style=none] (83) at (20, 2) {};
		\node [style=none] (84) at (17, 0) {};
		\node [style=none] (85) at (17, -1.5) {};
		\node [style=none] (86) at (13, -1) {};
		\node [style=none] (87) at (13, -3) {};
		\node [style=none] (88) at (15, -3) {};
		\node [style=none] (89) at (15, -1) {};
		\node [style=none] (90) at (15, -1.5) {};
		\node [style=none] (91) at (13, -2) {};
		\node [style=none] (92) at (17, -2.5) {};
		\node [style=none] (93) at (15, -2.5) {};
		\node [style=none] (94) at (20, -2.5) {};
		\node [style=none] (95) at (19, -2.5) {};
		\node [style=none] (96) at (14, 2) {$g$};
		\node [style=none] (97) at (14, -2) {$g^*$};
		\node [style=none] (98) at (7.25, 2) {$B$};
		\node [style=none] (99) at (7.25, -2.75) {$B'$};
	\end{pgfonlayer}
	\begin{pgfonlayer}{edgelayer}
		\draw (20.center) to (22.center);
		\draw (22.center) to (23.center);
		\draw (23.center) to (21.center);
		\draw (21.center) to (20.center);
		\draw (39.center) to (40.center);
		\draw [style=pointy] (43.center) to (44.center);
		\draw [style=pointy, in=-180, out=0] (36.center) to (41.center);
		\draw [style=pointy, in=180, out=0] (36.center) to (42.center);
		\draw (40.center) to (36.center);
		\draw [style=pointy] (41.center) to (35.center);
		\draw (34.center) to (45.center);
		\draw [style=pointy] (45.center) to (46.center);
		\draw (30.center) to (33.center);
		\draw (33.center) to (34.center);
		\draw (34.center) to (32.center);
		\draw (32.center) to (31.center);
		\draw (31.center) to (35.center);
		\draw (35.center) to (30.center);
		\draw [style=pointy] (42.center) to (47.center);
		\draw [style=pointy, bend left=90, looseness=1.75] (47.center) to (48.center);
		\draw (49.center) to (52.center);
		\draw (52.center) to (53.center);
		\draw (53.center) to (51.center);
		\draw (51.center) to (50.center);
		\draw (50.center) to (54.center);
		\draw (54.center) to (49.center);
		\draw [style=pointy] (48.center) to (53.center);
		\draw [style=pointy] (55.center) to (56.center);
		\draw (54.center) to (43.center);
		\draw (57.center) to (58.center);
		\draw (58.center) to (55.center);
		\draw (61.center) to (63.center);
		\draw (63.center) to (64.center);
		\draw (64.center) to (62.center);
		\draw (62.center) to (61.center);
		\draw (76.center) to (77.center);
		\draw [style=pointy] (80.center) to (81.center);
		\draw [style=pointy, in=-180, out=0] (75.center) to (78.center);
		\draw [style=pointy, in=180, out=0] (75.center) to (79.center);
		\draw (77.center) to (75.center);
		\draw [style=pointy] (78.center) to (74.center);
		\draw (73.center) to (82.center);
		\draw [style=pointy] (82.center) to (83.center);
		\draw (69.center) to (72.center);
		\draw (72.center) to (73.center);
		\draw (73.center) to (71.center);
		\draw (71.center) to (70.center);
		\draw (70.center) to (74.center);
		\draw (74.center) to (69.center);
		\draw [style=pointy] (79.center) to (84.center);
		\draw [style=pointy, bend left=90, looseness=1.75] (84.center) to (85.center);
		\draw (86.center) to (89.center);
		\draw (89.center) to (90.center);
		\draw (90.center) to (88.center);
		\draw (88.center) to (87.center);
		\draw (87.center) to (91.center);
		\draw (91.center) to (86.center);
		\draw [style=pointy] (85.center) to (90.center);
		\draw [style=pointy] (92.center) to (93.center);
		\draw (91.center) to (80.center);
		\draw (94.center) to (95.center);
		\draw (95.center) to (92.center);
		\draw [in=180, out=0] (46.center) to (76.center);
		\draw [in=-180, out=0] (57.center) to (81.center);
	\end{pgfonlayer}
\end{tikzpicture}}
\end{equation*}

and the formula for the composite in terms of equations (as described above) follows from this.  \\

We will often write lenses without the inside wires exposed, thinking of the entire lens as a black-box:

\begin{equation*}
   \scalebox{0.8}{\begin{tikzpicture}
	\begin{pgfonlayer}{nodelayer}
		\node [style=none] (20) at (-1, 1) {};
		\node [style=none] (21) at (-1, -1) {};
		\node [style=none] (22) at (1, 1) {};
		\node [style=none] (23) at (1, -1) {};
		\node [style=none] (26) at (-3, 0.5) {$A$};
		\node [style=none] (27) at (-3, -0.5) {$A'$};
		\node [style=none] (28) at (3, 0.5) {$B$};
		\node [style=none] (29) at (3, -0.5) {$B'$};
		\node [style=none] (39) at (-2, 0.5) {};
		\node [style=none] (40) at (-1, 0.5) {};
		\node [style=none] (43) at (-1, -0.5) {};
		\node [style=none] (44) at (-2, -0.5) {};
		\node [style=none] (45) at (1, 0.5) {};
		\node [style=none] (46) at (2, 0.5) {};
		\node [style=none] (57) at (2, -0.5) {};
		\node [style=none] (58) at (1, -0.5) {};
		\node [style=none] (59) at (0, 0) {$(f, f^*)$};
	\end{pgfonlayer}
	\begin{pgfonlayer}{edgelayer}
		\draw (20.center) to (22.center);
		\draw (22.center) to (23.center);
		\draw (23.center) to (21.center);
		\draw (21.center) to (20.center);
		\draw [style=pointy] (39.center) to (40.center);
		\draw [style=pointy] (43.center) to (44.center);
		\draw [style=pointy] (45.center) to (46.center);
		\draw [style=pointy] (57.center) to (58.center);
	\end{pgfonlayer}
\end{tikzpicture}}
\end{equation*}

Note $\lens{\Ca}$ is a monoidal category, with $(A,A') \otimes (B,B')$ defined as $(A \times B, A' \times B')$.
 However, in general $\lens{\Ca}$ is not itself Cartesian.  This is easy to see when looking at even a terminal object: if $T$ is a terminal object in $\Ca$, then in general $(T,T)$ will not be a terminal object in $\lens{\Ca}$ ---  it if was, there would be a unique lens 
	$(!_A,!_A^*): (A,A') \to (T,T)$ whose put part would need to be a (unique) map $A \times T \to A'$, but in general there are many such maps.  

\subsection{Parametric Lenses}\label{sec:paralenses}

The fundamental category where supervised learning takes place is the composite
of the two constructions in the previous sections.  As noted in the previous
section, for a Cartesian category $\Ca$, $\lens{\Ca}$ is monoidal, and so we can
form the the category $\Para{\lens{\Ca}}$, which we shall call the category of
\textbf{parametric lenses} of $\Ca$.  The definition of this category follows
automatically from definitions of $\para$ and $\lens{\Ca}$:

\begin{definition} The category $\Para{\lens{\Ca}}$ of parametric lenses on $\Ca$ is defined as follows.
\begin{itemize}
	\item An object is a pair of objects $(A,A')$ from $\Ca$ 
	\item A morphism from $(A,A')$ to $(B,B')$, called a parametric
    lens\footnote{In \cite{BackpropAsFunctor}, these are called
      \textit{learners}. However, in this paper we study them in a much broader
      light; see section \ref{section:related-work}.}, is a choice of parameter pair $(P,P')$ and a lens
		\[ (f,f^*): (A,A') \times (P,P') \to (B,B') \]
	so that $f: P \times A \to B$ and $f^*: P \times A \times B' \to P' \times A'$
\end{itemize}
\end{definition}

By the previous two sections, we get a graphical language for $\Para{\lens{\Ca}}$ which uses the graphical language for $\lens{\Ca}$ from section \ref{sec:lenses} as a base, then augments it with parameters as described in section \ref{sec:para}.  Thus a morphism of $\Para{\lens{\Ca}}$ from $(A,A')$ to $(B,B')$ is a box with input/output of $(A,A')$ on the left, input/output of $(B,B')$ on the right, and input/output of $(P,P')$ (the parameter space) on top:
\begin{equation}\label{eq:paralens-morphism}
   \scalebox{0.8}{\begin{tikzpicture}
	\begin{pgfonlayer}{nodelayer}
		\node [style=none] (20) at (-1, 1) {};
		\node [style=none] (21) at (-1, -1) {};
		\node [style=none] (22) at (1, 1) {};
		\node [style=none] (23) at (1, -1) {};
		\node [style=none] (26) at (-2.5, 0.5) {$A$};
		\node [style=none] (27) at (-2.5, -0.5) {$A'$};
		\node [style=none] (28) at (2.5, 0.5) {$B$};
		\node [style=none] (29) at (2.5, -0.5) {$B'$};
		\node [style=none] (39) at (-2, 0.5) {};
		\node [style=none] (40) at (-1, 0.5) {};
		\node [style=none] (43) at (-1, -0.5) {};
		\node [style=none] (44) at (-2, -0.5) {};
		\node [style=none] (45) at (1, 0.5) {};
		\node [style=none] (46) at (2, 0.5) {};
		\node [style=none] (57) at (2, -0.5) {};
		\node [style=none] (58) at (1, -0.5) {};
		\node [style=none] (60) at (-0.5, 1) {};
		\node [style=none] (61) at (0.5, 1) {};
		\node [style=none] (62) at (-0.5, 2) {};
		\node [style=none] (63) at (0.5, 2) {};
		\node [style=none] (64) at (-0.5, 2.5) {$P$};
		\node [style=none] (65) at (0.5, 2.5) {$P'$};
	\end{pgfonlayer}
	\begin{pgfonlayer}{edgelayer}
		\draw (20.center) to (22.center);
		\draw (22.center) to (23.center);
		\draw (23.center) to (21.center);
		\draw (21.center) to (20.center);
		\draw [style=pointy] (39.center) to (40.center);
		\draw [style=pointy] (43.center) to (44.center);
		\draw [style=pointy] (45.center) to (46.center);
		\draw [style=pointy] (57.center) to (58.center);
		\draw [style=pointy] (62.center) to (60.center);
		\draw [style=pointy] (61.center) to (63.center);
	\end{pgfonlayer}
\end{tikzpicture}}
\end{equation}

Composition is again quite natural in this formulation: given one box with input/output wires $(B,B')$ on the right and another box with input/output wires $(B,B')$ on the left, one simply hooks up those input/output wires to get the desired composite:
\begin{equation}\label{eq:paralens-composition}
  \scalebox{0.8}{ \begin{tikzpicture}
	\begin{pgfonlayer}{nodelayer}
		\node [style=none] (20) at (-3, 1) {};
		\node [style=none] (21) at (-3, -1) {};
		\node [style=none] (22) at (-1, 1) {};
		\node [style=none] (23) at (-1, -1) {};
		\node [style=none] (26) at (-4.5, 0.5) {$A$};
		\node [style=none] (27) at (-4.5, -0.5) {$A'$};
		\node [style=none] (28) at (0, 1) {$B$};
		\node [style=none] (29) at (0, -1) {$B'$};
		\node [style=none] (39) at (-4, 0.5) {};
		\node [style=none] (40) at (-3, 0.5) {};
		\node [style=none] (43) at (-3, -0.5) {};
		\node [style=none] (44) at (-4, -0.5) {};
		\node [style=none] (45) at (-1, 0.5) {};
		\node [style=none] (46) at (0, 0.5) {};
		\node [style=none] (57) at (0, -0.5) {};
		\node [style=none] (58) at (-1, -0.5) {};
		\node [style=none] (60) at (-2.5, 1) {};
		\node [style=none] (61) at (-1.5, 1) {};
		\node [style=none] (62) at (-2.5, 2) {};
		\node [style=none] (63) at (-1.5, 2) {};
		\node [style=none] (64) at (-2.5, 2.5) {$P$};
		\node [style=none] (65) at (-1.5, 2.5) {$P'$};
		\node [style=none] (66) at (1, 1) {};
		\node [style=none] (67) at (1, -1) {};
		\node [style=none] (68) at (3, 1) {};
		\node [style=none] (69) at (3, -1) {};
		\node [style=none] (72) at (4.5, 0.5) {$C$};
		\node [style=none] (73) at (4.5, -0.5) {$C'$};
		\node [style=none] (74) at (0, 0.5) {};
		\node [style=none] (75) at (1, 0.5) {};
		\node [style=none] (76) at (1, -0.5) {};
		\node [style=none] (77) at (0, -0.5) {};
		\node [style=none] (78) at (3, 0.5) {};
		\node [style=none] (79) at (4, 0.5) {};
		\node [style=none] (80) at (4, -0.5) {};
		\node [style=none] (81) at (3, -0.5) {};
		\node [style=none] (82) at (1.5, 1) {};
		\node [style=none] (83) at (2.5, 1) {};
		\node [style=none] (84) at (1.5, 2) {};
		\node [style=none] (85) at (2.5, 2) {};
		\node [style=none] (86) at (1.5, 2.5) {$Q$};
		\node [style=none] (87) at (2.5, 2.5) {$Q'$};
	\end{pgfonlayer}
	\begin{pgfonlayer}{edgelayer}
		\draw (20.center) to (22.center);
		\draw (22.center) to (23.center);
		\draw (23.center) to (21.center);
		\draw (21.center) to (20.center);
		\draw [style=pointy] (39.center) to (40.center);
		\draw [style=pointy] (43.center) to (44.center);
		\draw (45.center) to (46.center);
		\draw [style=pointy] (57.center) to (58.center);
		\draw [style=pointy] (62.center) to (60.center);
		\draw [style=pointy] (61.center) to (63.center);
		\draw (66.center) to (68.center);
		\draw (68.center) to (69.center);
		\draw (69.center) to (67.center);
		\draw (67.center) to (66.center);
		\draw [style=pointy] (74.center) to (75.center);
		\draw (76.center) to (77.center);
		\draw [style=pointy] (78.center) to (79.center);
		\draw [style=pointy] (80.center) to (81.center);
		\draw [style=pointy] (84.center) to (82.center);
		\draw [style=pointy] (83.center) to (85.center);
	\end{pgfonlayer}
\end{tikzpicture}}
\end{equation}

A reparameterisation in $\Para{\lens{\Ca}}$ is depicted graphically as drawing a box on top of the $(P,P')$ wires:
\begin{equation}\label{eq:paralens-reparametrisation}
  \scalebox{0.8}{ \begin{tikzpicture}
	\begin{pgfonlayer}{nodelayer}
		\node [style=none] (20) at (-1, 1) {};
		\node [style=none] (21) at (-1, -1) {};
		\node [style=none] (22) at (1, 1) {};
		\node [style=none] (23) at (1, -1) {};
		\node [style=none] (26) at (-2.5, 0.5) {$A$};
		\node [style=none] (27) at (-2.5, -0.5) {$A'$};
		\node [style=none] (28) at (2.5, 0.5) {$B$};
		\node [style=none] (29) at (2.5, -0.5) {$B'$};
		\node [style=none] (39) at (-2, 0.5) {};
		\node [style=none] (40) at (-1, 0.5) {};
		\node [style=none] (43) at (-1, -0.5) {};
		\node [style=none] (44) at (-2, -0.5) {};
		\node [style=none] (45) at (1, 0.5) {};
		\node [style=none] (46) at (2, 0.5) {};
		\node [style=none] (57) at (2, -0.5) {};
		\node [style=none] (58) at (1, -0.5) {};
		\node [style=none] (60) at (-0.5, 1) {};
		\node [style=none] (61) at (0.5, 1) {};
		\node [style=none] (62) at (-0.5, 3) {};
		\node [style=none] (63) at (0.5, 3) {};
		\node [style=none] (64) at (-1, 2) {$P$};
		\node [style=none] (65) at (1, 2) {$P'$};
		\node [style=none] (66) at (-1, 5) {};
		\node [style=none] (67) at (-1, 3) {};
		\node [style=none] (68) at (1, 5) {};
		\node [style=none] (69) at (1, 3) {};
		\node [style=none] (74) at (-0.5, 5) {};
		\node [style=none] (75) at (0.5, 5) {};
		\node [style=none] (76) at (-0.5, 6) {};
		\node [style=none] (77) at (0.5, 6) {};
		\node [style=none] (78) at (-0.5, 6.5) {$Q$};
		\node [style=none] (79) at (0.5, 6.5) {$Q'$};
	\end{pgfonlayer}
	\begin{pgfonlayer}{edgelayer}
		\draw (20.center) to (22.center);
		\draw (22.center) to (23.center);
		\draw (23.center) to (21.center);
		\draw (21.center) to (20.center);
		\draw [style=pointy] (39.center) to (40.center);
		\draw [style=pointy] (43.center) to (44.center);
		\draw [style=pointy] (45.center) to (46.center);
		\draw [style=pointy] (57.center) to (58.center);
		\draw [style=pointy] (62.center) to (60.center);
		\draw [style=pointy] (61.center) to (63.center);
		\draw (66.center) to (68.center);
		\draw (68.center) to (69.center);
		\draw (69.center) to (67.center);
		\draw (67.center) to (66.center);
		\draw [style=pointy] (76.center) to (74.center);
		\draw [style=pointy] (75.center) to (77.center);
	\end{pgfonlayer}
\end{tikzpicture}}
\end{equation}

Given a generic morphism $f$ in $\Para{\lens{\Ca}}$ as depicted in \eqref{eq:paralens-morphism}, one can see how it is possible to  ``learn'' new values from $f$: it takes as input an input $A$, a parameter $P$, and a change $B'$, and outputs a change in $A$, a value of $B$, and a change $P'$.  This last element is the key component for supervised learning: intuitively, it says how to change the parameter values to get the neural network closer to the true value of the desired function.

The question, then, is how one is to define such a parametric lens given nothing more than a neural network, ie., a parameterized map $(P,f): A \to B$.  This is precisely what the gradient operation provides, and its generality to categories is explored in the next subsection.

\subsection{Cartesian Reverse Differential Categories}\label{sec:crdc}

Fundamental to all gradient-based learning is, of course, the gradient
operation.
In most cases this gradient operation is performed in the category of smooth
maps between Euclidean spaces.
However, recent work \cite{rda} has shown that gradient-based learning can also
work well in other categories; for example, in a category of boolean circuits.
Thus, to encompass these examples in a single framework, it is helpful to work
in a category with an abstract gradient operation.
Specifically, we will work in a Cartesian reverse differential category (first
defined in \cite{CRDC}), a category in which every map has an associated reverse
derivative.




\begin{definition}\label{defn:CRDC} ~
\begin{itemize}
  \item \cite[Defn. 1]{CRDC} A \textbf{Cartesian left additive category}
    consists of a category $\Ca$ with chosen finite products (including a
    terminal object), and an addition operation and zero morphism in each
    homset, satisfying various axioms.
  \item \cite[Defn. 13]{CRDC} A \textbf{Cartesian reverse differential category} (CRDC) consists of a Cartesian left additive category $\Ca$, together with an operation which provides, for each map $f: A \to B \mbox{ in } \Ca$, a map
    \[ R[f]: A \times B \to A \]
    satisfying seven axioms (for full details, see the appendix).  
\end{itemize}
\end{definition}

Why are reverse derivatives helpful for learning?  For $f: A \to B$, the pair $(f,R[f])$ form a lens from $(A,A)$ to $(B,B)$, with $R[f]$ acting as backwards map. Thus having a reverse derivative already provides a way to turn an ordinary map in the category into which one can pass information backwards, that is, a map which can ``learn''.  

Note assigning type $A \times B \to A$ to $R[f]$ hides some relevant information: $B$-values in the domain and $A$-values in the codomain of $R[f]$ do not play the same role as values of the same types in $f \colon A \to B$: in $R[f]$, they really take in a tangent vector at $B$ and output a tangent vector at $A$ (\emph{cf.} the definition of $R[f]$ in $\Smooth$, Example~\ref{ex:smoothrdc} below).  To emphasise this, we will type $R[f]$ as a map $A \times B' \to A'$ (even though in reality $A = A'$ and $B=B'$), thus meaning that $(f, R[f])$ is actually a lens from $(A,A')$ to $(B,B')$.  This typing distinction will be helpful later on, when we want to add additional components to our learning algorithms. 

Graphically, then, we represent the pair $(f,R[f])$ as a lens: 
\begin{equation*}
  \scalebox{0.7}{\begin{tikzpicture}
	\begin{pgfonlayer}{nodelayer}
		\node [style=none] (20) at (-5, 4) {};
		\node [style=none] (21) at (-5, -4) {};
		\node [style=none] (22) at (5, 4) {};
		\node [style=none] (23) at (5, -4) {};
		\node [style=none] (26) at (-7, 1) {$A$};
		\node [style=none] (27) at (-7, -2) {$A'$};
		\node [style=none] (28) at (7, 2) {$B$};
		\node [style=none] (29) at (7, -2.5) {$B'$};
		\node [style=none] (30) at (-1, 3) {};
		\node [style=none] (31) at (-1, 1) {};
		\node [style=none] (32) at (1, 1) {};
		\node [style=none] (33) at (1, 3) {};
		\node [style=none] (34) at (1, 2) {};
		\node [style=none] (35) at (-1, 2) {};
		\node [style=none] (36) at (-4.25, 1) {};
		\node [style=none] (39) at (-6, 1) {};
		\node [style=none] (40) at (-5, 1) {};
		\node [style=none] (41) at (-3, 2) {};
		\node [style=none] (42) at (-3, 0) {};
		\node [style=none] (43) at (-5, -2) {};
		\node [style=none] (44) at (-6, -2) {};
		\node [style=none] (45) at (5, 2) {};
		\node [style=none] (46) at (6, 2) {};
		\node [style=none] (47) at (3, 0) {};
		\node [style=none] (48) at (3, -1.5) {};
		\node [style=none] (49) at (-1, -1) {};
		\node [style=none] (50) at (-1, -3) {};
		\node [style=none] (51) at (1, -3) {};
		\node [style=none] (52) at (1, -1) {};
		\node [style=none] (53) at (1, -1.5) {};
		\node [style=none] (54) at (-1, -2) {};
		\node [style=none] (55) at (3, -2.5) {};
		\node [style=none] (56) at (1, -2.5) {};
		\node [style=none] (57) at (6, -2.5) {};
		\node [style=none] (58) at (5, -2.5) {};
		\node [style=none] (59) at (0, 2) {$f$};
		\node [style=none] (60) at (0, -2) {$R[f]$};
	\end{pgfonlayer}
	\begin{pgfonlayer}{edgelayer}
		\draw (20.center) to (22.center);
		\draw (22.center) to (23.center);
		\draw (23.center) to (21.center);
		\draw (21.center) to (20.center);
		\draw (39.center) to (40.center);
		\draw [style=pointy] (43.center) to (44.center);
		\draw [style=pointy, in=-180, out=0] (36.center) to (41.center);
		\draw [style=pointy, in=180, out=0] (36.center) to (42.center);
		\draw (40.center) to (36.center);
		\draw [style=pointy] (41.center) to (35.center);
		\draw (34.center) to (45.center);
		\draw [style=pointy] (45.center) to (46.center);
		\draw (30.center) to (33.center);
		\draw (33.center) to (34.center);
		\draw (34.center) to (32.center);
		\draw (32.center) to (31.center);
		\draw (31.center) to (35.center);
		\draw (35.center) to (30.center);
		\draw [style=pointy] (42.center) to (47.center);
		\draw [style=pointy, bend left=90, looseness=1.75] (47.center) to (48.center);
		\draw (49.center) to (52.center);
		\draw (52.center) to (53.center);
		\draw (53.center) to (51.center);
		\draw (51.center) to (50.center);
		\draw (50.center) to (54.center);
		\draw (54.center) to (49.center);
		\draw [style=pointy] (48.center) to (53.center);
		\draw [style=pointy] (55.center) to (56.center);
		\draw (54.center) to (43.center);
		\draw (57.center) to (58.center);
		\draw (58.center) to (55.center);
	\end{pgfonlayer}
\end{tikzpicture}}
\end{equation*}

This point of view also makes clear the usefulness of the reverse chain rule (axiom [RD.5] from \cite[Defn. 13]{CRDC}): it tells us that the operation which takes a map $f$ and produces the lens $(f,R[f])$ preserves composition: that is,

\begin{equation*}
  \scalebox{0.55}{\begin{tikzpicture}
	\begin{pgfonlayer}{nodelayer}
		\node [style=none] (20) at (-5, 4) {};
		\node [style=none] (21) at (-5, -4) {};
		\node [style=none] (22) at (5, 4) {};
		\node [style=none] (23) at (5, -4) {};
		\node [style=none] (26) at (-7, 1) {$A$};
		\node [style=none] (27) at (-7, -2) {$A'$};
		\node [style=none] (28) at (7, 2) {$C$};
		\node [style=none] (29) at (7, -2.5) {$C'$};
		\node [style=none] (30) at (-1, 3) {};
		\node [style=none] (31) at (-1, 1) {};
		\node [style=none] (32) at (1, 1) {};
		\node [style=none] (33) at (1, 3) {};
		\node [style=none] (34) at (1, 2) {};
		\node [style=none] (35) at (-1, 2) {};
		\node [style=none] (36) at (-4.25, 1) {};
		\node [style=none] (39) at (-6, 1) {};
		\node [style=none] (40) at (-5, 1) {};
		\node [style=none] (41) at (-3, 2) {};
		\node [style=none] (42) at (-3, 0) {};
		\node [style=none] (43) at (-5, -2) {};
		\node [style=none] (44) at (-6, -2) {};
		\node [style=none] (45) at (5, 2) {};
		\node [style=none] (46) at (6, 2) {};
		\node [style=none] (47) at (3, 0) {};
		\node [style=none] (48) at (3, -1.5) {};
		\node [style=none] (49) at (-1, -1) {};
		\node [style=none] (50) at (-1, -3) {};
		\node [style=none] (51) at (1, -3) {};
		\node [style=none] (52) at (1, -1) {};
		\node [style=none] (53) at (1, -1.5) {};
		\node [style=none] (54) at (-1, -2) {};
		\node [style=none] (55) at (3, -2.5) {};
		\node [style=none] (56) at (1, -2.5) {};
		\node [style=none] (57) at (6, -2.5) {};
		\node [style=none] (58) at (5, -2.5) {};
		\node [style=none] (59) at (0, 2) {$fg$};
		\node [style=none] (60) at (0, -2) {$R[fg]$};
	\end{pgfonlayer}
	\begin{pgfonlayer}{edgelayer}
		\draw (20.center) to (22.center);
		\draw (22.center) to (23.center);
		\draw (23.center) to (21.center);
		\draw (21.center) to (20.center);
		\draw (39.center) to (40.center);
		\draw [style=pointy] (43.center) to (44.center);
		\draw [style=pointy, in=-180, out=0] (36.center) to (41.center);
		\draw [style=pointy, in=180, out=0] (36.center) to (42.center);
		\draw (40.center) to (36.center);
		\draw [style=pointy] (41.center) to (35.center);
		\draw (34.center) to (45.center);
		\draw [style=pointy] (45.center) to (46.center);
		\draw (30.center) to (33.center);
		\draw (33.center) to (34.center);
		\draw (34.center) to (32.center);
		\draw (32.center) to (31.center);
		\draw (31.center) to (35.center);
		\draw (35.center) to (30.center);
		\draw [style=pointy] (42.center) to (47.center);
		\draw [style=pointy, bend left=90, looseness=1.75] (47.center) to (48.center);
		\draw (49.center) to (52.center);
		\draw (52.center) to (53.center);
		\draw (53.center) to (51.center);
		\draw (51.center) to (50.center);
		\draw (50.center) to (54.center);
		\draw (54.center) to (49.center);
		\draw [style=pointy] (48.center) to (53.center);
		\draw [style=pointy] (55.center) to (56.center);
		\draw (54.center) to (43.center);
		\draw (57.center) to (58.center);
		\draw (58.center) to (55.center);
	\end{pgfonlayer}
\end{tikzpicture}}
  \qquad = \qquad
  \scalebox{0.55}{\begin{tikzpicture}
	\begin{pgfonlayer}{nodelayer}
		\node [style=none] (20) at (-5, 4) {};
		\node [style=none] (21) at (-5, -4) {};
		\node [style=none] (22) at (5, 4) {};
		\node [style=none] (23) at (5, -4) {};
		\node [style=none] (26) at (-7, 1) {$A$};
		\node [style=none] (27) at (-7, -2) {$A'$};
		\node [style=none] (30) at (-1, 3) {};
		\node [style=none] (31) at (-1, 1) {};
		\node [style=none] (32) at (1, 1) {};
		\node [style=none] (33) at (1, 3) {};
		\node [style=none] (34) at (1, 2) {};
		\node [style=none] (35) at (-1, 2) {};
		\node [style=none] (36) at (-4.25, 1) {};
		\node [style=none] (39) at (-6, 1) {};
		\node [style=none] (40) at (-5, 1) {};
		\node [style=none] (41) at (-3, 2) {};
		\node [style=none] (42) at (-3, 0) {};
		\node [style=none] (43) at (-5, -2) {};
		\node [style=none] (44) at (-6, -2) {};
		\node [style=none] (45) at (5, 2) {};
		\node [style=none] (46) at (6, 2) {};
		\node [style=none] (47) at (3, 0) {};
		\node [style=none] (48) at (3, -1.5) {};
		\node [style=none] (49) at (-1, -1) {};
		\node [style=none] (50) at (-1, -3) {};
		\node [style=none] (51) at (1, -3) {};
		\node [style=none] (52) at (1, -1) {};
		\node [style=none] (53) at (1, -1.5) {};
		\node [style=none] (54) at (-1, -2) {};
		\node [style=none] (55) at (3, -2.5) {};
		\node [style=none] (56) at (1, -2.5) {};
		\node [style=none] (57) at (6, -2.5) {};
		\node [style=none] (58) at (5, -2.5) {};
		\node [style=none] (59) at (0, 2) {$f$};
		\node [style=none] (60) at (0, -2) {$R[f]$};
		\node [style=none] (61) at (9, 4) {};
		\node [style=none] (62) at (9, -4) {};
		\node [style=none] (63) at (19, 4) {};
		\node [style=none] (64) at (19, -4) {};
		\node [style=none] (67) at (21, 2) {$C$};
		\node [style=none] (68) at (21, -2.5) {$C'$};
		\node [style=none] (69) at (13, 3) {};
		\node [style=none] (70) at (13, 1) {};
		\node [style=none] (71) at (15, 1) {};
		\node [style=none] (72) at (15, 3) {};
		\node [style=none] (73) at (15, 2) {};
		\node [style=none] (74) at (13, 2) {};
		\node [style=none] (75) at (9.75, 1) {};
		\node [style=none] (76) at (8, 1) {};
		\node [style=none] (77) at (9, 1) {};
		\node [style=none] (78) at (11, 2) {};
		\node [style=none] (79) at (11, 0) {};
		\node [style=none] (80) at (9, -2) {};
		\node [style=none] (81) at (8, -2) {};
		\node [style=none] (82) at (19, 2) {};
		\node [style=none] (83) at (20, 2) {};
		\node [style=none] (84) at (17, 0) {};
		\node [style=none] (85) at (17, -1.5) {};
		\node [style=none] (86) at (13, -1) {};
		\node [style=none] (87) at (13, -3) {};
		\node [style=none] (88) at (15, -3) {};
		\node [style=none] (89) at (15, -1) {};
		\node [style=none] (90) at (15, -1.5) {};
		\node [style=none] (91) at (13, -2) {};
		\node [style=none] (92) at (17, -2.5) {};
		\node [style=none] (93) at (15, -2.5) {};
		\node [style=none] (94) at (20, -2.5) {};
		\node [style=none] (95) at (19, -2.5) {};
		\node [style=none] (96) at (14, 2) {$g$};
		\node [style=none] (97) at (14, -2) {$R[g]$};
		\node [style=none] (98) at (7.25, 2) {$B$};
		\node [style=none] (99) at (7.25, -2.75) {$B'$};
	\end{pgfonlayer}
	\begin{pgfonlayer}{edgelayer}
		\draw (20.center) to (22.center);
		\draw (22.center) to (23.center);
		\draw (23.center) to (21.center);
		\draw (21.center) to (20.center);
		\draw (39.center) to (40.center);
		\draw [style=pointy] (43.center) to (44.center);
		\draw [style=pointy, in=-180, out=0] (36.center) to (41.center);
		\draw [style=pointy, in=180, out=0] (36.center) to (42.center);
		\draw (40.center) to (36.center);
		\draw [style=pointy] (41.center) to (35.center);
		\draw (34.center) to (45.center);
		\draw [style=pointy] (45.center) to (46.center);
		\draw (30.center) to (33.center);
		\draw (33.center) to (34.center);
		\draw (34.center) to (32.center);
		\draw (32.center) to (31.center);
		\draw (31.center) to (35.center);
		\draw (35.center) to (30.center);
		\draw [style=pointy] (42.center) to (47.center);
		\draw [style=pointy, bend left=90, looseness=1.75] (47.center) to (48.center);
		\draw (49.center) to (52.center);
		\draw (52.center) to (53.center);
		\draw (53.center) to (51.center);
		\draw (51.center) to (50.center);
		\draw (50.center) to (54.center);
		\draw (54.center) to (49.center);
		\draw [style=pointy] (48.center) to (53.center);
		\draw [style=pointy] (55.center) to (56.center);
		\draw (54.center) to (43.center);
		\draw (57.center) to (58.center);
		\draw (58.center) to (55.center);
		\draw (61.center) to (63.center);
		\draw (63.center) to (64.center);
		\draw (64.center) to (62.center);
		\draw (62.center) to (61.center);
		\draw (76.center) to (77.center);
		\draw [style=pointy] (80.center) to (81.center);
		\draw [style=pointy, in=-180, out=0] (75.center) to (78.center);
		\draw [style=pointy, in=180, out=0] (75.center) to (79.center);
		\draw (77.center) to (75.center);
		\draw [style=pointy] (78.center) to (74.center);
		\draw (73.center) to (82.center);
		\draw [style=pointy] (82.center) to (83.center);
		\draw (69.center) to (72.center);
		\draw (72.center) to (73.center);
		\draw (73.center) to (71.center);
		\draw (71.center) to (70.center);
		\draw (70.center) to (74.center);
		\draw (74.center) to (69.center);
		\draw [style=pointy] (79.center) to (84.center);
		\draw [style=pointy, bend left=90, looseness=1.75] (84.center) to (85.center);
		\draw (86.center) to (89.center);
		\draw (89.center) to (90.center);
		\draw (90.center) to (88.center);
		\draw (88.center) to (87.center);
		\draw (87.center) to (91.center);
		\draw (91.center) to (86.center);
		\draw [style=pointy] (85.center) to (90.center);
		\draw [style=pointy] (92.center) to (93.center);
		\draw (91.center) to (80.center);
		\draw (94.center) to (95.center);
		\draw (95.center) to (92.center);
		\draw [in=180, out=0] (46.center) to (76.center);
		\draw [in=-180, out=0] (57.center) to (81.center);
	\end{pgfonlayer}
\end{tikzpicture}}
\end{equation*}

Combined with axiom [RD.3] for a CRDC, this justifies the following fact, which we record for later use.

\begin{proposition} \cite[Prop. 31]{CRDC} \label{prop:crdctolens}
If $\Ca$ is a CRDC, there is a functor $\textbf{R} \colon \Ca \to \lens{\Ca}$ which on objects sends $A$ to the pair $(A,A)$ and on maps sends $f: A \to B$ to the pair $(f,R[f])$.
\end{proposition}

The following two examples of CRDCs will serve as the basis for the learning scenarios of the upcoming sections.

\begin{example}\label{ex:smoothrdc}
The category $\Smooth$ has as objects natural numbers and maps $n \rightarrow m$ are m-tuples of 
smooth maps $f: \R^n \to \R$. $\Smooth$ is Cartesian with product given by addition. $\Smooth$
is a Cartesian reverse differential category: given a smooth map $f: \R^n \to
\R^m$, the map 
  \[ R[f]: \R^n \times \R^m \to \R^n \]
sends a pair $(x,v)$ to $J[f]^{T}(x)\cdot v$: the transpose of the Jacobian of
$f$ at $x$ in the direction $v$.  \\

For example, if $f: \R^2 \to \R^3$ is defined as $f(x_1,x_2) := (x_1^3 + 2x_1x_2,x_2,\sin(x_1))$, then $R[f]: \R^2 \times \R^3 \to \R^2$ is given by
  $(x,v) \mapsto
  \begin{bmatrix}
  3x_1 + 2x_2 & 0 & \cos(x_1) \\
  2x_1 & 1 & 0
  \end{bmatrix}
  \cdot
  \begin{bmatrix}
  v_1 \\
  v_2 \\
  v_3
  \end{bmatrix}
  $.
Using the reverse derivative (as opposed to the forward derivative\footnote{Forward derivatives can analogously be modelled with Cartesian (forward) differential categories \cite{CFDC}. Given a map $f : A \to B $ the forward derivative $D[f] : A \times A \to B$ uses the Jacobian of $f$, instead of its transpose.}) is well-known to be much more computationally efficient for functions $f: \R^n \to \R^m$ when $m \ll n$ (for example, see \cite{griewank_walther}), as is the case in most supervised learning situations (where often $m = 1$).

\end{example}

\begin{example}
  Another RDC is the PROP $\PolyZ$ \cite[Example 14]{CRDC}, whose morphisms $f : A \to B$
  are $B$-tuples of polynomials $\Z_2[x_1 \ldots x_A]$.
  When presented by generators and relations these morphisms can be viewed as a
  syntax for boolean circuits, with parametric lenses for such circuits (and their reverse derivative) described in
  \cite{rda}.  
\end{example}



\section{Components of learning as Parametric Lenses}
\label{section:components-as-lenses}
As seen in the introduction, in the learning process there are many components at work: a model, an optimiser, a loss map, a learning rate, etc. In this section we show how each such component can be understood as a parametric lens.  Moreover, for each component, we show how our framework encompasses several variations of the gradient-descent algorithms, thus offering a unifying perspective on many different approaches that appear in the literature.

\subsection{Models as Parametric Lenses}\label{subsec:models}

We begin by characterising the models used for training as a parametric lenses. In essence, our approach identifies a set of abstract requirements necessary to perform training by gradient descent, which covers the case studies that we will consider in the next sections. 

The leading intuition is that a suitable model is a parametrised map, equipped with a reverse derivative operator. Using the formal developments of Section~\ref{section:background}, this amounts to assuming that a model is a morphism in $\Para{\Ca}$, for a CRDC $\Ca$. In order to visualise such morphism as a parametric lens, it then suffices to apply $\textbf{R}$ from Proposition~\ref{prop:crdctolens} under $\Para{-}$, yielding a functor  
\begin{equation}
 \para(\textbf{R}): \para(\Ca) \to \para( \lens{\Ca})\label{eq:para_rdc}
\end{equation}



Pictorially, $\Para{\textbf{R}}$ send a map as on the left below to a parametric lens as on the right.
\begin{equation*}
\scalebox{0.7}{\begin{tikzpicture}
	\begin{pgfonlayer}{nodelayer}
		\node [style={medium_box}] (0) at (0, 0) {$f$};
		\node [style=none] (1) at (0, 1.5) {};
		\node [style=none] (2) at (-2, 0) {};
		\node [style=none] (3) at (0, 0.75) {};
		\node [style=none] (4) at (-0.5, 0) {};
		\node [style=none] (5) at (0.25, 0) {};
		\node [style=none] (6) at (2, 0) {};
		\node [style=none] (7) at (0, 2) {$P$};
		\node [style=none] (8) at (-3, 0) {$A$};
		\node [style=none] (9) at (3, 0) {$B$};
	\end{pgfonlayer}
	\begin{pgfonlayer}{edgelayer}
		\draw [style=pointy, in=-180, out=0, looseness=0.75] (2.center) to (4.center);
		\draw [style=pointy] (5.center) to (6.center);
		\draw [style=pointy] (1.center) to (3.center);
	\end{pgfonlayer}
\end{tikzpicture}}
 \qquad\qquad \mapsto\qquad \qquad \scalebox{0.7}{\begin{tikzpicture}
	\begin{pgfonlayer}{nodelayer}
		\node [style=none] (20) at (-8, 5) {};
		\node [style=none] (21) at (-8, -4) {};
		\node [style=none] (22) at (8, 5) {};
		\node [style=none] (23) at (8, -4) {};
		\node [style=none] (26) at (-10, 1) {$A$};
		\node [style=none] (27) at (-10, -2) {$A'$};
		\node [style=none] (28) at (10, 2.5) {$B$};
		\node [style=none] (29) at (10, -2.5) {$B'$};
		\node [style=none] (30) at (3, 3.5) {};
		\node [style=none] (31) at (3, 1.5) {};
		\node [style=none] (32) at (5, 1.5) {};
		\node [style=none] (33) at (5, 3.5) {};
		\node [style=none] (34) at (5, 2.5) {};
		\node [style=none] (35) at (3, 2.25) {};
		\node [style=none] (36) at (-4.25, 1) {};
		\node [style=none] (39) at (-9, 1) {};
		\node [style=none] (40) at (-8, 1) {};
		\node [style=none] (41) at (-3, 2.25) {};
		\node [style=none] (42) at (-3, 0.5) {};
		\node [style=none] (43) at (-8, -2.5) {};
		\node [style=none] (44) at (-9, -2.5) {};
		\node [style=none] (45) at (8, 2.5) {};
		\node [style=none] (46) at (9, 2.5) {};
		\node [style=none] (47) at (6, 0.5) {};
		\node [style=none] (48) at (6, -2) {};
		\node [style=none] (49) at (3, -1) {};
		\node [style=none] (50) at (3, -3) {};
		\node [style=none] (51) at (5, -3) {};
		\node [style=none] (52) at (5, -1) {};
		\node [style=none] (53) at (5, -2) {};
		\node [style=none] (54) at (3, -2.5) {};
		\node [style=none] (55) at (6, -2.5) {};
		\node [style=none] (56) at (5, -2.5) {};
		\node [style=none] (57) at (9, -2.5) {};
		\node [style=none] (58) at (8, -2.5) {};
		\node [style=none] (59) at (4, 2.5) {$f$};
		\node [style=none] (60) at (4, -2) {$R[f]$};
		\node [style=none] (61) at (-5, 3) {};
		\node [style=none] (62) at (-4.25, 2) {};
		\node [style=none] (63) at (-4.25, 2) {};
		\node [style=none] (64) at (-3, 2.75) {};
		\node [style=none] (65) at (-3, 1) {};
		\node [style=none] (66) at (3, 2.75) {};
		\node [style=none] (67) at (6, 1) {};
		\node [style=none] (68) at (6, 1) {};
		\node [style=none] (69) at (6, -1.5) {};
		\node [style=none] (70) at (5, -1.5) {};
		\node [style=none] (71) at (-5, 5) {};
		\node [style=none] (72) at (3, -1.75) {};
		\node [style=none] (73) at (2, -1.75) {};
		\node [style=none] (74) at (1, -1) {};
		\node [style=none] (75) at (1, 5) {};
		\node [style=none] (76) at (-5, 6) {};
		\node [style=none] (77) at (1, 6) {};
		\node [style=none] (78) at (-5, 7) {$P$};
		\node [style=none] (79) at (1, 7) {$P'$};
	\end{pgfonlayer}
	\begin{pgfonlayer}{edgelayer}
		\draw (20.center) to (22.center);
		\draw (22.center) to (23.center);
		\draw (23.center) to (21.center);
		\draw (21.center) to (20.center);
		\draw (39.center) to (40.center);
		\draw [style=pointy] (43.center) to (44.center);
		\draw [style=pointy, in=-180, out=0] (36.center) to (41.center);
		\draw [style=pointy, in=180, out=0] (36.center) to (42.center);
		\draw (40.center) to (36.center);
		\draw [style=pointy] (41.center) to (35.center);
		\draw (34.center) to (45.center);
		\draw [style=pointy] (45.center) to (46.center);
		\draw (30.center) to (33.center);
		\draw (33.center) to (34.center);
		\draw (34.center) to (32.center);
		\draw (32.center) to (31.center);
		\draw (31.center) to (35.center);
		\draw (35.center) to (30.center);
		\draw [style=pointy] (42.center) to (47.center);
		\draw [style=pointy, bend left=90, looseness=1.75] (47.center) to (48.center);
		\draw (49.center) to (52.center);
		\draw (52.center) to (53.center);
		\draw (53.center) to (51.center);
		\draw (51.center) to (50.center);
		\draw (50.center) to (54.center);
		\draw (54.center) to (49.center);
		\draw [style=pointy] (48.center) to (53.center);
		\draw [style=pointy] (55.center) to (56.center);
		\draw (54.center) to (43.center);
		\draw (57.center) to (58.center);
		\draw (58.center) to (55.center);
		\draw [in=180, out=-90] (61.center) to (62.center);
		\draw [style=pointy, in=-180, out=0] (63.center) to (64.center);
		\draw [style=pointy, in=180, out=0] (63.center) to (65.center);
		\draw [style=pointy] (64.center) to (66.center);
		\draw [style=pointy] (65.center) to (67.center);
		\draw [style=pointy, bend left=90, looseness=1.75] (68.center) to (69.center);
		\draw [style=pointy] (69.center) to (70.center);
		\draw [style=pointy] (71.center) to (61.center);
		\draw [style=pointy] (72.center) to (73.center);
		\draw [in=-90, out=-180] (73.center) to (74.center);
		\draw (74.center) to (75.center);
		\draw [style=pointy] (75.center) to (77.center);
		\draw [style=pointy] (76.center) to (71.center);
	\end{pgfonlayer}
\end{tikzpicture}}
\end{equation*}

\begin{example}[Neural networks]
As noted previously, to learn a function of type $\Rb^n \to \Rb^m$, one constructs a neural network, which can be seen as a function of type $\Rb^p \times \Rb^n \to \Rb^m$ where $\Rb^p$ is the space of parameters of the neural network.  As seen in Example~\ref{ex:parasmooth}, this is a map in the category $\Para{\Smooth}$ of type $\Rb^n \to \Rb^m$ with parameter space $\Rb^p$.  Then one can apply the functor in \eqref{eq:para_rdc} to present a neural network together with its reverse derivative operator as a parameterised lens, i.e. a morphism in $\Para{\lens{\Smooth}}$.
\end{example}

\begin{example}[Boolean circuits]
For learning of Boolean circuits as described in \cite{rda}, almost everything is the same as the previous example except that the base category is changed to polynomials over $\mathbb{Z}_2$, $\PolyZ$. To learn a function of type $\mathbb{Z}_2^n \to \mathbb{Z}_2^m$, one constructs some map of type $\mathbb{Z}^p \times \mathbb{Z}^n \to \mathbb{Z}^m$, which one can view as a map in the category $\Para{\PolyZ}$ of type $\mathbb{Z}^n \to \mathbb{Z}^m$ with parameter space $\mathbb{Z}^p$, and again applying the functor in \eqref{eq:para_rdc} for $\PolyZ$ yields a parameterised lens.
\end{example}

Note a model/parametric lens $f$ can take as inputs an element of $A$, an element of $B'$ (a change in $B$) and a parameter $P$ and outputs an element of $B$, a change in $A$, and a change in $P$.  This is not yet sufficient to do machine learning!  When we perform  learning, we want to input a parameter $P$ and a pair $A \times B$ and receive a new parameter $P$.  Instead, $f$ expects a change in $B$ (not an element of $B$) and outputs a change in $P$ (not an element of $P$).  Deep dreaming, on the other hand, wants to return an element of $A$ (not a change in $A$).  Thus, to do machine learning (or deep dreaming) we need to add additional components to $f$; we will consider these additional components in the next sections.

\subsection{Loss Maps as Parametric Lenses}\label{subsec:lossmaps}

Another key component of any learning algorithm is the choice of loss map.  This
gives a measurement of how far the current output of the model is from the
desired output.  In standard learning in $\Smooth$, this loss map is viewed as a
map of type $B \times B \to \Rb$.  However, in our setup, this is naturally
viewed as a parameterised map from $B$ to $\Rb$ with parameter space $B$.\footnote{Here the loss map has its parameter space equal to its input space. However, putting loss maps on the same footing as models lends itself to further generalizations where the parameter space is different, and where the loss map can itself be learned. See Generative Adversarial Networks, \cite[Figure 7.]{TowardsCatCyber}.} We
also generalize the codomain to an arbitrary object $L$.

\begin{definition}
A \textbf{loss map on $B$} consists of a $\para(\Ca)$ map $(\mbox{loss},B): B \to L$ for some object $L$.
\end{definition}

We can then compose such a map with a neural network $(N,P): A \to B$ to get the composite

\begin{equation}\label{eq:paralens-loss-function-with-model}
   \scalebox{0.8}{\begin{tikzpicture}
	\begin{pgfonlayer}{nodelayer}
		\node [style=none] (20) at (-3, 1) {};
		\node [style=none] (21) at (-3, -1) {};
		\node [style=none] (22) at (-1, 1) {};
		\node [style=none] (23) at (-1, -1) {};
		\node [style=none] (26) at (-4.5, 0.5) {$A$};
		\node [style=none] (27) at (-4.5, -0.5) {$A'$};
		\node [style=none] (28) at (0, 1) {$B$};
		\node [style=none] (29) at (0, -1) {$B'$};
		\node [style=none] (39) at (-4, 0.5) {};
		\node [style=none] (40) at (-3, 0.5) {};
		\node [style=none] (43) at (-3, -0.5) {};
		\node [style=none] (44) at (-4, -0.5) {};
		\node [style=none] (45) at (-1, 0.5) {};
		\node [style=none] (46) at (0, 0.5) {};
		\node [style=none] (57) at (0, -0.5) {};
		\node [style=none] (58) at (-1, -0.5) {};
		\node [style=none] (60) at (-2.5, 1) {};
		\node [style=none] (61) at (-1.5, 1) {};
		\node [style=none] (62) at (-2.5, 2) {};
		\node [style=none] (63) at (-1.5, 2) {};
		\node [style=none] (64) at (-2.5, 2.5) {$P$};
		\node [style=none] (65) at (-1.5, 2.5) {$P'$};
		\node [style=none] (66) at (1, 1) {};
		\node [style=none] (67) at (1, -1) {};
		\node [style=none] (68) at (3, 1) {};
		\node [style=none] (69) at (3, -1) {};
		\node [style=none] (72) at (4.5, 0.5) {$L$};
		\node [style=none] (73) at (4.5, -0.5) {$L'$};
		\node [style=none] (74) at (0, 0.5) {};
		\node [style=none] (75) at (1, 0.5) {};
		\node [style=none] (76) at (1, -0.5) {};
		\node [style=none] (77) at (0, -0.5) {};
		\node [style=none] (78) at (3, 0.5) {};
		\node [style=none] (79) at (4, 0.5) {};
		\node [style=none] (80) at (4, -0.5) {};
		\node [style=none] (81) at (3, -0.5) {};
		\node [style=none] (82) at (1.5, 1) {};
		\node [style=none] (83) at (2.5, 1) {};
		\node [style=none] (84) at (1.5, 2) {};
		\node [style=none] (85) at (2.5, 2) {};
		\node [style=none] (86) at (1.5, 2.5) {$B$};
		\node [style=none] (87) at (2.5, 2.5) {$B'$};
		\node [style=none] (88) at (-2, 0) {$N$};
		\node [style=none] (89) at (2, 0) {$\mathsf{loss}$};
	\end{pgfonlayer}
	\begin{pgfonlayer}{edgelayer}
		\draw (20.center) to (22.center);
		\draw (22.center) to (23.center);
		\draw (23.center) to (21.center);
		\draw (21.center) to (20.center);
		\draw [style=pointy] (39.center) to (40.center);
		\draw [style=pointy] (43.center) to (44.center);
		\draw (45.center) to (46.center);
		\draw [style=pointy] (57.center) to (58.center);
		\draw [style=pointy] (62.center) to (60.center);
		\draw [style=pointy] (61.center) to (63.center);
		\draw (66.center) to (68.center);
		\draw (68.center) to (69.center);
		\draw (69.center) to (67.center);
		\draw (67.center) to (66.center);
		\draw [style=pointy] (74.center) to (75.center);
		\draw (76.center) to (77.center);
		\draw [style=pointy] (78.center) to (79.center);
		\draw [style=pointy] (80.center) to (81.center);
		\draw [style=pointy] (84.center) to (82.center);
		\draw [style=pointy] (83.center) to (85.center);
	\end{pgfonlayer}
\end{tikzpicture}}
\end{equation}

Note we can apply $\para$ to either composite $NL$ or $N$ and $L$
individually, giving a parametric lens

\begin{equation}\label{eq:para-commuting-diagram}
   \scalebox{0.8}{\input{para-commuting-diagram.tikz}}
\end{equation}

This is getting closer to the parametric lens we want: it can now receive inputs of type
$B$.  However, this is at the cost of now needing an input to $L'$; we consider
how to handle this in the next section.

\begin{example}[Quadratic error]\label{ex:l2_loss}
In $\Smooth$, the standard loss function on $\Rb^b$ is quadratic error: it uses $L = \Rb$ and has parameterised map $e: \Rb^b  \times \Rb^b \to
\Rb$ given by
	\[ e(b_t,b_p) = \frac{1}{2} \sum_{i=1}^b ((b_p)_i - (b_t)_i)^2. \]\
where we think of $b_t$ as the ``true'' value and $b_p$ the predicted value.
This has reverse derivative $R[e]: \Rb^b \times \Rb^b \times \Rb \to \Rb^b
\times \Rb^b$ given by\footnote{The last argument in the reverse derivative
  implementation is suggestively named $\alpha$ to envoke the idea of \textit{learning rate}, described in the next subsection.}
$R[e](b_t,b_p,\alpha) = \alpha \cdot (b_p-b_t, b_t-b_p)$.
\end{example}




\begin{example}[Boolean error]\label{ex:boolean_error}
In $\PolyZ$, the loss function on $\Z^b$ which is implictly used in \cite{rda} is a bit different: it uses $L = \Z^b$ and has parameterised map $e: \Z^b \times \Z^b \to \Z^b$ given by
	\[ e(b_t,b_p) = b_t + b_p. \]
(Note that this is $+$ in $Z_2$; equivalently this is given by XOR.)   Its reverse derivative is of type $R[e]: \Z^b \times \Z^b \times \Z^b \to \Z^b \times \Z^b$ given by $R[e](b_t,b_p,\alpha) = (\alpha,\alpha)$.
\end{example}

\begin{example}[Softmax cross entropy]\label{ex:softmax_ce}
The Softmax cross entropy loss is a $\Rb^b$-parameterized map $\Rb^b \to \Rb$
defined by
\[
e(b_t, b_p) = \sum_{i:N}(b_t)_i((b_p)_i- \log(\Softmax(b_p)_i)
\]

where $\Softmax(b_p) = \frac{\exp((b_p)_i)}{\sum_{j:N}\exp((b_p)_j)}$ is defined
componentwise for $i:N$.
\end{example}

We note that, although $q$ needs to be a probability distribution, at the moment
there is no need to ponder the question of interaction of probability
distributions with the reverse derivative framework: one can simply consider $q$
as the image of some logits under the $\Softmax$ function.

\begin{example}[Dot product]\label{ex:dot_product}
In Deep Dreaming (Section \ref{subsec:deep_dreaming}) we often want to focus
only on a particular element of the network output $\Rb^b$. This is done by
supplying a one-hot vector $b_t$ as the ground truth to the loss function which
takes two vectors and computes their dot product:
\[
e(b_t, b_p) = b_t \cdot b_p
\]
The reverse derivative of the dot product has the type $\Rb^b \times \Rb^b
\times \Rb \to \Rb^b \times \Rb^b$ and the following implementation
\[
  R[e](b_t, b_p, \alpha) = (\alpha \cdot b_p, \alpha \cdot b_t)
\]
\end{example}

If the ground truth vector $y$ is a one-hot vector (active at the $i$-th
element), then the dot product essentially performs masking of all inputs except
the $i$-th one.

\subsection{Learning Rates as Parametric Lenses}\label{subsec:learningrate}


With a loss function, we are getting closer to our goal of having a parameterised lens which represents a learning process.  We have the following parametrised lens:

\begin{equation}\label{eq:paralens-loss-function-with-model-2}
   \scalebox{0.8}{\begin{tikzpicture}
	\begin{pgfonlayer}{nodelayer}
		\node [style=none] (20) at (-3.5, 1) {};
		\node [style=none] (21) at (-3.5, -1) {};
		\node [style=none] (22) at (-1, 1) {};
		\node [style=none] (23) at (-1, -1) {};
		\node [style=none] (26) at (-5, 0.5) {$A$};
		\node [style=none] (27) at (-5, -0.5) {$A'$};
		\node [style=none] (28) at (0, 1) {$B$};
		\node [style=none] (29) at (0, -1) {$B'$};
		\node [style=none] (39) at (-4.5, 0.5) {};
		\node [style=none] (40) at (-3.5, 0.5) {};
		\node [style=none] (43) at (-3.5, -0.5) {};
		\node [style=none] (44) at (-4.5, -0.5) {};
		\node [style=none] (45) at (-1, 0.5) {};
		\node [style=none] (46) at (0, 0.5) {};
		\node [style=none] (57) at (0, -0.5) {};
		\node [style=none] (58) at (-1, -0.5) {};
		\node [style=none] (60) at (-2.75, 1) {};
		\node [style=none] (61) at (-1.75, 1) {};
		\node [style=none] (62) at (-2.75, 2) {};
		\node [style=none] (63) at (-1.75, 2) {};
		\node [style=none] (64) at (-2.75, 2.5) {$P$};
		\node [style=none] (65) at (-1.75, 2.5) {$P'$};
		\node [style=none] (66) at (1, 1) {};
		\node [style=none] (67) at (1, -1) {};
		\node [style=none] (68) at (3.5, 1) {};
		\node [style=none] (69) at (3.5, -1) {};
		\node [style=none] (72) at (5, 0.5) {$L$};
		\node [style=none] (73) at (5, -0.5) {$L'$};
		\node [style=none] (74) at (0, 0.5) {};
		\node [style=none] (75) at (1, 0.5) {};
		\node [style=none] (76) at (1, -0.5) {};
		\node [style=none] (77) at (0, -0.5) {};
		\node [style=none] (78) at (3.5, 0.5) {};
		\node [style=none] (79) at (4.5, 0.5) {};
		\node [style=none] (80) at (4.5, -0.5) {};
		\node [style=none] (81) at (3.5, -0.5) {};
		\node [style=none] (82) at (1.75, 1) {};
		\node [style=none] (83) at (2.75, 1) {};
		\node [style=none] (84) at (1.75, 2) {};
		\node [style=none] (85) at (2.75, 2) {};
		\node [style=none] (86) at (1.75, 2.5) {$B$};
		\node [style=none] (87) at (2.75, 2.5) {$B'$};
		\node [style=none] (88) at (-2.25, 0.5) {$f$};
		\node [style=none] (89) at (2.25, 0.5) {$\mathsf{loss}$};
		\node [style=none] (90) at (-2.25, -0.5) {$R[f]$};
		\node [style=none] (91) at (2.25, -0.5) {$R[\mathsf{loss}]$};
	\end{pgfonlayer}
	\begin{pgfonlayer}{edgelayer}
		\draw (20.center) to (22.center);
		\draw (22.center) to (23.center);
		\draw (23.center) to (21.center);
		\draw (21.center) to (20.center);
		\draw [style=pointy] (39.center) to (40.center);
		\draw [style=pointy] (43.center) to (44.center);
		\draw (45.center) to (46.center);
		\draw [style=pointy] (57.center) to (58.center);
		\draw [style=pointy] (62.center) to (60.center);
		\draw [style=pointy] (61.center) to (63.center);
		\draw (66.center) to (68.center);
		\draw (68.center) to (69.center);
		\draw (69.center) to (67.center);
		\draw (67.center) to (66.center);
		\draw [style=pointy] (74.center) to (75.center);
		\draw (76.center) to (77.center);
		\draw [style=pointy] (78.center) to (79.center);
		\draw [style=pointy] (80.center) to (81.center);
		\draw [style=pointy] (84.center) to (82.center);
		\draw [style=pointy] (83.center) to (85.center);
	\end{pgfonlayer}
\end{tikzpicture}}
\end{equation}

In this section we focus on the right side of the diagram.  There is an output of $L$ and an input of $L'$ to the diagram.  This is precisely the place where gradient-based learning algorithms input a \emph{learning rate}.

\begin{definition}
A \textbf{learning rate} $\alpha$ on $L$ consists of a lens from $(L,L')$ to $(1,1)$ where $1$ is a terminal object in $\Ca$.
\end{definition}

Note that the get component of such a lens must be the unique map to $1$, while the put component is a map $L \times 1 \to L'$; that is, simply a map $\alpha^*: L \to L'$.   Moreover, we can view $\alpha$ as a $\para(\lens{\Ca})$ map from $(L,L') \to (1,1)$ (with trivial parameter space).  We write such a morphism as a cap, and compose it with the parameterised map above to get

\begin{figure}[h]
  \centering
   \scalebox{0.8}{\begin{tikzpicture}
	\begin{pgfonlayer}{nodelayer}
		\node [style=none] (20) at (-3.5, 1) {};
		\node [style=none] (21) at (-3.5, -1) {};
		\node [style=none] (22) at (-1, 1) {};
		\node [style=none] (23) at (-1, -1) {};
		\node [style=none] (26) at (-5, 0.5) {$A$};
		\node [style=none] (27) at (-5, -0.5) {$A'$};
		\node [style=none] (28) at (0, 1) {$B$};
		\node [style=none] (29) at (0, -1) {$B'$};
		\node [style=none] (39) at (-4.5, 0.5) {};
		\node [style=none] (40) at (-3.5, 0.5) {};
		\node [style=none] (43) at (-3.5, -0.5) {};
		\node [style=none] (44) at (-4.5, -0.5) {};
		\node [style=none] (45) at (-1, 0.5) {};
		\node [style=none] (46) at (0, 0.5) {};
		\node [style=none] (57) at (0, -0.5) {};
		\node [style=none] (58) at (-1, -0.5) {};
		\node [style=none] (60) at (-2.75, 1) {};
		\node [style=none] (61) at (-1.75, 1) {};
		\node [style=none] (62) at (-2.75, 2) {};
		\node [style=none] (63) at (-1.75, 2) {};
		\node [style=none] (64) at (-2.75, 2.5) {$P$};
		\node [style=none] (65) at (-1.75, 2.5) {$P'$};
		\node [style=none] (66) at (1, 1) {};
		\node [style=none] (67) at (1, -1) {};
		\node [style=none] (68) at (3.5, 1) {};
		\node [style=none] (69) at (3.5, -1) {};
		\node [style=none] (72) at (4.5, 1) {$L$};
		\node [style=none] (73) at (4.5, -1) {$L'$};
		\node [style=none] (74) at (0, 0.5) {};
		\node [style=none] (75) at (1, 0.5) {};
		\node [style=none] (76) at (1, -0.5) {};
		\node [style=none] (77) at (0, -0.5) {};
		\node [style=none] (78) at (3.5, 0.5) {};
		\node [style=none] (79) at (4.5, 0.5) {};
		\node [style=none] (80) at (4.5, -0.5) {};
		\node [style=none] (81) at (3.5, -0.5) {};
		\node [style=none] (82) at (1.75, 1) {};
		\node [style=none] (83) at (2.75, 1) {};
		\node [style=none] (84) at (1.75, 2) {};
		\node [style=none] (85) at (2.75, 2) {};
		\node [style=none] (86) at (1.75, 2.5) {$B$};
		\node [style=none] (87) at (2.75, 2.5) {$B'$};
		\node [style=none] (88) at (-2.25, 0.5) {$f$};
		\node [style=none] (89) at (2.25, 0.5) {$\mathsf{loss}$};
		\node [style=none] (90) at (-2.25, -0.5) {$R[f]$};
		\node [style=none] (91) at (2.25, -0.5) {$R[\mathsf{loss}]$};
		\node [style=none] (92) at (5.5, 1) {};
		\node [style=none] (93) at (5.5, -1) {};
		\node [style=none] (94) at (7.5, -1) {};
		\node [style=none] (95) at (8.5, 0) {};
		\node [style=none] (96) at (7.5, 1) {};
		\node [style=none] (97) at (5.5, -0.5) {};
		\node [style=none] (98) at (5.5, 0.5) {};
		\node [style=none] (99) at (6.75, 0) {$\alpha$};
	\end{pgfonlayer}
	\begin{pgfonlayer}{edgelayer}
		\draw (20.center) to (22.center);
		\draw (22.center) to (23.center);
		\draw (23.center) to (21.center);
		\draw (21.center) to (20.center);
		\draw [style=pointy] (39.center) to (40.center);
		\draw [style=pointy] (43.center) to (44.center);
		\draw (45.center) to (46.center);
		\draw [style=pointy] (57.center) to (58.center);
		\draw [style=pointy] (62.center) to (60.center);
		\draw [style=pointy] (61.center) to (63.center);
		\draw (66.center) to (68.center);
		\draw (68.center) to (69.center);
		\draw (69.center) to (67.center);
		\draw (67.center) to (66.center);
		\draw [style=pointy] (74.center) to (75.center);
		\draw (76.center) to (77.center);
		\draw (78.center) to (79.center);
		\draw [style=pointy] (80.center) to (81.center);
		\draw [style=pointy] (84.center) to (82.center);
		\draw [style=pointy] (83.center) to (85.center);
		\draw (92.center) to (96.center);
		\draw (96.center) to (95.center);
		\draw (95.center) to (94.center);
		\draw (94.center) to (93.center);
		\draw (93.center) to (92.center);
		\draw [style=pointy] (79.center) to (98.center);
		\draw [style=pointy] (97.center) to (80.center);
	\end{pgfonlayer}
\end{tikzpicture}}

  \caption{Model composed with a loss function and a learning rate}
  \label{fig:paralens-model-loss-cap}
\end{figure}
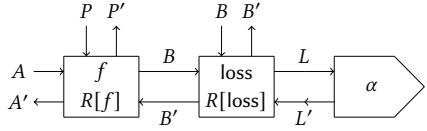
\begin{example}\label{ex:smooth_learning_rate}
In standard supervised learning in $\Smooth$, one fixes some $\epsilon > 0$ as a learning rate, and this is used to define $\alpha$: $\alpha$ is simply constantly $-\epsilon$, ie., $\alpha(l) = -\epsilon$ for any $l \in L$.
\end{example}

\begin{example}\label{ex:boolean_learning_rate}
In supervised learning in $\PolyZ$, the standard learning rate is quite different: for a given $L$ it is defined as the identity function, $\alpha(l) = l$.
\end{example}

Other learning rate morphisms are possible as well: for example, one could fix some $\epsilon > 0$ and define a learning rate in $\Smooth$ by $\alpha(l) = -\epsilon \cdot l$.
Such a learning rate would take into account how far away the network is from
its desired goal and adjust its learning rate accordingly.

\subsection{Optimisers as Reparameterisations}\label{subsec:optimisers}

In the previous sections we have seen how to incorporate the loss map and learning rate into our formalism.  In this section we consider how to implement gradient descent (and its variants) into the picture.  Recall that we are writing our model graphically as the parameterised lens $\scalebox{0.7}{\begin{tikzpicture}
	\begin{pgfonlayer}{nodelayer}
		\node [style=none] (20) at (-1, 1) {};
		\node [style=none] (21) at (-1, -1) {};
		\node [style=none] (22) at (1, 1) {};
		\node [style=none] (23) at (1, -1) {};
		\node [style=none] (26) at (-2.5, 0.5) {$A$};
		\node [style=none] (27) at (-2.5, -0.5) {$A'$};
		\node [style=none] (28) at (2.5, 0.5) {$B$};
		\node [style=none] (29) at (2.5, -0.5) {$B'$};
		\node [style=none] (39) at (-2, 0.5) {};
		\node [style=none] (40) at (-1, 0.5) {};
		\node [style=none] (43) at (-1, -0.5) {};
		\node [style=none] (44) at (-2, -0.5) {};
		\node [style=none] (45) at (1, 0.5) {};
		\node [style=none] (46) at (2, 0.5) {};
		\node [style=none] (57) at (2, -0.5) {};
		\node [style=none] (58) at (1, -0.5) {};
		\node [style=none] (60) at (-0.5, 1) {};
		\node [style=none] (61) at (0.5, 1) {};
		\node [style=none] (62) at (-0.5, 2) {};
		\node [style=none] (63) at (0.5, 2) {};
		\node [style=none] (64) at (-0.5, 2.5) {$P$};
		\node [style=none] (65) at (0.5, 2.5) {$P'$};
	\end{pgfonlayer}
	\begin{pgfonlayer}{edgelayer}
		\draw (20.center) to (22.center);
		\draw (22.center) to (23.center);
		\draw (23.center) to (21.center);
		\draw (21.center) to (20.center);
		\draw [style=pointy] (39.center) to (40.center);
		\draw [style=pointy] (43.center) to (44.center);
		\draw [style=pointy] (45.center) to (46.center);
		\draw [style=pointy] (57.center) to (58.center);
		\draw [style=pointy] (62.center) to (60.center);
		\draw [style=pointy] (61.center) to (63.center);
	\end{pgfonlayer}
\end{tikzpicture}}$.


Note that this diagram outputs a $P'$, which represents a \emph{change} in the parameter space.  But we would like to receive not just the requested change in the parameter, but the new parameter itself.  Thus, we need to add a box above the $P/P'$ wires in the image above to get something whose input and output are both $P$.  Recall that a box in the graphical language is a lens; thus, we are asking for a lens of type $(P,P) \to (P,P')$.  This is precisely what gradient descent accomplishes.

\begin{definition}
In any CRDC $\Ca$ we can define gradient update as a map $G$ in $\lens{\Ca}$ from $(P,P)$ to $(P,P')$ consisting of \[ (G,G^*): (P,P) \to (P,P') \] where $G(p) = p$ and $G^*(p,p') = p + p'$.
\end{definition}
Note that gradient descent is not typically seen as a lens - but it precisely fits this way into the picture we are creating!

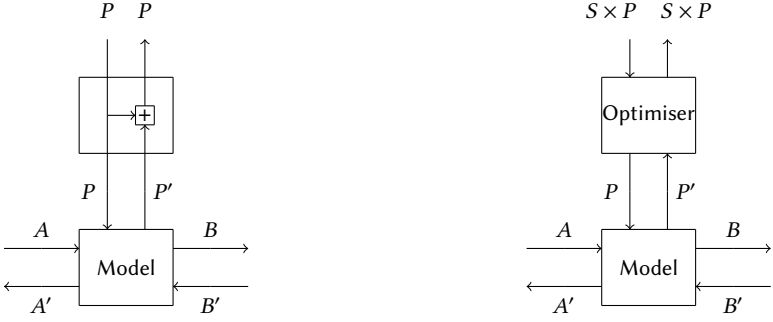
\begin{figure}[h]
\centering
\begin{subfigure}{.5\textwidth}
   \centering
  {\scalefont{0.85}
    \begin{tikzpicture}
	\begin{pgfonlayer}{nodelayer}
		\node [style=none] (20) at (-3.5, 1) {};
		\node [style=none] (21) at (-3.5, -1) {};
		\node [style=none] (22) at (-1, 1) {};
		\node [style=none] (23) at (-1, -1) {};
		\node [style=none] (26) at (-4.5, 1) {$A$};
		\node [style=none] (27) at (-4.5, -1) {$A'$};
		\node [style=none] (28) at (0, 1) {$B$};
		\node [style=none] (29) at (0, -1) {$B'$};
		\node [style=none] (39) at (-4.5, 0.5) {};
		\node [style=none] (40) at (-3.5, 0.5) {};
		\node [style=none] (43) at (-3.5, -0.5) {};
		\node [style=none] (44) at (-4.5, -0.5) {};
		\node [style=none] (45) at (-1, 0.5) {};
		\node [style=none] (46) at (0, 0.5) {};
		\node [style=none] (57) at (0, -0.5) {};
		\node [style=none] (58) at (-1, -0.5) {};
		\node [style=none] (60) at (-2.75, 1) {};
		\node [style=none] (61) at (-1.75, 1) {};
		\node [style=none] (62) at (-2.75, 2) {};
		\node [style=none] (63) at (-1.75, 2) {};
		\node [style=none] (64) at (-3.25, 2) {$P$};
		\node [style=none] (65) at (-1.25, 2) {$P'$};
		\node [style=none] (66) at (1, 1) {};
		\node [style=none] (67) at (1, -1) {};
		\node [style=none] (74) at (0, 0.5) {};
		\node [style=none] (75) at (1, 0.5) {};
		\node [style=none] (76) at (1, -0.5) {};
		\node [style=none] (77) at (0, -0.5) {};
		\node [style=none] (100) at (-3.5, 3) {};
		\node [style=none] (101) at (-1, 3) {};
		\node [style=none] (102) at (-1, 5) {};
		\node [style=none] (103) at (-3.5, 5) {};
		\node [style=none] (105) at (-2.75, 3) {};
		\node [style=none] (106) at (-1.75, 3) {};
		\node [style=none] (107) at (-2.75, 4) {};
		\node [style=none] (108) at (-2.75, 5) {};
		\node [style=none] (110) at (-1.75, 6) {};
		\node [style=none] (111) at (-2.75, 6) {};
		\node [style=none] (112) at (-2, 4.25) {};
		\node [style=none] (113) at (-2, 3.75) {};
		\node [style=none] (114) at (-1.75, 3.75) {};
		\node [style=none] (115) at (-1.5, 3.75) {};
		\node [style=none] (116) at (-1.5, 4.25) {};
		\node [style=none] (117) at (-1.75, 4.25) {};
		\node [style=none] (118) at (-2, 4) {};
		\node [style=none] (119) at (-1.75, 4) {$+$};
		\node [style=none] (122) at (-5.5, 1) {};
		\node [style=none] (123) at (-5.5, -1) {};
		\node [style=none] (126) at (-5.5, 0.5) {};
		\node [style=none] (127) at (-5.5, -0.5) {};
		\node [style=none] (130) at (-2.25, 0) {\textsf{Model}};
		\node [style=none] (136) at (-2.75, 6.75) {$P$};
		\node [style=none] (137) at (-1.75, 6.75) {$P$};
	\end{pgfonlayer}
	\begin{pgfonlayer}{edgelayer}
		\draw (20.center) to (22.center);
		\draw (22.center) to (23.center);
		\draw (23.center) to (21.center);
		\draw (21.center) to (20.center);
		\draw [style=pointy] (39.center) to (40.center);
		\draw (43.center) to (44.center);
		\draw (45.center) to (46.center);
		\draw [style=pointy] (57.center) to (58.center);
		\draw [style=pointy] (62.center) to (60.center);
		\draw (61.center) to (63.center);
		\draw [style=pointy] (74.center) to (75.center);
		\draw (76.center) to (77.center);
		\draw (103.center) to (102.center);
		\draw (102.center) to (101.center);
		\draw (101.center) to (100.center);
		\draw (100.center) to (103.center);
		\draw (62.center) to (105.center);
		\draw (63.center) to (106.center);
		\draw (111.center) to (108.center);
		\draw (108.center) to (107.center);
		\draw (107.center) to (105.center);
		\draw [style=pointy] (107.center) to (118.center);
		\draw [style=pointy] (117.center) to (110.center);
		\draw [style=pointy] (106.center) to (114.center);
		\draw (113.center) to (112.center);
		\draw (112.center) to (116.center);
		\draw (116.center) to (115.center);
		\draw (115.center) to (113.center);
		\draw (126.center) to (39.center);
		\draw [style=pointy] (44.center) to (127.center);
	\end{pgfonlayer}
\end{tikzpicture}
  }
   
\end{subfigure}%
\begin{subfigure}{.5\textwidth}
  \centering
  {\scalefont{0.85}
    \begin{tikzpicture}
	\begin{pgfonlayer}{nodelayer}
		\node [style=none] (20) at (-3.5, 1) {};
		\node [style=none] (21) at (-3.5, -1) {};
		\node [style=none] (22) at (-1, 1) {};
		\node [style=none] (23) at (-1, -1) {};
		\node [style=none] (26) at (-4.5, 1) {$A$};
		\node [style=none] (27) at (-4.5, -1) {$A'$};
		\node [style=none] (28) at (0, 1) {$B$};
		\node [style=none] (29) at (0, -1) {$B'$};
		\node [style=none] (39) at (-4.5, 0.5) {};
		\node [style=none] (40) at (-3.5, 0.5) {};
		\node [style=none] (43) at (-3.5, -0.5) {};
		\node [style=none] (44) at (-4.5, -0.5) {};
		\node [style=none] (45) at (-1, 0.5) {};
		\node [style=none] (46) at (0, 0.5) {};
		\node [style=none] (57) at (0, -0.5) {};
		\node [style=none] (58) at (-1, -0.5) {};
		\node [style=none] (60) at (-2.75, 1) {};
		\node [style=none] (61) at (-1.75, 1) {};
		\node [style=none] (62) at (-2.75, 2) {};
		\node [style=none] (63) at (-1.75, 2) {};
		\node [style=none] (64) at (-3.25, 2) {$P$};
		\node [style=none] (65) at (-1.25, 2) {$P'$};
		\node [style=none] (66) at (1, 1) {};
		\node [style=none] (67) at (1, -1) {};
		\node [style=none] (74) at (0, 0.5) {};
		\node [style=none] (75) at (1, 0.5) {};
		\node [style=none] (76) at (1, -0.5) {};
		\node [style=none] (77) at (0, -0.5) {};
		\node [style=none] (100) at (-3.5, 3) {};
		\node [style=none] (101) at (-1, 3) {};
		\node [style=none] (102) at (-1, 5) {};
		\node [style=none] (103) at (-3.5, 5) {};
		\node [style=none] (105) at (-2.75, 3) {};
		\node [style=none] (106) at (-1.75, 3) {};
		\node [style=none] (108) at (-2.75, 5) {};
		\node [style=none] (110) at (-1.75, 6) {};
		\node [style=none] (111) at (-2.75, 6) {};
		\node [style=none] (117) at (-1.75, 5) {};
		\node [style=none] (122) at (-5.5, 1) {};
		\node [style=none] (123) at (-5.5, -1) {};
		\node [style=none] (126) at (-5.5, 0.5) {};
		\node [style=none] (127) at (-5.5, -0.5) {};
		\node [style=none] (130) at (-2.25, 0) {\textsf{Model}};
		\node [style=none] (136) at (-3.25, 6.75) {$S \times P$};
		\node [style=none] (137) at (-1.25, 6.75) {$S \times P$};
		\node [style=none] (139) at (-2.25, 4) {\textsf{Optimiser}};
	\end{pgfonlayer}
	\begin{pgfonlayer}{edgelayer}
		\draw (20.center) to (22.center);
		\draw (22.center) to (23.center);
		\draw (23.center) to (21.center);
		\draw (21.center) to (20.center);
		\draw [style=pointy] (39.center) to (40.center);
		\draw (43.center) to (44.center);
		\draw (45.center) to (46.center);
		\draw [style=pointy] (57.center) to (58.center);
		\draw [style=pointy] (62.center) to (60.center);
		\draw (61.center) to (63.center);
		\draw [style=pointy] (74.center) to (75.center);
		\draw (76.center) to (77.center);
		\draw (103.center) to (102.center);
		\draw (102.center) to (101.center);
		\draw (101.center) to (100.center);
		\draw (100.center) to (103.center);
		\draw (62.center) to (105.center);
		\draw [style=pointy] (63.center) to (106.center);
		\draw [style=pointy] (111.center) to (108.center);
		\draw [style=pointy] (117.center) to (110.center);
		\draw (126.center) to (39.center);
		\draw [style=pointy] (44.center) to (127.center);
	\end{pgfonlayer}
\end{tikzpicture}
  }

\end{subfigure}
\caption{Model reparameterised by basic gradient descent (left) and a
  generic stateful optimiser (right).}
\label{fig:model_and_optimiser}
\end{figure}

Gradient descent allows one to receive the requested change in parameter and implement that change by adding that value to the current parameter.  We attach this lens, seen as a
reparameterisation, to the top of the diagram above, giving us Figure \ref{fig:model_and_optimiser} (left).

\begin{example}[Gradient update in Smooth]\label{ex:grad_desc_lens}
In $\Smooth$, the gradient descent reparameterisation will take the output from $P'$ and add it to the current value of $P$ to get a new value of~$P$.
\end{example}

\begin{example}[Gradient update in Boolean circuits]\label{ex:boolean_circuit_descent}
In the CRDC $\PolyZ$, the gradient descent reparameterisation will again take the output from $P'$ and add it to the current value of $P$ to get a new value of $P$; however, since $+$ in $\mathbb{Z}_2$ is the same as XOR, this can be also be seen as taking the XOR of the current parameter and the requested change; this is exactly how this algorithm is implemented in \cite{rda}.
\end{example}

Moreover, other variants of gradient descent also fit naturally into this
framework by allowing for additional input/output data with $P$.    In
particular, many important variants of gradient descent keep track of the
history of previous updates and use that to inform the next one.  This is easy to model in our setup: instead of asking for a lens from $(P,P)$ to $(P,P')$, we ask instead for a lens from $(S \times P, S \times P)$ to $(P,P')$ where $S$ is some other object which holds a ``state''.

\begin{definition}
A \textbf{stateful parameter update} consists of a choice of object $S$ (the \textbf{state} object) and a lens $U: (S \times P, S \times P) \to (P, P')$.
\end{definition}

Again, we view this optimizer as a reparameterisation and attach it above the
$P/P'$ wires to the image from the previous section, giving us Figure
~\ref{fig:model_and_optimiser} (right). Let us consider how several well-known optimizers can be implemented in this way.

\begin{example}[Momentum]
In the momentum variant of gradient descent, one keeps track of the previous
change and uses this to inform how the current parameter should be changed.
Thus, in this case, we set $S = P$, fix some $\gamma > 0$, and define the
\textbf{momentum} lens 
	\[ (U,U^*): (P \times P, P \times P) \to (P, P') \]
by $U(s,p) = p$ and $ U^*(s,p,p') = (s', p + s')$, where $ s' = -\gamma s + p'$. Note momentum recovers
  gradient descent when $\gamma = 0$.
\end{example}

In both standard gradient descent and momentum, our lens representation has trivial get/forward part.  Thus it is reasonable to wonder whether this formulation is really capturing the essence of what is going on.  However, as soon as we move to more complicated variants, having non-trivial forward part of the lens is important, and Nesterov momentum is a key example of this.

\begin{example}[Nesterov momentum]\label{ex:nesterov_momentum}
In Nesterov momentum, one makes a modification to the input of the derivative by the previous update.  We can precisely capture this by using a small variation of the lens in the previous example.  Again, we set $S = P$, fix some $\gamma > 0$, and define the \textbf{Nesterov momentum} lens
	\[ (U,U^*): (P \times P, P \times P) \to (P, P') \]
by $U(s,p) = p + \gamma s$ and $U^*$ as in the previous example.
\end{example}

\begin{example}[Adagrad]\label{ex:adagrad}
Given any fixed $\epsilon > 0$ and $\delta \sim 10^{-7}$, Adagrad \cite{Adagrad}
is given by $S = P$, with the lens whose \gett{} part is $(g, p)
\mapsto p$. The \putt{} is $(g, p, p') \mapsto (g', p + \frac{\epsilon}{\delta + \sqrt{g'}} \odot p')$
where $g' = g + p' \odot p'$ and $\odot$ is the elementwise (Hadamard) product.

Unlike with other optimization algorithms where the learning rate is the same
for all parameters, Adagrad divides the learning rate of each individual
parameter with the square root of the past accumulated gradients.
\end{example}

\begin{example}[Adam]\label{ex:adam}
Adaptive Moment Estimation (Adam) \cite{Adam}  is another method that computes adaptive
learning rates for each parameter by storing exponentially decaying average of past
gradients ($m$) and past squared gradients ($v$). Fixed $\beta_1, \beta_2 \in [0, 1)$, $\epsilon > 0$, and $\delta \sim 10^{-8}$, Adam is given by $S = P \times P$, with the lens whose \gett{} part is $(m, v, p) \mapsto p$ and whose \putt{} part is
\[
\putt{}(m, v, p, p') = (\widehat{m}', \widehat{v}', p + \frac{\epsilon}{\delta + \sqrt{\widehat{v}'}} \odot \widehat{m}')
\]

where $m' = \beta_1m + (1 - \beta_1)p'$, $v' = \beta_2v + (1 - \beta_2)p'^2$, and
	$\widehat{m}' = \frac{m'}{1 - \beta_1^t}, \widehat{v}' = \frac{v'}{1 - \beta_2^t}$.

\end{example}

Note that, so far, optimsers/reparameterisations have been added to the $P/P'$ wires, in order to change the model's parameters. We will see in section \ref{subsec:deep_dreaming} how we can also attach them to the $A/A'$ wires instead, giving \emph{deep dreaming}.


\section{Learning with Parametric Lenses}
\label{section:learning-with-lenses}
In the previous section we have seen how all the components of learning can be
modeled as parametric lenses. We now study how all these components can be put together to form supervised learning systems. In addition to studying the most common
examples of supervised learning: systems that learn \textit{parameters}, we also
study different kinds systems: those that learn their \textit{inputs}. This is a
technique commonly known as deep dreaming, and we present it as a natural
counterpart of supervised learning of parameters.

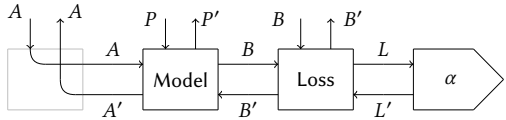
\begin{figure}[h]
  \centering
    \scalebox{0.8}{ \begin{tikzpicture}
	\begin{pgfonlayer}{nodelayer}
		\node [style=none] (20) at (-3.5, 1) {};
		\node [style=none] (21) at (-3.5, -1) {};
		\node [style=none] (22) at (-1, 1) {};
		\node [style=none] (23) at (-1, -1) {};
		\node [style=none] (26) at (-4.5, 1) {$A$};
		\node [style=none] (27) at (-4.5, -1) {$A'$};
		\node [style=none] (28) at (0, 1) {$B$};
		\node [style=none] (29) at (0, -1) {$B'$};
		\node [style=none] (39) at (-4.5, 0.5) {};
		\node [style=none] (40) at (-3.5, 0.5) {};
		\node [style=none] (43) at (-3.5, -0.5) {};
		\node [style=none] (44) at (-4.5, -0.5) {};
		\node [style=none] (45) at (-1, 0.5) {};
		\node [style=none] (46) at (0, 0.5) {};
		\node [style=none] (57) at (0, -0.5) {};
		\node [style=none] (58) at (-1, -0.5) {};
		\node [style=none] (60) at (-2.75, 1) {};
		\node [style=none] (61) at (-1.75, 1) {};
		\node [style=none] (62) at (-2.75, 2) {};
		\node [style=none] (63) at (-1.75, 2) {};
		\node [style=none] (64) at (-3.25, 2) {$P$};
		\node [style=none] (65) at (-1.25, 2) {$P'$};
		\node [style=none] (66) at (1, 1) {};
		\node [style=none] (67) at (1, -1) {};
		\node [style=none] (68) at (3.5, 1) {};
		\node [style=none] (69) at (3.5, -1) {};
		\node [style=none] (72) at (4.5, 1) {$L$};
		\node [style=none] (73) at (4.5, -1) {$L'$};
		\node [style=none] (74) at (0, 0.5) {};
		\node [style=none] (75) at (1, 0.5) {};
		\node [style=none] (76) at (1, -0.5) {};
		\node [style=none] (77) at (0, -0.5) {};
		\node [style=none] (78) at (3.5, 0.5) {};
		\node [style=none] (79) at (4.5, 0.5) {};
		\node [style=none] (80) at (4.5, -0.5) {};
		\node [style=none] (81) at (3.5, -0.5) {};
		\node [style=none] (82) at (1.75, 1) {};
		\node [style=none] (83) at (2.75, 1) {};
		\node [style=none] (84) at (1.75, 2) {};
		\node [style=none] (85) at (2.75, 2) {};
		\node [style=none] (86) at (1, 2) {$B$};
		\node [style=none] (87) at (3.5, 2) {$B'$};
		\node [style=none] (89) at (2.25, 0) {\textsf{Loss}};
		\node [style=none] (92) at (5.5, 1) {};
		\node [style=none] (93) at (5.5, -1) {};
		\node [style=none] (94) at (7.5, -1) {};
		\node [style=none] (95) at (8.5, 0) {};
		\node [style=none] (96) at (7.5, 1) {};
		\node [style=none] (97) at (5.5, -0.5) {};
		\node [style=none] (98) at (5.5, 0.5) {};
		\node [style=none] (99) at (6.75, 0) {$\alpha$};
		\node [style=none] (120) at (-8, 1) {};
		\node [style=none] (121) at (-8, -1) {};
		\node [style=none] (122) at (-5.5, 1) {};
		\node [style=none] (123) at (-5.5, -1) {};
		\node [style=none] (124) at (-8, 0.5) {};
		\node [style=none] (125) at (-8, -0.5) {};
		\node [style=none] (126) at (-5.5, 0.5) {};
		\node [style=none] (127) at (-5.5, -0.5) {};
		\node [style=none] (128) at (-7.25, 1) {};
		\node [style=none] (129) at (-6.25, 1) {};
		\node [style=none] (130) at (-2.25, 0) {\textsf{Model}};
		\node [style=none] (131) at (-6.75, 0.5) {};
		\node [style=none] (132) at (-6.25, 0) {};
		\node [style=none] (133) at (-5.75, -0.5) {};
		\node [style=none] (134) at (-7.25, 2) {};
		\node [style=none] (135) at (-6.25, 2) {};
		\node [style=none] (138) at (-7.75, 2.25) {$A$};
		\node [style=none] (139) at (-5.75, 2.25) {$A$};
	\end{pgfonlayer}
	\begin{pgfonlayer}{edgelayer}
		\draw (20.center) to (22.center);
		\draw (22.center) to (23.center);
		\draw (23.center) to (21.center);
		\draw (21.center) to (20.center);
		\draw [style=pointy] (39.center) to (40.center);
		\draw (43.center) to (44.center);
		\draw (45.center) to (46.center);
		\draw [style=pointy] (57.center) to (58.center);
		\draw [style=pointy] (62.center) to (60.center);
		\draw [style=pointy] (61.center) to (63.center);
		\draw (66.center) to (68.center);
		\draw (68.center) to (69.center);
		\draw (69.center) to (67.center);
		\draw (67.center) to (66.center);
		\draw [style=pointy] (74.center) to (75.center);
		\draw (76.center) to (77.center);
		\draw (78.center) to (79.center);
		\draw [style=pointy] (80.center) to (81.center);
		\draw [style=pointy] (84.center) to (82.center);
		\draw [style=pointy] (83.center) to (85.center);
		\draw (92.center) to (96.center);
		\draw (96.center) to (95.center);
		\draw (95.center) to (94.center);
		\draw (94.center) to (93.center);
		\draw (93.center) to (92.center);
		\draw [style=pointy] (79.center) to (98.center);
		\draw (97.center) to (80.center);
		\draw [style=lightnone] (120.center) to (122.center);
		\draw [style=lightnone] (122.center) to (123.center);
		\draw [style=lightnone] (123.center) to (121.center);
		\draw [style=lightnone] (121.center) to (120.center);
		\draw (126.center) to (39.center);
		\draw (44.center) to (127.center);
		\draw (126.center) to (131.center);
		\draw [in=-90, out=-180, looseness=1.25] (131.center) to (128.center);
		\draw [bend right=315, looseness=1.25] (133.center) to (132.center);
		\draw (127.center) to (133.center);
		\draw (132.center) to (129.center);
		\draw [style=pointy] (134.center) to (128.center);
		\draw [style=pointy] (129.center) to (135.center);
	\end{pgfonlayer}
\end{tikzpicture}}
  \caption{Closed parametric lens whose all inputs and outputs are now vertical wires.}
  \label{fig:supervised}
\end{figure}

Before we describe these systems, it will be convenient to represent all the
inputs and outputs of our parametric lenses as parameters. In Figure ~\ref{fig:paralens-model-loss-cap}, we see the $P/P'$ and $B/B'$ inputs and outputs as parameters; however, the
$A/A'$ wires are not. To view the $A/A'$ inputs as parameters, we compose that
system with the parameterised lens $\eta$ we now define. The parameterised lens
$\eta$ has the type $(1,1) \to (A,A')$ with parameter space $(A, A')$ defined by
$(\gett{}_{\eta} = 1_A, \putt{}_{\eta} = \pi_1)$ and can be depicted graphically as $\scalebox{0.7}{ \begin{tikzpicture}
	\begin{pgfonlayer}{nodelayer}
		\node [style=none] (26) at (-4.5, 1) {$A$};
		\node [style=none] (27) at (-4.5, -1) {$A'$};
		\node [style=none] (39) at (-4.5, 0.5) {};
		\node [style=none] (44) at (-4.5, -0.5) {};
		\node [style=none] (120) at (-8, 1) {};
		\node [style=none] (121) at (-8, -1) {};
		\node [style=none] (122) at (-5.5, 1) {};
		\node [style=none] (123) at (-5.5, -1) {};
		\node [style=none] (124) at (-8, 0.5) {};
		\node [style=none] (125) at (-8, -0.5) {};
		\node [style=none] (126) at (-5.5, 0.5) {};
		\node [style=none] (127) at (-5.5, -0.5) {};
		\node [style=none] (128) at (-7.25, 1) {};
		\node [style=none] (129) at (-6.25, 1) {};
		\node [style=none] (131) at (-6.75, 0.5) {};
		\node [style=none] (132) at (-6.25, 0) {};
		\node [style=none] (133) at (-5.75, -0.5) {};
		\node [style=none] (134) at (-7.25, 2) {};
		\node [style=none] (135) at (-6.25, 2) {};
		\node [style=none] (138) at (-7.75, 2.25) {$A$};
	\end{pgfonlayer}
	\begin{pgfonlayer}{edgelayer}
		\draw [style=lightnone] (120.center) to (122.center);
		\draw [style=lightnone] (122.center) to (123.center);
		\draw [style=lightnone] (123.center) to (121.center);
		\draw [style=lightnone] (121.center) to (120.center);
		\draw [style=pointy] (126.center) to (39.center);
		\draw (44.center) to (127.center);
		\draw (126.center) to (131.center);
		\draw [in=-90, out=-180, looseness=1.25] (131.center) to (128.center);
		\draw [bend right=315, looseness=1.25] (133.center) to (132.center);
		\draw (127.center) to (133.center);
		\draw (132.center) to (129.center);
		\draw (134.center) to (128.center);
		\draw [style=pointy] (129.center) to (135.center);
	\end{pgfonlayer}
\end{tikzpicture}}$.
Composing $\eta$ with the rest of the learning system in Figure
~\ref{fig:paralens-model-loss-cap} gives us the closed parametric lens in Figure ~\ref{fig:supervised}.

This composite is now a map in $\para(\lens{\Ca})$ from $(1,1)$ to $(1,1)$; all
its inputs and outputs are now vertical wires, ie., parameters. Unpacking it
further, this is a lens of type $(A \times P \times B, A' \times P' \times B')
\to (1, 1)$ whose $\gett{}$ map is the terminal map, and whose $\putt{}$ map is
of the type $A \times P \times B \to A' \times P' \times B'$. It can be unpacked
as the composite
\begin{align*}
  \label{eq:putmap}
  \putt{}(a, p, b_t) = (a', p', b_t') \qquad \texttt{ where } \quad \qquad b_p &= f(p, a) \\
  (b'_t, b'_p) &= R[\loss](b_t, b_p, \alpha(\loss(b_t, b_p))) \\
  (p', a') &= R[f](p, a, b'_p)
\end{align*}

In the next two sections we consider further additions to the image above which
correspond to different types of supervised learning: supervised learning of
parameters and supervised learning of inputs.

\subsection{Supervised Learning of Parameters}\label{subsec:learning-parameters}

\begin{figure}[h]
  \centering
  {\scalefont{0.8}
    \begin{tikzpicture}
	\begin{pgfonlayer}{nodelayer}
		\node [style=none] (20) at (-3.5, 1) {};
		\node [style=none] (21) at (-3.5, -1) {};
		\node [style=none] (22) at (-1, 1) {};
		\node [style=none] (23) at (-1, -1) {};
		\node [style=none] (26) at (-4.5, 1) {$A$};
		\node [style=none] (27) at (-4.5, -1) {$A'$};
		\node [style=none] (28) at (0, 1) {$B$};
		\node [style=none] (29) at (0, -1) {$B'$};
		\node [style=none] (39) at (-4.5, 0.5) {};
		\node [style=none] (40) at (-3.5, 0.5) {};
		\node [style=none] (43) at (-3.5, -0.5) {};
		\node [style=none] (44) at (-4.5, -0.5) {};
		\node [style=none] (45) at (-1, 0.5) {};
		\node [style=none] (46) at (0, 0.5) {};
		\node [style=none] (57) at (0, -0.5) {};
		\node [style=none] (58) at (-1, -0.5) {};
		\node [style=none] (60) at (-2.75, 1) {};
		\node [style=none] (61) at (-1.75, 1) {};
		\node [style=none] (62) at (-2.75, 2) {};
		\node [style=none] (63) at (-1.75, 2) {};
		\node [style=none] (64) at (-3.25, 2) {$P$};
		\node [style=none] (65) at (-1.25, 2) {$P'$};
		\node [style=none] (66) at (1, 1) {};
		\node [style=none] (67) at (1, -1) {};
		\node [style=none] (68) at (3.5, 1) {};
		\node [style=none] (69) at (3.5, -1) {};
		\node [style=none] (72) at (4.5, 1) {$L$};
		\node [style=none] (73) at (4.5, -1) {$L'$};
		\node [style=none] (74) at (0, 0.5) {};
		\node [style=none] (75) at (1, 0.5) {};
		\node [style=none] (76) at (1, -0.5) {};
		\node [style=none] (77) at (0, -0.5) {};
		\node [style=none] (78) at (3.5, 0.5) {};
		\node [style=none] (79) at (4.5, 0.5) {};
		\node [style=none] (80) at (4.5, -0.5) {};
		\node [style=none] (81) at (3.5, -0.5) {};
		\node [style=none] (82) at (1.75, 1) {};
		\node [style=none] (83) at (2.75, 1) {};
		\node [style=none] (84) at (1.75, 6) {};
		\node [style=node] (85) at (2.75, 2.5) {};
		\node [style=none] (86) at (1.75, 6.5) {$B$};
		\node [style=none] (87) at (3.25, 1.5) {$B'$};
		\node [style=none] (89) at (2.25, 0) {\textsf{Loss}};
		\node [style=none] (92) at (5.5, 1) {};
		\node [style=none] (93) at (5.5, -1) {};
		\node [style=none] (94) at (7.5, -1) {};
		\node [style=none] (95) at (8.5, 0) {};
		\node [style=none] (96) at (7.5, 1) {};
		\node [style=none] (97) at (5.5, -0.5) {};
		\node [style=none] (98) at (5.5, 0.5) {};
		\node [style=none] (99) at (6.75, 0) {$\alpha$};
		\node [style=none] (100) at (-3.5, 3) {};
		\node [style=none] (101) at (-1, 3) {};
		\node [style=none] (102) at (-1, 5) {};
		\node [style=none] (103) at (-3.5, 5) {};
		\node [style=none] (105) at (-2.75, 3) {};
		\node [style=none] (106) at (-1.75, 3) {};
		\node [style=none] (108) at (-2.75, 5) {};
		\node [style=none] (110) at (-1.75, 6) {};
		\node [style=none] (111) at (-2.75, 6) {};
		\node [style=none] (117) at (-1.75, 5) {};
		\node [style=none] (120) at (-8, 1) {};
		\node [style=none] (121) at (-8, -1) {};
		\node [style=none] (122) at (-5.5, 1) {};
		\node [style=none] (123) at (-5.5, -1) {};
		\node [style=none] (124) at (-8, 0.5) {};
		\node [style=none] (125) at (-8, -0.5) {};
		\node [style=none] (126) at (-5.5, 0.5) {};
		\node [style=none] (127) at (-5.5, -0.5) {};
		\node [style=none] (128) at (-7.25, 1) {};
		\node [style=none] (129) at (-6.25, 1) {};
		\node [style=none] (130) at (-2.25, 0) {\textsf{Model}};
		\node [style=none] (131) at (-6.75, 0.5) {};
		\node [style=none] (132) at (-6.25, 0) {};
		\node [style=none] (133) at (-5.75, -0.5) {};
		\node [style=none] (134) at (-7.25, 6) {};
		\node [style=node] (135) at (-6.25, 2.5) {};
		\node [style=none] (136) at (-3.25, 6.75) {$S \times P$};
		\node [style=none] (137) at (-1.25, 6.75) {$S \times P$};
		\node [style=none] (138) at (-7.25, 6.75) {$A$};
		\node [style=none] (139) at (-2.25, 4) {\textsf{Optimiser}};
	\end{pgfonlayer}
	\begin{pgfonlayer}{edgelayer}
		\draw (20.center) to (22.center);
		\draw (22.center) to (23.center);
		\draw (23.center) to (21.center);
		\draw (21.center) to (20.center);
		\draw [style=pointy] (39.center) to (40.center);
		\draw (43.center) to (44.center);
		\draw (45.center) to (46.center);
		\draw [style=pointy] (57.center) to (58.center);
		\draw [style=pointy] (62.center) to (60.center);
		\draw (61.center) to (63.center);
		\draw (66.center) to (68.center);
		\draw (68.center) to (69.center);
		\draw (69.center) to (67.center);
		\draw (67.center) to (66.center);
		\draw [style=pointy] (74.center) to (75.center);
		\draw (76.center) to (77.center);
		\draw (78.center) to (79.center);
		\draw [style=pointy] (80.center) to (81.center);
		\draw [style=pointy] (84.center) to (82.center);
		\draw (83.center) to (85);
		\draw (92.center) to (96.center);
		\draw (96.center) to (95.center);
		\draw (95.center) to (94.center);
		\draw (94.center) to (93.center);
		\draw (93.center) to (92.center);
		\draw [style=pointy] (79.center) to (98.center);
		\draw [style=pointy] (97.center) to (80.center);
		\draw (103.center) to (102.center);
		\draw (102.center) to (101.center);
		\draw (101.center) to (100.center);
		\draw (100.center) to (103.center);
		\draw (62.center) to (105.center);
		\draw (63.center) to (106.center);
		\draw [style=pointy] (111.center) to (108.center);
		\draw [style=pointy] (117.center) to (110.center);
		\draw [style=lightnone] (120.center) to (122.center);
		\draw [style=lightnone] (122.center) to (123.center);
		\draw [style=lightnone] (123.center) to (121.center);
		\draw [style=lightnone] (121.center) to (120.center);
		\draw (126.center) to (39.center);
		\draw [style=pointy] (44.center) to (127.center);
		\draw (126.center) to (131.center);
		\draw [in=-90, out=-180, looseness=1.25] (131.center) to (128.center);
		\draw [bend right=315, looseness=1.25] (133.center) to (132.center);
		\draw (127.center) to (133.center);
		\draw (132.center) to (129.center);
		\draw [style=pointy] (134.center) to (128.center);
		\draw (129.center) to (135);
	\end{pgfonlayer}
\end{tikzpicture}
  }
  \caption[Supervised learning diagram 2]{Closed parametric lens whose parameters are being learned. An
    animation of this supervised learning system is available online. \footnotemark }
  \label{fig:paralens-full-learner}
\end{figure}
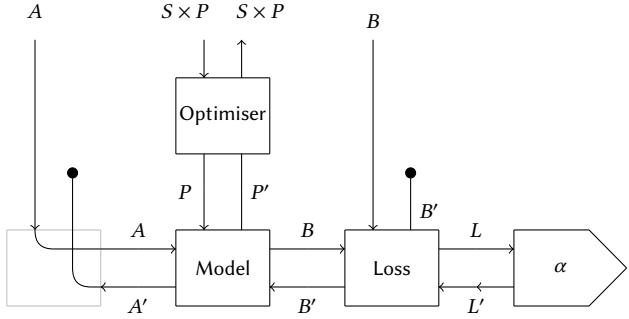

The most common type of learning that is performed on the image in Figure
\ref{fig:supervised} is supervised learning of \textit{parameters}.
This is done by reparameterising the image (Def. \ref{def:reparameterisation})
in the following manner. The parameter ports are reparameterised by one of the
(potentially stateful) optimisers described in the previous section, while the backward wires $A'$ of inputs and $B'$ of outputs are discarded.
This finally gives us the complete picture of a system which learns the
parameters in a supervised manner (Figure ~\ref{fig:paralens-full-learner}).

\footnotetext{See footnote 1.}
Fixing a particular optimiser $(U, U^*) : (S \times P, S \times P) \to (P, P')$
we again unpack the entire construction. This is a map in $\Para{\lens{\Ca}}$
from $(1, 1)$ to $(1, 1)$ whose parameter space is $(A \times S \times P \times
B, S \times P)$. In other words, this is a lens of type $ (A \times S \times P \times B, S \times P)  \to (1, 1)$ whose $\gett{}$ component is identity. Its $\putt{}$ map has the type $A \times S \times P \times B \to S \times P$ and unpacks to

\begin{align*}
  \putt{}(a, s, p, b_t) = U^*(s, p, p') \qquad \texttt{ where } \quad \qquad \overline{p}  &= U(s, p) \\
  b_p &= f(\overline{p}, a) \\
  (b'_t, b'_p) &= R[\loss](b_t, b_p, \alpha(\loss(b_t, b_p))) \\
  (p', a') &= R[f](\overline{p}, a, b'_p)
\end{align*}

While this formulation might seem daunting, we note that it just explicitly
specifies the computation performed by a supervised learning system. The
variable $\overline{p}$ represents the parameter supplied to the network by the
stateful gradient update rule (in many cases this is equal to $p$); $b_p$
represents the prediction of the network (contrast this with $b_t$ which
represents the ground truth from the dataset), Variables with a tick $'$
represent changes: $ b'_p$ and $b'_t$ are the changes on predictions and true
values respectively, while $p'$ and $a'$ are changes on the parameters and
inputs.
Furthermore, this arises automatically out of the rule for lens composition
(Figure ~\ref{eq:lens-composition}); what we needed to specify is just the lenses themselves.

We justify and illustrate our approach on a series of case studies drawn from the machine learning literature, showing how in each case the parameters of our framework (in particular, loss functions and gradient updates) instantiate to familiar concepts. This presentation has the advantage of treating all these case studies uniformly in terms of our basic constructs, highlighting their similarities and differences.

We start in $\Smooth$, fixing some parameterised map $(\Rb^p, f) :
\Para{\Smooth}(\Rb^a, \Rb^b)$ and the constant \textit{negative} learning rate $\alpha : \Rb$ (Example ~\ref{ex:smooth_learning_rate}). We then vary the loss
function and the gradient update, seeing how the $\putt{}$ map above reduces to
many of the known cases in the literature.

\begin{example}[Quadratic error, basic gradient descent]\label{ex:quad_err_grad_desc}
Fix the quadratic error (Example \ref{ex:l2_loss}) as the loss map and basic
gradient update (Example \ref{ex:grad_desc_lens}). Then the aforementioned $\putt{}$ map
simplifies. Since there is no state, its type reduces to $A \times P \times B
\to P$, and its implementation to: $\putt{}(a, p, b_t) = p + p'$, where $(p',
a') = R[f](p, a, \alpha \cdot (f(p, a) - b_t))$.

Note that $\alpha$ here is simply a constant, and due to the linearity of the
reverse derivative (Def ~\ref{def:crdc}), we can slide the $\alpha$ from the costate
into the gradient descent lens. Rewriting this update, and performing this
sliding we obtain a closed form update step
\begin{align*}
  \putt{}(a, p, b_t) &= p + \alpha \cdot (R[f](p, a, f(p, a) - b_t);\pi_0)
\end{align*}
where the negative \textit{descent} component of gradient descent is here contained in the choice of the negative constant $\alpha$.
\end{example}

This example gives us a variety of \textit{regression} algorithms solved
iteratively by gradient descent: it embeds some parameterised map $(\R^p, f) :
\para(\Smooth)(\R^a, \R^b)$ into the system which performs regression on input
data - where $a$ denotes the input to the model and $b_t$ denotes the ground truth. If the corresponding map $f$ is linear and $m = 1$, we recover simple linear
regression with gradient descent. If the codomain is additionally
multi-dimensional, i.e. we're predicting multiple scalars, then we recover
multivariate linear regression. Likewise, we can model a
multi-layer perceptron or even more complex neural network architectures
performing supervised learning of parameters simply
by changing the underlying parameterised map.

\begin{example}[Softmax cross entropy, basic gradient descent]
Fix Softmax cross entropy (Example \ref{ex:softmax_ce}) as the loss map and
basic gradient update (Example \ref{ex:grad_desc_lens}). Again the $\putt{}$ map simplifies. The type
reduces to $A \times P \times B \to P$ and the implementation to
\begin{align*}
  \putt{}(a, p, b_t) &= p + p'
\end{align*}
where $(p', a') = R[f](\overline{p}, a, \alpha \cdot (\Softmax(f(p, a)) - b_t))$.
The same rewriting performed on the previous example can be done here.
\end{example}

This example recovers \textit{logistic regression}, e.g. classification.

\begin{example}[Mean squared error, Nesterov Momentum]
Fix the quadratic error (Example \ref{ex:l2_loss}) as the loss map and Nesterov
momentum (Example \ref{ex:nesterov_momentum}) as the gradient update. This time the
$\putt{}$ map doesn't have a simplified type, it is still $A \times S \times P
\times B \to S \times P$. The implementation of $\putt{}$ reduces to

\begin{align*}
  \putt{}(a, s, p, b_t) = (s', p + s') \qquad \texttt{ where } \quad \qquad  \overline{p}  &= p + \gamma s \\
  (p', a') &= R[f](\overline{p}, a, \alpha \cdot (f(\overline{p}, a) - b_t)) \\
  s' &= - \gamma s + p' 
\end{align*}
\end{example}

This example with Nesterov momentum differs in two key points from all the other
ones: i) the optimiser is stateful, and ii) its $\gett{}$ map is not trivial.
While many other optimisers are stateful, the non-triviality of the $\gett{}$
map here showcases the importance of lenses. They allow us to make precise the notion
of computing a ``lookahead'' value for Nesterov momentum, something that is in
practice usually handled in ad-hoc ways. Here, the algebra of lens composition handles this case naturally by
using the $\gett{}$ map, a seemingly trivial, unused piece of data for previous optimisers.

We finish off these examples by moving to a different base category $\PolyZ$.
This example shows that our framework describes learning in not just
continuous, but discrete settings too.
Again, we fix a parameterised map $(\Zb^p, f) : \PolyZ(\Rb^a, \Rb^b)$ but this
time we fix the identity learning rate (Example \ref{ex:boolean_learning_rate}),
instead of a constant one.

\begin{example}[Basic learning in Boolean circuits]
Fix XOR as the loss map (Example \ref{ex:boolean_error}) and the basic gradient update (Example \ref{ex:boolean_circuit_descent}). The put map again simplifies. The type reduces to $A \times P \times B \to P$ and the implementation to $\putt{}(a, p, b_t) = p + p'$ where $(p', a') = R[f](p, a, f(p, a) + b_t)$.
\end{example}

\textbf{A sketch of learning iteration.}
Having described a number of examples in supervised learning, we outline how to
model learning iteration in our framework.
Recall the aforementioned  \putt{} map whose type is $A \times P
\times B \to P$ (for simplicity here modelled without state $S$). This map takes an
input-output pair $(a_0, b_0)$, the current parameter $p_i$ and produces an updated
parameter $p_{i + 1}$. At the next time step, it takes a potentially different
input-output pair $(a_1, b_1)$, the updated parameter $p_{i + 1}$ and produces
$p_{i + 2}$. This process is then repeated. We can model this iteration as a composition of the $\putt{}$ map with itself, as a composite $(A \times \putt{} \times B);\putt{}$
whose type is $A \times A \times P \times B \times B \to P$. This map takes two
input-output pairs $A \times B$, a parameter and produces a new parameter by
processing these datapoints in sequence. One can see how
this process can be iterated any number of times, and even represented as a
string diagram.

But we note that with a slight reformulation of the \putt{} map, it is possible to obtain a conceptually much simpler definition. The key insight lies in seeing that the
map $\putt{} : A \times P \times B \to P$ is essentially an endo-map $P \to P$ with some extra inputs $A \times B$; it's a parameterised map!

In other words, we can recast the $\putt{}$ map as a parameterised map $(A
\times B, \putt{}) : \Para{\Ca}(P, P)$. Since it is an endo-map, it can be
composed with itself. The resulting composite is too an endo-map, which now
takes two ``parameters'': input-output pair at the time step $0$ and time step
$1$. This process can then be repeated, with $\para$ composition automatically taking care of the algebra of iteration.

\begin{equation*}
  \scalebox{0.8}{\begin{tikzpicture}
	\begin{pgfonlayer}{nodelayer}
		\node [style=none] (27) at (-6, 0.5) {};
		\node [style=none] (28) at (-4, -0.5) {};
		\node [style=none] (29) at (-4, 0.5) {};
		\node [style=none] (30) at (-6, -0.5) {};
		\node [style=none] (31) at (-6, 0) {};
		\node [style=none] (33) at (-5, 0.5) {};
		\node [style=none] (34) at (-4, 0) {};
		\node [style=none] (35) at (-2, 0) {};
		\node [style=none] (37) at (-7, 0) {};
		\node [style=none] (38) at (-7.5, 0) {$P$};
		\node [style=none] (39) at (-5, 0) {\textsf{put}};
		\node [style=none] (41) at (-5, 1.75) {};
		\node [style=none] (43) at (-5, 2.25) {$A \times B$};
		\node [style=none] (44) at (-2, 0.5) {};
		\node [style=none] (45) at (0, -0.5) {};
		\node [style=none] (46) at (0, 0.5) {};
		\node [style=none] (47) at (-2, -0.5) {};
		\node [style=none] (48) at (-2, 0) {};
		\node [style=none] (50) at (-1, 0.5) {};
		\node [style=none] (51) at (0, 0) {};
		\node [style=none] (52) at (1, 0) {};
		\node [style=none] (55) at (-3, 0.5) {$P$};
		\node [style=none] (56) at (-1, 0) {\textsf{put}};
		\node [style=none] (58) at (-1, 1.75) {};
		\node [style=none] (60) at (-1, 2.25) {$A \times B$};
		\node [style=none] (61) at (3.75, 0.5) {};
		\node [style=none] (62) at (5.75, -0.5) {};
		\node [style=none] (63) at (5.75, 0.5) {};
		\node [style=none] (64) at (3.75, -0.5) {};
		\node [style=none] (65) at (3.75, 0) {};
		\node [style=none] (67) at (4.75, 0.5) {};
		\node [style=none] (68) at (5.75, 0) {};
		\node [style=none] (69) at (6.75, 0) {};
		\node [style=none] (70) at (7.25, 0) {$P$};
		\node [style=none] (71) at (2.75, 0) {};
		\node [style=none] (72) at (4.75, 0) {\textsf{put}};
		\node [style=none] (74) at (4.75, 1.75) {};
		\node [style=none] (76) at (4.75, 2.25) {$A \times B$};
		\node [style=none] (77) at (1.75, 0) {$\overset{n}{\ldots}$};
	\end{pgfonlayer}
	\begin{pgfonlayer}{edgelayer}
		\draw (27.center) to (31.center);
		\draw (31.center) to (30.center);
		\draw (28.center) to (30.center);
		\draw (28.center) to (34.center);
		\draw (34.center) to (29.center);
		\draw (29.center) to (33.center);
		\draw [style=pointy] (37.center) to (31.center);
		\draw [style=pointy] (34.center) to (35.center);
		\draw [style=pointy] (41.center) to (33.center);
		\draw (44.center) to (48.center);
		\draw (48.center) to (47.center);
		\draw (45.center) to (47.center);
		\draw (45.center) to (51.center);
		\draw (51.center) to (46.center);
		\draw (46.center) to (50.center);
		\draw [style=pointy] (51.center) to (52.center);
		\draw [style=pointy] (58.center) to (50.center);
		\draw (61.center) to (65.center);
		\draw (65.center) to (64.center);
		\draw (62.center) to (64.center);
		\draw (62.center) to (68.center);
		\draw (68.center) to (63.center);
		\draw (63.center) to (67.center);
		\draw [style=pointy] (71.center) to (65.center);
		\draw [style=pointy] (68.center) to (69.center);
		\draw [style=pointy] (74.center) to (67.center);
		\draw (27.center) to (33.center);
		\draw (44.center) to (50.center);
		\draw (61.center) to (67.center);
	\end{pgfonlayer}
\end{tikzpicture}}
\end{equation*}

This reformulation captures the essence of parameter iteration: one can think of it as a
trajectory $p_i, p_{i + 1}, p_{i + 2}, ...$ through the parameter space; but
it is a \textit{trajectory parameterised by the dataset}. With different datasets the
algorithm will take a different path through this space and learn different things.

\subsection{Deep Dreaming: Supervised Learning of Inputs}\label{subsec:deep_dreaming}

We have seen that attaching gradient descent to the parameter port of the
parametric lens as a reparameterisation allows us to learn the parameters in a supervised way.
In this section we describe how attaching the gradient descent lens to the \textit{input
port} provides us with a way to enhance an input image to elicit a particular
interpretation. This is the idea behind the technique called Deep Dreaming,
appearing in the literature in many forms \cite{DeepDreaming1, DeepDreaming2,
  DeepDreaming3, DeepDreaming4}.

\begin{figure}[h]
  \centering
  {\scalefont{0.8}
    \begin{tikzpicture}
	\begin{pgfonlayer}{nodelayer}
		\node [style=none] (20) at (-3.5, 1) {};
		\node [style=none] (21) at (-3.5, -1) {};
		\node [style=none] (22) at (-1, 1) {};
		\node [style=none] (23) at (-1, -1) {};
		\node [style=none] (26) at (-4.5, 1) {$A$};
		\node [style=none] (27) at (-4.5, -1) {$A'$};
		\node [style=none] (28) at (0, 1) {$B$};
		\node [style=none] (29) at (0, -1) {$B'$};
		\node [style=none] (39) at (-4.5, 0.5) {};
		\node [style=none] (40) at (-3.5, 0.5) {};
		\node [style=none] (43) at (-3.5, -0.5) {};
		\node [style=none] (44) at (-4.5, -0.5) {};
		\node [style=none] (45) at (-1, 0.5) {};
		\node [style=none] (46) at (0, 0.5) {};
		\node [style=none] (57) at (0, -0.5) {};
		\node [style=none] (58) at (-1, -0.5) {};
		\node [style=none] (60) at (-7.25, 1) {};
		\node [style=none] (61) at (-6.25, 1) {};
		\node [style=none] (62) at (-7.25, 2) {};
		\node [style=none] (63) at (-6.25, 2) {};
		\node [style=none] (64) at (-7.75, 2) {$A$};
		\node [style=none] (65) at (-5.75, 2) {$A'$};
		\node [style=none] (66) at (1, 1) {};
		\node [style=none] (67) at (1, -1) {};
		\node [style=none] (68) at (3.5, 1) {};
		\node [style=none] (69) at (3.5, -1) {};
		\node [style=none] (72) at (4.5, 1) {$L$};
		\node [style=none] (73) at (4.5, -1) {$L'$};
		\node [style=none] (74) at (0, 0.5) {};
		\node [style=none] (75) at (1, 0.5) {};
		\node [style=none] (76) at (1, -0.5) {};
		\node [style=none] (77) at (0, -0.5) {};
		\node [style=none] (78) at (3.5, 0.5) {};
		\node [style=none] (79) at (4.5, 0.5) {};
		\node [style=none] (80) at (4.5, -0.5) {};
		\node [style=none] (81) at (3.5, -0.5) {};
		\node [style=none] (82) at (1.75, 1) {};
		\node [style=none] (83) at (2.75, 1) {};
		\node [style=none] (84) at (1.75, 6) {};
		\node [style=node] (85) at (2.75, 2.5) {};
		\node [style=none] (86) at (1.75, 6.75) {$B$};
		\node [style=none] (87) at (3.25, 1.5) {$B'$};
		\node [style=none] (89) at (2.25, 0) {\textsf{Loss}};
		\node [style=none] (92) at (5.5, 1) {};
		\node [style=none] (93) at (5.5, -1) {};
		\node [style=none] (94) at (7.5, -1) {};
		\node [style=none] (95) at (8.5, 0) {};
		\node [style=none] (96) at (7.5, 1) {};
		\node [style=none] (97) at (5.5, -0.5) {};
		\node [style=none] (98) at (5.5, 0.5) {};
		\node [style=none] (99) at (6.75, 0) {$\alpha$};
		\node [style=none] (100) at (-8, 3) {};
		\node [style=none] (101) at (-5.5, 3) {};
		\node [style=none] (102) at (-5.5, 5) {};
		\node [style=none] (103) at (-8, 5) {};
		\node [style=none] (105) at (-7.25, 3) {};
		\node [style=none] (106) at (-6.25, 3) {};
		\node [style=none] (108) at (-7.25, 5) {};
		\node [style=none] (110) at (-6.25, 6) {};
		\node [style=none] (111) at (-7.25, 6) {};
		\node [style=none] (120) at (-8, 1) {};
		\node [style=none] (121) at (-8, -1) {};
		\node [style=none] (122) at (-5.5, 1) {};
		\node [style=none] (123) at (-5.5, -1) {};
		\node [style=none] (124) at (-8, 0.5) {};
		\node [style=none] (125) at (-8, -0.5) {};
		\node [style=none] (126) at (-5.5, 0.5) {};
		\node [style=none] (127) at (-5.5, -0.5) {};
		\node [style=none] (128) at (-7.25, 1) {};
		\node [style=none] (129) at (-6.25, 1) {};
		\node [style=none] (130) at (-2.25, 0) {\textsf{Model}};
		\node [style=none] (131) at (-6.75, 0.5) {};
		\node [style=none] (132) at (-6.25, 0) {};
		\node [style=none] (133) at (-5.75, -0.5) {};
		\node [style=none] (136) at (-7.75, 6.75) {$S \times A$};
		\node [style=none] (137) at (-5.75, 6.75) {$S \times A$};
		\node [style=none] (138) at (-2.75, 6) {};
		\node [style=none] (139) at (-2.75, 6.75) {$P$};
		\node [style=none] (140) at (-2.75, 1) {};
		\node [style=none] (141) at (-1.75, 1) {};
		\node [style=node] (142) at (-1.75, 2.5) {};
		\node [style=none] (143) at (-6.25, 5) {};
		\node [style=none] (144) at (-6.75, 4) {\textsf{Optimiser}};
	\end{pgfonlayer}
	\begin{pgfonlayer}{edgelayer}
		\draw (20.center) to (22.center);
		\draw (22.center) to (23.center);
		\draw (23.center) to (21.center);
		\draw (21.center) to (20.center);
		\draw [style=pointy] (39.center) to (40.center);
		\draw (43.center) to (44.center);
		\draw (45.center) to (46.center);
		\draw [style=pointy] (57.center) to (58.center);
		\draw [style=pointy] (62.center) to (60.center);
		\draw (61.center) to (63.center);
		\draw (66.center) to (68.center);
		\draw (68.center) to (69.center);
		\draw (69.center) to (67.center);
		\draw (67.center) to (66.center);
		\draw [style=pointy] (74.center) to (75.center);
		\draw (76.center) to (77.center);
		\draw (78.center) to (79.center);
		\draw [style=pointy] (80.center) to (81.center);
		\draw [style=pointy] (84.center) to (82.center);
		\draw (83.center) to (85);
		\draw (92.center) to (96.center);
		\draw (96.center) to (95.center);
		\draw (95.center) to (94.center);
		\draw (94.center) to (93.center);
		\draw (93.center) to (92.center);
		\draw [style=pointy] (79.center) to (98.center);
		\draw (97.center) to (80.center);
		\draw (103.center) to (102.center);
		\draw (102.center) to (101.center);
		\draw (101.center) to (100.center);
		\draw (100.center) to (103.center);
		\draw (62.center) to (105.center);
		\draw [style=pointy] (63.center) to (106.center);
		\draw [style=pointy] (111.center) to (108.center);
		\draw [style=lightnone] (120.center) to (122.center);
		\draw [style=lightnone] (122.center) to (123.center);
		\draw [style=lightnone] (123.center) to (121.center);
		\draw [style=lightnone] (121.center) to (120.center);
		\draw (126.center) to (39.center);
		\draw [style=pointy] (44.center) to (127.center);
		\draw (126.center) to (131.center);
		\draw [in=-90, out=-180, looseness=1.25] (131.center) to (128.center);
		\draw [bend right=315, looseness=1.25] (133.center) to (132.center);
		\draw (127.center) to (133.center);
		\draw (132.center) to (129.center);
		\draw (141.center) to (142);
		\draw [style=pointy] (138.center) to (140.center);
		\draw [style=pointy] (143.center) to (110.center);
	\end{pgfonlayer}
\end{tikzpicture}
  }
  \caption{Deep dreaming: supervised learning of inputs}
  \label{fig:deep_dreaming}
\end{figure}
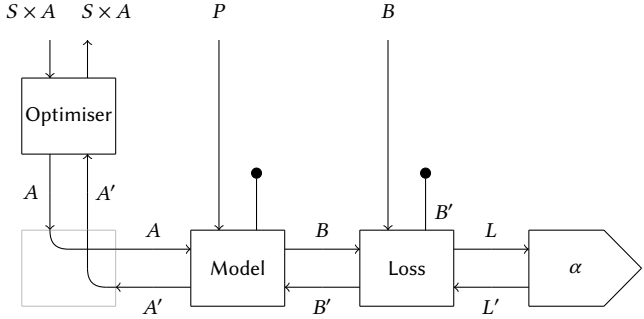

Deep dreaming is a technique which uses the parameters $p$ of some trained
classifier network to iteratively dream up, or amplify some features of a class $b$ on a chosen input $a$.
For example, if we start with an image of a landscape $a_0$, a label $b$ of a ``cat''
and a parameter $p$ of a sufficiently well-trained classifier, we can start performing
``learning'' as usual: computing the predicted class for the landscape $a_0$
for the network with parameters $p$, and then computing the distance between the
prediction and our label of a cat $b$. When performing backpropagation, the
respective changes computed for each layer tell us how the activations of that
layer should have been changed to be more ``cat'' like. This includes the first
(input) layer of the landscape $a_0$. Usually, we discard this changes and
apply gradient update to the parameters.
In deep dreaming we \textit{discard the parameters} and \textit{apply gradient update to the input} (Figure ~\ref{fig:deep_dreaming}).
Gradient update here takes these changes and computes a new image $a_1$ which
is the same image of the landscape, but changed slightly so to look more like
whatever the network thinks a cat looks like.
This is the essence of deep dreaming, where iteration of this process allows
networks to dream up features and shapes on a particular chosen image
\cite{GoogleInceptionism}.

Just like in the previous subsection, we can write this deep dreaming system as
a map in $\Para{\lens{\Ca}}$ from $(1, 1)$ to $(1, 1)$ whose parameter space is
$(S \times A \times P \times B, S \times A)$. In other words, this is a lens of
type $(S \times A \times P \times B, S \times A) \to (1, 1)$ whose $\gett{}$ map is trivial. Its $\putt{}$ map has the type $S \times A \times P \times B \to S \times A$ and unpacks to
\begin{align*}
  \putt{}(s, a, p, b_t) = U^*(s, a, a') \qquad \texttt{ where} \quad \qquad \overline{a}  &= U(s, a) \\
  b_p &= f(p, \overline{a}) \\
  (b'_t, b'_p) &= R[\loss](b_t, b_p, \alpha(\loss(b_t, b_p))) \\
  (p', a') &= R[f](p, \overline{a}, b'_p)
\end{align*}

We note that deep dreaming is usually presented without any loss function as a
maximisation of a particular activation in the last layer of the network output
\cite[Section 2.]{DeepDreaming4}. This maximisation is done with gradient
ascent, as opposed to gradient descent.
However, this is just a special case of our framework where the loss function is
the dot product (Example \ref{ex:dot_product}). The choice of the particular
activation is encoded as a one-hot vector, and the loss function in that case essentially masks the network output, leaving active only the particular chosen activation. The final component is the gradient \textit{ascent}: this is simply recovered by choosing
a positive, instead of a negative learning rate \cite{DeepDreaming4}. We
explicitly unpack this in the following example.

\begin{example}[Deep dreaming, dot product loss, gradient descent] Fix the base category to $\Smooth$ and a parameterised map $(\Rb^p, f) :
\Para{\Smooth}(\Rb^a, \Rb^b)$.
Fix the dot product loss (Example \ref{ex:dot_product}), basic gradient
descent (Example \ref{ex:grad_desc_lens}), and a \textit{positive} learning rate
$\alpha : \Rb$. Then the above put map simplifies.
Since there is no state, its type reduces to $A \times P \times B \to A$ and its
implementation to
\begin{align*}
  \putt{}(a, p, b_t) = a + a' &\qquad \qquad \qquad \qquad \texttt{  where }
  (p', a') = R[f](p, a, \alpha \cdot b_t).
\end{align*}
Like in Example \ref{ex:quad_err_grad_desc}, this update can be rewritten as
\[
  \putt{}(a, p, b_t) = a + \alpha \cdot (R[f](p, a, b_t);\pi_1)
\]
making a few things apparent. This update does not depend on the prediction
$f(p, a)$: no matter what the network has predicted, the goal is always to
maximize particular activations. Which activations? The ones chosen by $b_t$.
When $b_t$ is a one-hot vector, this picks out the activation of just one class
to maximize, which is often done in practice.
\end{example}

While we present only the most basic image, there is plenty of room left for
exploration. The work of \cite[Section 2.]{DeepDreaming4} adds an extra
regularization term to the image. In general, the neural network $f$ is sometimes changed to copy a number of internal activations which are then exposed on the output layer. Maximizing all these activations often produces more visually appealing results.
In the literature we did not find an example which uses the Softmax-cross
entropy (Example \ref{ex:softmax_ce}) as a loss function in deep dreaming, which
seems like the more natural choice in this setting. Furthermore, while deep
dreaming is commonly done with basic gradient descent, there is nothing
preventing one from doing deep dreaming with any of the optimizer lenses
discussed in the previous section, or even doing deep dreaming in the context of
Boolean circuits. Lastly, learning iteration which was described in at the end
of previous subsection can be modelled here in an analogous way.

\section{Implementation}
\label{section:implementation}
We provide a proof-of-concept implementation
as a Python library.\footnote{
  Full usage examples, source code, and experiments using our proof-of-concept can be
  found at \href{\codelink}{\codelink}.
}
We demonstrate the correctness of our library empirically using a number of
experiments implemented both in our library and in Keras\cite{Keras}, a popular
framework for deep learning.
For example, one experiment is a model for the MNIST image classification
problem\cite{mnist}: we implement the same model in both frameworks and achieve
comparable accuracy.

Our implementation also demonstrates the advantages of our approach.
Firstly, computing the gradients of the network is greatly simplified
through the use of lens composition.
Secondly, model architectures can be expressed in a principled, mathematical
language; as morphisms of a monoidal category.
Finally, the modularity of our approach makes it easy to see how various aspects
of training can be modified: for example, one can define a new optimization
algorithm simply by defining an appropriate lens. We now give a brief sketch of our implementation.

\subsection{Constructing a Model with Lens and Para}

We model a lens $(f, f^*)$ in our library with the \texttt{Lens} class, which
consists of a pair of maps \texttt{fwd} and \texttt{rev} corresponding to $f$
and $f^*$, respectively.
For example, we write the identity lens $(1_A, \pi_2)$ as follows:

\begin{lstlisting}[
  %caption=Identity Lens,
  label=lst:identity-lens,
]
identity = Lens(lambda x: x, lambda x_dy: x_dy[1])
\end{lstlisting}

The composition (in diagrammatic order) of \texttt{Lens} values \texttt{f} and
\texttt{g} is written \texttt{f >> g}, and monoidal composition as \texttt{f @
g}.
Similarly, the type of $\para$ maps is modeled by the \texttt{Para} class, with
composition and monoidal product written the same way.
Our library provides several primitive \texttt{Lens} and \texttt{Para} values.

Let us now see how to construct a single layer neural network from the
composition of such primitives.
Diagramatically, we wish to construct the following model, representing a
single `dense' layer of a neural network:

\begin{equation}\label{eq:paralens-dense-layer}
   \scalebox{0.8}{\begin{tikzpicture}
	\begin{pgfonlayer}{nodelayer}
		\node [style=none] (20) at (-3, 1) {};
		\node [style=none] (21) at (-3, -1) {};
		\node [style=none] (22) at (-1, 1) {};
		\node [style=none] (23) at (-1, -1) {};
		\node [style=none] (26) at (-4.5, 0.5) {$\Rb^a$};
		\node [style=none] (27) at (-4.5, -0.5) {$\Rb^a$};
		\node [style=none] (28) at (0, 1) {$\Rb^b$};
		\node [style=none] (39) at (-4, 0.5) {};
		\node [style=none] (40) at (-3, 0.5) {};
		\node [style=none] (43) at (-3, -0.5) {};
		\node [style=none] (44) at (-4, -0.5) {};
		\node [style=none] (45) at (-1, 0.5) {};
		\node [style=none] (46) at (0, 0.5) {};
		\node [style=none] (57) at (0, -0.5) {};
		\node [style=none] (58) at (-1, -0.5) {};
		\node [style=none] (60) at (-2.5, 1) {};
		\node [style=none] (61) at (-1.5, 1) {};
		\node [style=none] (62) at (-2.5, 2) {};
		\node [style=none] (63) at (-1.5, 2) {};
		\node [style=none] (64) at (-2.75, 2.5) {$\Rb^{b \times a}$};
		\node [style=none] (65) at (-1, 2.5) {$\Rb^{b \times a}$};
		\node [style=none] (66) at (1, 1) {};
		\node [style=none] (67) at (1, -1) {};
		\node [style=none] (68) at (3, 1) {};
		\node [style=none] (69) at (3, -1) {};
		\node [style=none] (74) at (0, 0.5) {};
		\node [style=none] (75) at (1, 0.5) {};
		\node [style=none] (76) at (1, -0.5) {};
		\node [style=none] (77) at (0, -0.5) {};
		\node [style=none] (78) at (3, 0.5) {};
		\node [style=none] (79) at (4, 0.5) {};
		\node [style=none] (80) at (4, -0.5) {};
		\node [style=none] (81) at (3, -0.5) {};
		\node [style=none] (82) at (1.5, 1) {};
		\node [style=none] (83) at (2.5, 1) {};
		\node [style=none] (84) at (1.5, 2) {};
		\node [style=none] (85) at (2.5, 2) {};
		\node [style=none] (86) at (1.5, 2.5) {$\Rb^b$};
		\node [style=none] (87) at (2.75, 2.5) {$\Rb^b$};
		\node [style=none] (88) at (-2, 0) {\textsf{linear}};
		\node [style=none] (89) at (2, 0) {\textsf{bias}};
		\node [style=none] (92) at (4, 0.5) {};
		\node [style=none] (93) at (4, -0.5) {};
		\node [style=none] (94) at (5, 1) {};
		\node [style=none] (95) at (5, -1) {};
		\node [style=none] (96) at (8, 1) {};
		\node [style=none] (97) at (8, -1) {};
		\node [style=none] (100) at (4, 0.5) {};
		\node [style=none] (101) at (5, 0.5) {};
		\node [style=none] (102) at (5, -0.5) {};
		\node [style=none] (103) at (4, -0.5) {};
		\node [style=none] (104) at (8, 0.5) {};
		\node [style=none] (105) at (9, 0.5) {};
		\node [style=none] (106) at (9, -0.5) {};
		\node [style=none] (107) at (8, -0.5) {};
		\node [style=none] (108) at (5.5, 1) {};
		\node [style=none] (109) at (7.5, 1) {};
		\node [style=none] (114) at (6.5, 0) {\textsf{activation}};
		\node [style=none] (115) at (0, -1) {$\Rb^b$};
		\node [style=none] (116) at (4, 1) {$\Rb^b$};
		\node [style=none] (117) at (4, -1) {$\Rb^b$};
		\node [style=none] (118) at (9.75, 0.5) {$\Rb^b$};
		\node [style=none] (119) at (9.75, -0.5) {$\Rb^b$};
	\end{pgfonlayer}
	\begin{pgfonlayer}{edgelayer}
		\draw (20.center) to (22.center);
		\draw (22.center) to (23.center);
		\draw (23.center) to (21.center);
		\draw (21.center) to (20.center);
		\draw [style=pointy] (39.center) to (40.center);
		\draw [style=pointy] (43.center) to (44.center);
		\draw (45.center) to (46.center);
		\draw [style=pointy] (57.center) to (58.center);
		\draw [style=pointy] (62.center) to (60.center);
		\draw [style=pointy] (61.center) to (63.center);
		\draw (66.center) to (68.center);
		\draw (68.center) to (69.center);
		\draw (69.center) to (67.center);
		\draw (67.center) to (66.center);
		\draw [style=pointy] (74.center) to (75.center);
		\draw (76.center) to (77.center);
		\draw [style=pointy] (78.center) to (79.center);
		\draw [style=pointy] (80.center) to (81.center);
		\draw [style=pointy] (84.center) to (82.center);
		\draw [style=pointy] (83.center) to (85.center);
		\draw (94.center) to (96.center);
		\draw (96.center) to (97.center);
		\draw (97.center) to (95.center);
		\draw (95.center) to (94.center);
		\draw [style=pointy] (100.center) to (101.center);
		\draw (102.center) to (103.center);
		\draw [style=pointy] (104.center) to (105.center);
		\draw [style=pointy] (106.center) to (107.center);
	\end{pgfonlayer}
\end{tikzpicture}}
\end{equation}

Here, the parameters of \texttt{linear} are the coefficients of a $b \times a$ matrix,
and the underlying lens has as its forward map the function $(M, x) \to M \cdot x$, where $M$ is the $b \times a$ matrix whose coefficients are the
$\Rb^{b \times a}$ parameters, and $x \in \Rb^a$ is the input vector.  The
\texttt{bias} map is even simpler: the forward map of the underlying lens is
simply pointwise addition of inputs and parameters: $(b, x) \to b + x$.
Finally, the \texttt{activation} map simply applies a nonlinear function (e.g., $\textsf{sigmoid}$) to the input,
and thus has the trivial (unit) parameter space.
The representation of this composition in code is straightforward: we can simply
compose the three primitive \texttt{Para} maps as in \eqref{eq:paralens-dense-layer}:

\begin{lstlisting}[
  %caption={The \texttt{dense} lens},
  label=lst:dense-lens,
]
def dense(a, b, activation):
  return linear(a, b) >> bias(b) >> activation
\end{lstlisting}

Note that by constructing model architectures in this way, the computation of
reverse derivatives is greatly simplified: we obtain the reverse
derivative `for free' as the $\putt$ map of the model.
Furthermore, adding new primitives is also simplified: the user need simply
provide a function and its reverse derivative in the form of a \texttt{Para}
map.
Finally, notice also that our approach is truly compositional: we can define a
hidden layer neural network with $n$ hidden units simply by composing two dense
layers, as follows:
\begin{lstlisting}[
  %caption={The \texttt{dense} lens},
  label=lst:dense-lens,
]
dense(a, n, activation) >> dense(n, b, activation)
\end{lstlisting}

\subsection{Learning}

Now that we have constructed a model, we also need to use it to \emph{learn} from data.
Concretely, we will construct a full parametric lens as in
Figure ~\ref{fig:paralens-full-learner}
then extract its $\putt$ map to iterate over the dataset.

By way of example, let us see how to construct the following parametric lens,
representing basic gradient descent over a single layer neural network with a
fixed learning rate:

\begin{equation}\label{eq:paralens-implementation-full-learner}
   \scalebox{0.8}{\begin{tikzpicture}
	\begin{pgfonlayer}{nodelayer}
		\node [style=none] (20) at (-3.5, 1) {};
		\node [style=none] (21) at (-3.5, -1) {};
		\node [style=none] (22) at (-1, 1) {};
		\node [style=none] (23) at (-1, -1) {};
		\node [style=none] (26) at (-4.5, 1) {$A$};
		\node [style=none] (27) at (-4.5, -1) {$A'$};
		\node [style=none] (28) at (0, 1) {$B$};
		\node [style=none] (29) at (0, -1) {$B'$};
		\node [style=none] (39) at (-4.5, 0.5) {};
		\node [style=none] (40) at (-3.5, 0.5) {};
		\node [style=none] (43) at (-3.5, -0.5) {};
		\node [style=none] (44) at (-4.5, -0.5) {};
		\node [style=none] (45) at (-1, 0.5) {};
		\node [style=none] (46) at (0, 0.5) {};
		\node [style=none] (57) at (0, -0.5) {};
		\node [style=none] (58) at (-1, -0.5) {};
		\node [style=none] (60) at (-2.75, 1) {};
		\node [style=none] (61) at (-1.75, 1) {};
		\node [style=none] (62) at (-2.75, 2) {};
		\node [style=none] (63) at (-1.75, 2) {};
		\node [style=none] (64) at (-3.25, 2) {$P$};
		\node [style=none] (65) at (-1.25, 2) {$P'$};
		\node [style=none] (66) at (1, 1) {};
		\node [style=none] (67) at (1, -1) {};
		\node [style=none] (68) at (3.5, 1) {};
		\node [style=none] (69) at (3.5, -1) {};
		\node [style=none] (72) at (4.5, 1) {$L$};
		\node [style=none] (73) at (4.5, -1) {$L'$};
		\node [style=none] (74) at (0, 0.5) {};
		\node [style=none] (75) at (1, 0.5) {};
		\node [style=none] (76) at (1, -0.5) {};
		\node [style=none] (77) at (0, -0.5) {};
		\node [style=none] (78) at (3.5, 0.5) {};
		\node [style=none] (79) at (4.5, 0.5) {};
		\node [style=none] (80) at (4.5, -0.5) {};
		\node [style=none] (81) at (3.5, -0.5) {};
		\node [style=none] (82) at (1.75, 1) {};
		\node [style=none] (83) at (2.75, 1) {};
		\node [style=none] (84) at (1.75, 6) {};
		\node [style=node] (85) at (2.75, 2.5) {};
		\node [style=none] (86) at (1.75, 6.5) {$B$};
		\node [style=none] (87) at (3.25, 1.5) {$B'$};
		\node [style=none] (89) at (2.25, 0) {$\mathsf{loss}$};
		\node [style=none] (92) at (5.5, 1) {};
		\node [style=none] (93) at (5.5, -1) {};
		\node [style=none] (94) at (7.5, -1) {};
		\node [style=none] (95) at (8.5, 0) {};
		\node [style=none] (96) at (7.5, 1) {};
		\node [style=none] (97) at (5.5, -0.5) {};
		\node [style=none] (98) at (5.5, 0.5) {};
		\node [style=none] (99) at (6.25, -0.5) {$\epsilon$};
		\node [style=none] (100) at (-3.5, 3) {};
		\node [style=none] (101) at (-1, 3) {};
		\node [style=none] (102) at (-1, 5) {};
		\node [style=none] (103) at (-3.5, 5) {};
		\node [style=none] (105) at (-2.75, 3) {};
		\node [style=none] (106) at (-1.75, 3) {};
		\node [style=none] (107) at (-2.75, 4) {};
		\node [style=none] (108) at (-2.75, 5) {};
		\node [style=none] (110) at (-1.75, 6) {};
		\node [style=none] (111) at (-2.75, 6) {};
		\node [style=none] (112) at (-2, 4.25) {};
		\node [style=none] (113) at (-2, 3.75) {};
		\node [style=none] (114) at (-1.75, 3.75) {};
		\node [style=none] (115) at (-1.5, 3.75) {};
		\node [style=none] (116) at (-1.5, 4.25) {};
		\node [style=none] (117) at (-1.75, 4.25) {};
		\node [style=none] (118) at (-2, 4) {};
		\node [style=none] (119) at (-1.75, 4) {$+$};
		\node [style=none] (120) at (-8, 1) {};
		\node [style=none] (121) at (-8, -1) {};
		\node [style=none] (122) at (-5.5, 1) {};
		\node [style=none] (123) at (-5.5, -1) {};
		\node [style=none] (124) at (-8, 0.5) {};
		\node [style=none] (125) at (-8, -0.5) {};
		\node [style=none] (126) at (-5.5, 0.5) {};
		\node [style=none] (127) at (-5.5, -0.5) {};
		\node [style=none] (128) at (-7.25, 1) {};
		\node [style=none] (129) at (-6.25, 1) {};
		\node [style=none] (130) at (-2.25, 0) {$\mathsf{dense}$};
		\node [style=none] (131) at (-6.75, 0.5) {};
		\node [style=none] (132) at (-6.25, 0) {};
		\node [style=none] (133) at (-5.75, -0.5) {};
		\node [style=none] (134) at (-7.25, 6) {};
		\node [style=node] (135) at (-6.25, 2.5) {};
		\node [style=none] (136) at (-2.75, 6.75) {$P$};
		\node [style=none] (137) at (-1.75, 6.75) {$P$};
		\node [style=none] (138) at (-7.25, 6.75) {$A$};
		\node [style=node] (139) at (6.5, 0.5) {};
		\node [style=none] (140) at (6, -0.25) {};
		\node [style=none] (141) at (6, -0.75) {};
		\node [style=none] (142) at (6.5, -0.75) {};
		\node [style=none] (143) at (6.5, -0.25) {};
		\node [style=none] (144) at (6.75, -0.5) {};
		\node [style=none] (145) at (6, -0.5) {};
	\end{pgfonlayer}
	\begin{pgfonlayer}{edgelayer}
		\draw (20.center) to (22.center);
		\draw (22.center) to (23.center);
		\draw (23.center) to (21.center);
		\draw (21.center) to (20.center);
		\draw [style=pointy] (39.center) to (40.center);
		\draw (43.center) to (44.center);
		\draw (45.center) to (46.center);
		\draw [style=pointy] (57.center) to (58.center);
		\draw [style=pointy] (62.center) to (60.center);
		\draw (61.center) to (63.center);
		\draw (66.center) to (68.center);
		\draw (68.center) to (69.center);
		\draw (69.center) to (67.center);
		\draw (67.center) to (66.center);
		\draw [style=pointy] (74.center) to (75.center);
		\draw (76.center) to (77.center);
		\draw (78.center) to (79.center);
		\draw [style=pointy] (80.center) to (81.center);
		\draw [style=pointy] (84.center) to (82.center);
		\draw (83.center) to (85);
		\draw (92.center) to (96.center);
		\draw (96.center) to (95.center);
		\draw (95.center) to (94.center);
		\draw (94.center) to (93.center);
		\draw (93.center) to (92.center);
		\draw [style=pointy] (79.center) to (98.center);
		\draw [style=pointy] (97.center) to (80.center);
		\draw (103.center) to (102.center);
		\draw (102.center) to (101.center);
		\draw (101.center) to (100.center);
		\draw (100.center) to (103.center);
		\draw (62.center) to (105.center);
		\draw (63.center) to (106.center);
		\draw (111.center) to (108.center);
		\draw (108.center) to (107.center);
		\draw (107.center) to (105.center);
		\draw [style=pointy] (107.center) to (118.center);
		\draw [style=pointy] (117.center) to (110.center);
		\draw [style=pointy] (106.center) to (114.center);
		\draw (113.center) to (112.center);
		\draw (112.center) to (116.center);
		\draw (116.center) to (115.center);
		\draw (115.center) to (113.center);
		\draw [style=lightnone] (120.center) to (122.center);
		\draw [style=lightnone] (122.center) to (123.center);
		\draw [style=lightnone] (123.center) to (121.center);
		\draw [style=lightnone] (121.center) to (120.center);
		\draw (126.center) to (39.center);
		\draw [style=pointy] (44.center) to (127.center);
		\draw (126.center) to (131.center);
		\draw [in=-90, out=-180, looseness=1.25] (131.center) to (128.center);
		\draw [bend right=315, looseness=1.25] (133.center) to (132.center);
		\draw (127.center) to (133.center);
		\draw (132.center) to (129.center);
		\draw [style=pointy] (134.center) to (128.center);
		\draw (129.center) to (135);
		\draw (98.center) to (139);
		\draw (143.center) to (144.center);
		\draw (144.center) to (142.center);
		\draw (142.center) to (141.center);
		\draw (141.center) to (140.center);
		\draw (140.center) to (143.center);
		\draw (97.center) to (145.center);
	\end{pgfonlayer}
\end{tikzpicture}}
\end{equation}

This morphism is constructed essentially as below, where
\texttt{apply\_update($\alpha$, $f$)} represents the `vertical stacking' of
$\alpha$ atop $f$:

\begin{lstlisting}[mathescape]
apply_update(basic_update, dense) >> loss >> learning_rate($\epsilon$)
\end{lstlisting}

Now, given the parametric lens of \eqref{eq:paralens-implementation-full-learner},
one can construct a morphism $\textsf{step} : B \times P \times A \to P$ which is simply the put map of the lens.
Training the model then consists of iterating the $\textsf{step}$ function over
dataset examples $(x, y) \in A \times B$ to optimise some initial choice of
parameters $\theta_0 \in P$, by letting
$\theta_{i+1} = \mathsf{step}(y_i, \theta_i, x_i)$.

Note that our library also provides a utility function to construct $\mathsf{step}$ from its various pieces:
\begin{lstlisting}[mathescape]
step = supervised_step(model, update, loss, learning_rate)
\end{lstlisting}

For an end-to-end example of model training and iteration, we refer the
interested reader to the experiments accompanying the code:
\href{\codelink}{\codelink}.


\section{Related Work}
\label{section:related-work}
The work \cite{BackpropAsFunctor} is closely related to ours, in that it
provides an abstract categorical model of back-propagation. However, it differs in
a number of key aspects.
We give a complete lens-theoretic explanation of {\em what} is
back-propagated via (i) the use of CRDCs to model gradients; and (ii) the $\para$ construction to model parameterized functions and
parameter update.
We thus can go well beyond \cite{BackpropAsFunctor} in terms of examples - their example of
smooth functions and basic gradient descent is covered in our subsection
\ref{subsec:learning-parameters}.

We also explain some of the constructions of \cite{BackpropAsFunctor} in a more
structured way.  For example, rather than considering the category $\Learn$ of
\cite{BackpropAsFunctor} as primitive, here we construct it as a composite of two
more basic constructions (the $\para$ and $\textbf{Lens}$ constructions).
The flexibility could be used, for example, to compositionally replace $\para$
with a variant allowing parameters to come from a different category, or lenses
with the category of optics \cite{Optics} enabling us to model things such as control flow using prisms. 

One more important thing is related to functoriality. We use a functor to augment a
parameterised map with its backward pass, just like \cite{BackpropAsFunctor}.
However, they additionally augmented this map with a loss map and gradient
descent using a functor as well. This added extra conditions on the partial
derivatives of the loss function: it needed to be invertible in the 2nd
variable. This constraint was not justified in \cite{BackpropAsFunctor}, nor is
it a constraint that appears in machine learning
practice. This led us to reexamine their constructions, coming up with our
reformulation that does not require it.
While loss maps and optimisers are mentioned in \cite{BackpropAsFunctor} as parts of the aforementioned functor, here they are extracted out and play a key role: loss maps are parameterised lenses and optimisers are reparameterisations.
Thus, in this paper we instead use $\para$-composition
to add the loss map to the model, and $\para$ 2-cells to add optimisers.
The mentioned inverse of the partial derivative of the loss map in the 2nd variable was also hypothesised to be relevant to deep dreaming. In our paper we have given a complete picture of deep dreaming systems, showing it is gradient update which is used to dream
up pictures.

We also correct a small issue in Theorem III.2 of \cite{BackpropAsFunctor}.
There, the morphisms of $\Learn$ were defined up to an equivalence (pg. 4 of
\cite{BackpropAsFunctor}) but, unfortunately, the functor defined in Theorem
III.2 does not respect this equivalence relation. Our approach instead uses
2-cells which comes from the universal property of $\para$ ---  a 2-cell from
$(P,f): A \to B$ to $(Q,g): A \to B$ is a lens, and hence has two components: a
map $\alpha: Q \to P$ and $\alpha^*: Q \times P \to Q$.  By comparison, we can
see the equivalence relation of \cite{BackpropAsFunctor} as being induced by
map $\alpha: Q \to P$, and not a lens.  Our approach highlights the importance
of the 2-categorical structure of learners. In addition, it does not treat the
functor $\Para{\Ca} \to \Learn$ as a primitive. In our case, this functor has
the type $\Para{\Ca} \to \Para{\lens{\Ca}}$ and arises from applying $\para$ to a
canonical functor $\Ca \to \lens{\Ca}$ existing for \textit{any} reverse derivative
category, not just $\Smooth$.
Lastly, in our paper we have taken the advantage of the graphical calculus for
$\para$, redrawing many diagrams appearing in \cite{BackpropAsFunctor} in a structured way.

Other than \cite{BackpropAsFunctor}, there are a few more relevant papers.
The work of (\cite{Dioptics}) contains a sketch of some of the ideas this paper
evolved from. They are based on the interplay of optics with parameterisation,
albeit framed in the setting diffeological spaces, and requiring cartesian and
local cartesian closed structure on the base category.
Lenses and Learners are studied in the eponymous work of
\cite{LensesAndLearners} which observes that lenses are parameterised learners. They do not explore any of the
relevant $\para$ or CRDC structure, but make the distinction between \textit{symmetric}
and \textit{asymmetric lenses}, studying how they are related to learners
defined in \cite{BackpropAsFunctor}. A lens-like implementation of automatic differentiation is the focus of
\cite{SimpleAD}, but learning algorithms aren't studied.
A relationship between category-theoretic perspective on probabilistic modeling
and gradient-based optimisation is studied in \cite{CatStochLik} which also
studies a variant of the $\para$ construction.
Usage of Cartesian
differential categories to study learning is found in \cite{DelayedTrace}. They
extend the differential operator to work on stateful maps, but do not study
lenses, parameterisation nor update maps.
The work of \cite{CompDL} studies deep learning in the context of
Cycle-consistent Generative Adversarial Networks \cite{CycleGAN} and formalises
it via free and quotient categories, making parallels to the categorical
formulations of database theory \cite{FunctorialDataMigration}. They do use the
$\para$ construction, but do not relate it to lenses nor reverse derivative categories.

Lastly, the concept of parameterised lenses has started appearing in recent formulations
of categorical game theory and cybernetics \cite{TowardsCatCyber, ExtensiveFormGamesAgency}. The work of \cite{TowardsCatCyber} generalises the study of parameterised lenses
into parameterised optics and connects it to game thereotic concepts such as
Nash equilibria.
A general survey of category theoretic approaches to machine learning, covering
many of the above papers in detail, can be found in \cite{CategoryTheoryMachineLearning}.


\section{Conclusions and Future Directions}
\label{section:conclusions}


We have given a categorical foundation of gradient-based learning algorithms which achieves a number of important goals.  The foundation is principled and mathematically clean, based on the fundamental idea of a \emph{parameterised lens}.  The foundation covers a wide variety of examples: it covers different optimisers and loss maps in gradient-based learning, it covers different settings where gradient-based learning happens (smooth functions vs. boolean circuits) and it covers both learning of parameters and learning of inputs (deep dreaming).  Finally, the foundation is more than a mere abstraction: we have also shown how it can be used to give a practical implementation of learning, as discussed in section \ref{section:implementation}.

There are a number of important directions which are possible to explore because of this work. 
One of the most exciting ones is the extension to more complex neural network
architectures. Our formulation of the loss map as a parameterised lens should
pave the way for Generative Adversarial Networks \cite{GAN}, an exciting new architecture whose loss map can be said to be \textit{learned} in tandem with the base network.
In all our settings we have fixed an optimiser beforehand. The work of
\cite{LTL} describes a \textit{meta-learning} approach which sees the optimiser
as a neural network whose parameters and gradient update rule can be learned.
This is an exciting prospect since one can model optimisers as parameterised
lenses; and our framework covers learning with parameterised lenses. Recurrent neural networks are another
example of a more complex architecture, which has already been studied in the
context of differential categories in \cite{DelayedTrace}. When it comes to
architectures, future work includes modelling some classical systems as well, such as the Support Vector Machines \cite{SVM}, which should be possible with the usage of loss maps such as Hinge loss.

We have not made use of the full power of the CRDC axioms; in particular, we did not
explicitly need axioms RD.6 or RD.7, which deal with the behaviour of
higher-order derivatives.  However, some supervised learning algorithms do use
the higher-order derivatives (for example, the Hessian) for additional
optimisations; as such, future work includes exploring how to use those axioms
to capture these optimisations. Taking this idea in a different direction, one
can see that much of our work can be applied to any functor of the form $F: \Ca
\to \lens{\Ca}$ - $F$ does not necessarily have to be of the form $f \mapsto
(f,R[f])$ for a CRDC $R$.  Moreover, by working with more generalised forms of
the lens category (such as dependent lenses), we may be able to capture ideas
related to supervised learning on manifolds. And, of course, we can vary
the parameter space to endow it with different structure from the functions we
wish to learn. In this vein, we wish to use fibrations/dependent types to model
the use of tangent bundles: this would foster the extension of the {\em correct
by construction} paradigm to machine learning, and thereby addressing the widely
acknowledged problem of trusted machine learning. The possibilities are made
much easier by the compositional nature of our framework.
Another key topic for future work is to link gradient-based learning with game theory. At a
high level, the former takes little incremental steps to achieve an equilibrium
while the later aims to do so in one fell swoop. Formalising this intuition is
possible with our lens-based framework and the lens-based framework for game
theory~\cite{CompositionalGameTheory}.  Finally, because our framework is quite
general, in future work we plan to consider further modifications and additions
to encompass non-supervised, probabilistic and non-gradient based learning. This
includes genetic algorithms and reinforcement learning.




\bibliography{references}

\newpage
\appendix
\section{Background on Cartesian Reverse Differential Categories}

Here we briefly review the definitions of Cartesian left additive category (CLAC), Cartesian reverse differential category (CRDC) and additive and linear maps in these categories.  Note that in this appendix we follow the convention of \cite{CRDC} and write composition in diagrammatic order by juxtaposition of terms (rather than a semicolon) to shorten the form of many of the expressions.  

\begin{definition} A category $\Ca$ is said to be {\bf Cartesian} when there are chosen binary products $\times$, with projection maps $\pi_i$ and pairing operation $\langle - , - \rangle$, and a chosen terminal object $T$, with unique maps $!$ to the terminal object. 
\end{definition}

\begin{definition} A {\bf left additive category} \cite[Definition  1.1.1]{journal:BCS:CDC} (CLAC) is a category $\Ca$ such that each hom-set has  commutative monoid structure, with addition operation $+$ and zero maps 0, such that composition on the left preserves the additive structure: for any appropriate $f, g, h$, $f(g+h) = fg + fh$ and $f0 = 0$.
\end{definition}  

\begin{definition}
A map $h: X \to Y$ in a CLAC is \textbf{additive} if it has the property that it preserves additive structure by composition on the right: for any maps $x,y: Z \to X$, $(x+y);h = x;h + y;h$, and $0;h = 0$. 
\end{definition}

\begin{definition}
A {\bf Cartesian left additive category} \cite[Definition 1.2.1]{journal:BCS:CDC} is a left additive category $\Ca$ which is Cartesian and such that all projection maps $\pi_i$ are additive. 
\end{definition}

The central definition of \cite{CRDC} is the following:

\begin{definition}\label{def:crdc}
A \textbf{Cartesian reverse differential category} (CRDC) is a Cartesian left additive category $\Ca$ which has, for each map $f: A \to B$ in $\Ca$, a map
	\[ R[f]: A \times B \to A \]
satisfying seven axioms: \\
\noindent 
 {\bf [RD.1]}   $R[f+g] = R[f] + R[g]$ and $R[0]=0$. \\
\noindent  
{\bf [RD.2]} $\<a,b+c\>R[f] = \<a,b\>R[f] + \<a,c\>R[f]$ and  $\<a,0\>R[f] = 0$. \\
{\bf [RD.3]} 
    $R[1] = \pi_1$, $R[\pi_0] = \pi_1 \iota_0$, and $R[\pi_1] = \pi_1 \iota_1$.  
\noindent {\bf [RD.4]} 
   For a tupling of maps $f$ and $g$, the following equality holds:      
$$R[\<f,g\>] = (1 \times \pi_0);R[f]+ (1 \times \pi_1);R[g]$$
And if $!_A: A \to T$ is the unique map to the terminal object, $R[!_A] = 0$.
        
\noindent {\bf [RD.5]} 
For composable maps $f$ and $g$,
	\[ R[fg] = \<\pi_0,\pi_0f,\pi_1\>\>(1 \times R[g])R[f] \]
\noindent {\bf [RD.6]} 
\[ \<1\times \pi_0,0\times \pi_1\>(\iota_0 \times 1)R[R[R[f]]]\pi_1 = (1\times \pi_1)R[f] .  \] \\ 
\noindent {\bf [RD.7]} 
\[ (\iota_0 \times 1);R[R[(\iota_0 \times 1)R[R[f]]\pi_1]];\pi_1  = \mathsf{ex};(\iota_0 \times 1)R[R[(\iota_0 \times 1)R[R[f]]\pi_1]]\pi_1 \] (where $\mathsf{ex}$ is the map that exchanges the middle two variables).  
\end{definition}

As discussed in \cite{CRDC}, these axioms correspond to familiar properties of the reverse derivative:
\begin{itemize}
	\item  {\bf [RD.1]} says that differentiation preserves addition of maps, while  {\bf [RD.2]} says that differentiation is additive in its vector variable.
	\item {\bf [RD.3]} and {\bf [RD.4]} handle the derivatives of identities, projections, and tuples.  
	\item {\bf [RD.5]} is the (reverse) chain rule.  
	\item {\bf [RD.6]} says that the reverse derivative is linear in its vector variable.
	\item {\bf [RD.7]} expresses the independence of order of mixed partial derivatives.  
\end{itemize}  


%

%

\end{document}